\documentclass{article}

\usepackage[preprint,nonatbib]{utils/neurips_2024}
\usepackage{subcaption}

\usepackage[table,x11names]{xcolor} 
\usepackage[utf8]{inputenc} 
\usepackage[T1]{fontenc}    
\usepackage{marvosym}
\usepackage{url}            
\usepackage{booktabs}       
\usepackage{amsfonts}       
\usepackage{nicefrac}       
\usepackage{microtype}      
\usepackage{multirow}
\usepackage{pifont}
\usepackage{svg}
\usepackage{graphicx}
\usepackage{booktabs} 
\usepackage{tabularx} 
\usepackage{mathtools}
\usepackage{wrapfig}
\usepackage{enumitem}

\usepackage{tabularx}
\usepackage{geometry}
\usepackage{graphicx}
\usepackage{booktabs}
\usepackage{tcolorbox}

\definecolor{citecolor}{HTML}{0071bc}
\definecolor{action}{RGB}{201, 218, 248}
\definecolor{high}{RGB}{255, 242, 207}  
\definecolor{middle}{RGB}{218, 234, 213}
\definecolor{citecolor}{HTML}{0071bc}
\definecolor{darkblue}{RGB}{30,144,255} 
\definecolor{darkyellow}{RGB}{255,165,0} 
\definecolor{darkgreen}{RGB}{34,139,34} 

\newcommand{\doth}{\textcolor{darkyellow}{\ding{108}} \textbf{High-level Planning}}
\newcommand{\dotm}{\textcolor{darkgreen}{\ding{108}} \textbf{Middle-level Planning}}
\newcommand{\dota}{\textcolor{darkblue}{\ding{108}} \textbf{Atomic-action Execution}}

\newcommand{\dothh}{\textcolor{darkyellow}{\ding{108}} \textbf{High-level Plans}}
\newcommand{\dotmm}{\textcolor{darkgreen}{\ding{108}} \textbf{Mid.-level Plans}}
\newcommand{\dotaa}{\textcolor{darkblue}{\ding{108}} \textbf{Atomic Actions}}

\usepackage[pagebackref=false,breaklinks=true,letterpaper=true,colorlinks,citecolor=citecolor,bookmarks=false]{hyperref}
\usepackage{etoolbox, subcaption}
\usepackage[font=footnotesize,labelfont=bf,aboveskip=2pt,belowskip=-8pt]{caption}

\newdimen\abovecrulesep
\newdimen\belowcrulesep
\makeatletter
\patchcmd{\@@@cmidrule}{\aboverulesep}{\abovecrulesep}{}{}
\patchcmd{\@xcmidrule}{\belowrulesep}{\belowcrulesep}{}{}

\definecolor{demphcolor}{RGB}{144, 144, 144}
\definecolor{mygray}{gray}{0.4}
\definecolor{lightgray}{rgb}{0.9, 0.9, 0.9}

\newlength\savewidth

\newcommand{\tablestyle}[2]{\setlength{\tabcolsep}{#1}\renewcommand{\arraystretch}{#2}\centering\footnotesize}
\makeatletter\renewcommand\paragraph{\@startsection{paragraph}{4}{\z@}{.5em\@plus1ex\@minus.2ex}{-.5em}{\normalfont\normalsize\bfseries}}
\makeatother

\newcolumntype{C}[1]{>{\centering\arraybackslash}p{#1}}
\newcolumntype{R}[1]{>{\raggedleft\arraybackslash}p{#1}}
\newcolumntype{L}[1]{>{\raggedright\arraybackslash}p{#1}}

\usepackage{etoolbox}
\preto\align{\small}
\expandafter\preto\csname align*\endcsname{\small}
\preto\equation{\par\nobreak\small\noindent}
\newcommand{\our}{VideoGUI}

\newcommand{\ppt}{Powerpoint}
\newcommand{\pr}{Premiere Pro}
\newcommand{\ps}{Photoshop}
\newcommand{\eg}{\textit{e.g.,}}
\newcommand{\ie}{\textit{i.e.,}}
\usepackage{makecell}

\newcommand{\cmark}{\ding{51}}%

\newcommand{\llama}{LLama3-70B~\cite{llama3}}
\newcommand{\mixtral}{Mixtral-8x22B~\cite{mixtral}}
\newcommand{\chatgpt}{GPT-3.5-Turbo~\cite{chatgpt}}
\newcommand{\gptv}{GPT-4-Turbo~\cite{gpt4report}}
\newcommand{\gpto}{GPT-4o~\cite{gpt4report}}
\newcommand{\gemini}{Gemini-Pro-V~\cite{gemini}}
\newcommand{\claude}{Claude-3-Opus~\cite{claude}}
\newcommand{\cogagent}{CogAgent~\cite{cogagent}}
\newcommand{\qwen}{Qwen-VL-Max~\cite{Qwen_technicalReport}}

\title{\includegraphics[scale=0.06, bb=-150 80 540 34]{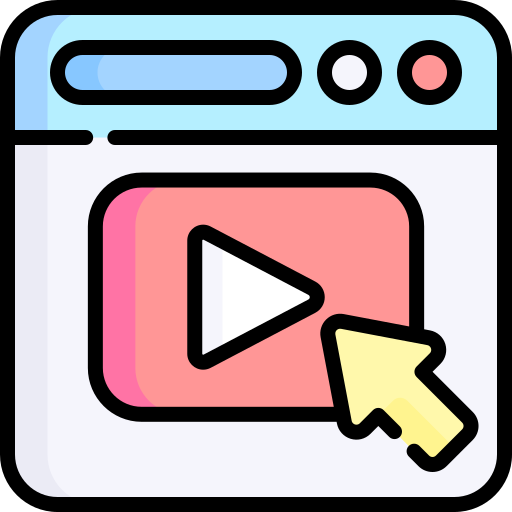}\our: A Benchmark for GUI Automation \\from Instructional Videos}

\author{%
	Kevin Qinghong Lin$^1$,~
	Linjie Li$^2$,~
	Difei Gao$^1$,~
	Qinchen Wu$^1$,~
        \\
        \textbf{
        Mingyi Yan$^1$,~ 
        Zhengyuan Yang$^2$,~ 
        Lijuan Wang$^2$,~
        Mike Zheng Shou$^1$\textsuperscript{\Letter}
        }
	\\
	$^1$Show Lab, National University of Singapore\quad
	$^2$Microsoft Gen AI\\
        \url{https://showlab.github.io/videogui/}
}

\begin{document}

\maketitle

\begin{abstract}
Graphical User Interface (GUI) automation holds significant promise for enhancing human productivity by assisting with computer tasks. Existing task formulations primarily focus on simple tasks that can be specified by a single, language-only instruction, such as ``Insert a new slide.'' 
In this work, we introduce \textbf{\our}, a novel multi-modal benchmark designed to evaluate GUI assistants on visual-centric GUI tasks.
Sourced from high-quality web instructional videos, our benchmark focuses on tasks involving professional and novel software (\eg~Adobe Photoshop or Stable Diffusion WebUI) and complex activities (\eg~video editing). 
\our~evaluates GUI assistants through a \textit{hierarchical} process, 
allowing for identification of the specific levels at which they may fail:
\textbf{($i$) high-level planning:} reconstruct procedural subtasks from visual conditions without language descriptions;
\textbf{($ii$) middle-level planning:} generate sequences of precise action narrations based on visual state (\ie~screenshot) and goals;
\textbf{($iii$) atomic action execution:} perform specific actions such as accurately clicking designated elements.
For each level, we design evaluation metrics across individual dimensions to provide clear signals, such as individual performance in clicking, dragging, typing, and scrolling for atomic action execution.
Our evaluation on \our~reveals that even the SoTA large multimodal model GPT4o performs poorly on visual-centric GUI tasks, especially for high-level planning. 




\end{abstract}

\vspace{-0.5cm}
\section{Introduction}
\vspace{-0.2cm}
In the digital age, individuals rely on computers for a vast array of daily activities (\eg web browsing, entertainment adn etc.).
These activities often necessitate the use of diverse software, which are accessed primarily through Graphical User Interfaces (GUIs).
Large language models (LLMs)~\cite{gpt4}, 
which excel in understanding complex language instructions and integrating various tools seamlessly, have shown great potential in GUI automation~\cite{copilot,webagentplan,webarena,multimodalweb}. They could streamline the navigation of digital interfaces and significantly enhance productivity, \eg assisting slide template creation in \ppt~with just a few keywords~\cite{copilot}.

Recently, 
notable efforts have been made in GUI automation evaluation, benchmarking model performances on Web~\cite{webarena,mind2web,miniwob++} or Smartphone GUI navigation~\cite{android_in_zoo,comprehensive_smartphone_agent}, given screenshots or HTML codes~\cite{aitw,empowering}.
Follow-up works ~\cite{openagents,agentstudio,pixel2act} develop executable environments with well-defined action spaces, which removes the dependencies on pre-defined inputs. 
Nonetheless, most existing GUI benchmarks~\cite{yao2022webshop,seeclick} restrict their applications to simpler domains and tasks that can be described with a single text instruction (\eg~``Insert a new slide on the second page''). In real-world scenarios, users rarely struggle with basic operations that can be clearly described in text. Rather, they often encounter difficulties in performing novel and advanced tasks (\eg~``Create a special animation effects in powerpoint''),
which extend far beyond basic operations, and rely more on visual signals than text instructions to complete such tasks.

\begin{figure}[!t]
	\centering
	\includegraphics[width=1.0\linewidth]{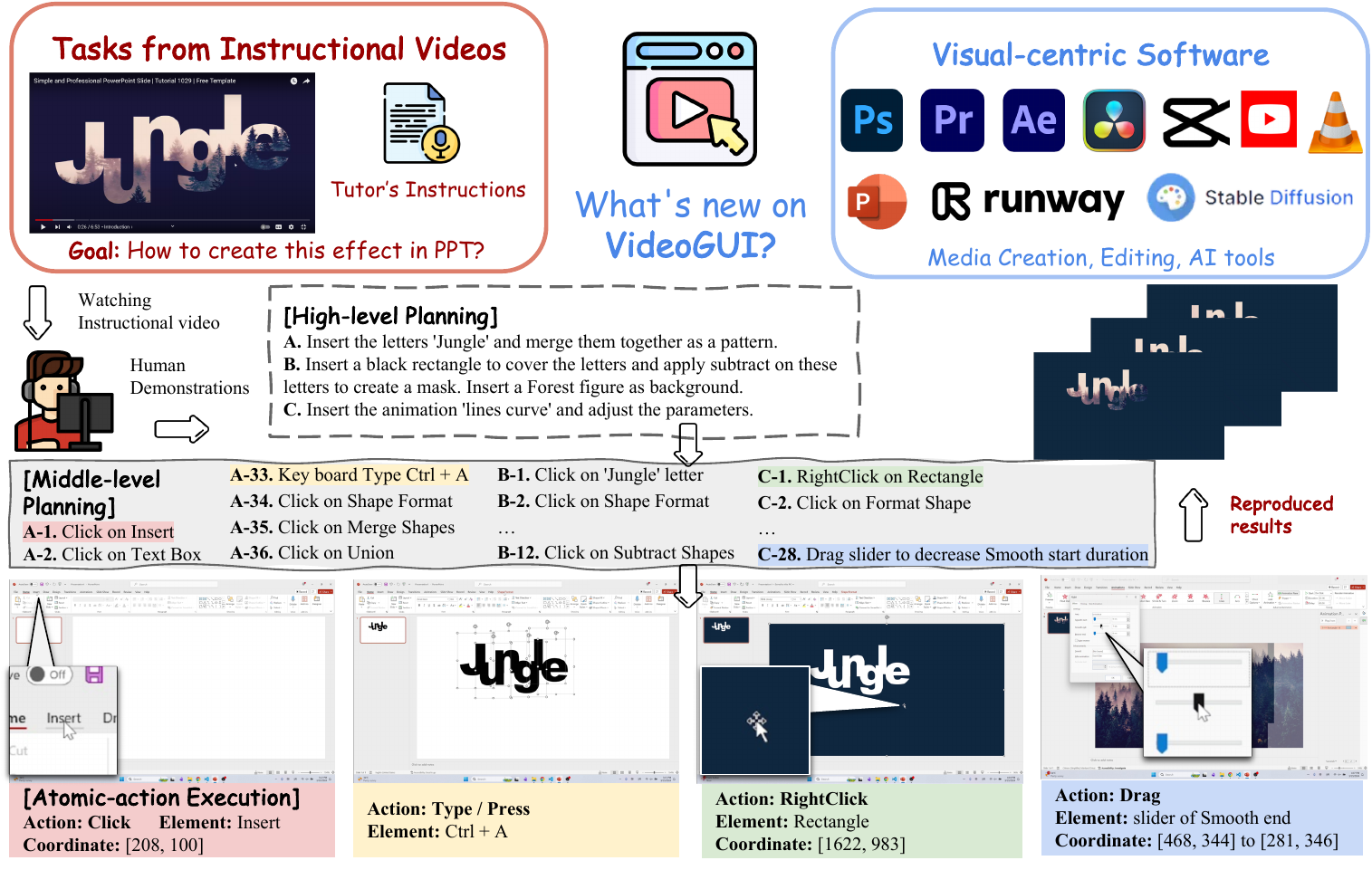}
	\caption{
\textbf{A brief illustration of \our}.
\our~focuses on professional and novel software like PR, AE for video editing, and Stable Diffusion, Runway for visual creation.
We source tasks from high-quality instructional videos  (see an example \href{www.youtube.com/watch?v=F2Enf6tjmy8}{here}), with annotators replicating these to reproduce effects;
We provide detailed annotations with planning procedures and recorded actions for hierarchical evaluation.
}
\label{fig:demo}
\end{figure}

Inspired by the abundant instructional videos that teach average users for performing novel and complex GUI tasks, we introduce  \our, a new  multi-modal GUI benchmark derived from high-quality web instructional videos. As shown in Fig.~\ref{fig:demo}, \our~provides high- quality annotations by having participants reproducing the instructional videos, capturing multi-level labels from procedural planning to atomic actions with element locations. 
\our~covers 11 visual-centric software applications and features 86 complex tasks (averaging 22.7 actions each) and 463 subtasks, alongside hierarchical manual planning and 2.7K manual action annotations (Tab.~\ref{tab:benchmarks}).

With \our,  we propose a comprehensive evaluation suite for GUI assistants via a \textit{hierarchical} process:
($i$) high-level planning involves reconstructing procedural subtasks from visual cues without language descriptions;
($ii$) middle-level planning details the steps for completing a subtask with a sequence of precise action narrations based on visual state and textual query;
($iii$) atomic action execution is to perform the target actions (\eg click on the designated element).
For each level, we design evaluation metrics across \textit{individual dimensions} to assess model performance, which help to pinpoint model limitations.

We conduct comprehensive evaluation of SoTA large multimodal models (LMMs) on \our, and find that even the current best model GPT-4o fails to complete a single full task in our benchmark. Our empirical results show that the bottleneck surprisingly lies in  planning rather than action execution, even though GPT-4o is not known for grounding. Moreover, planning from textual queries is much easier than planning from visual previews for almost all models evaluated, which further implies the difficulty of visual-centric GUI tasks. Our findings shed lights on the directions for developing the next generation of models or agent systems towards GUI automation. 

\begin{table*}[t]
    \centering
    \tablestyle{1.8pt}{0.98} 
    \resizebox{\textwidth}{!}
    {%
    \begin{tabular}{llccccclccc}
    \toprule
        \multirow{2}{*}{\textbf{Benchmark}} 
        & \multirow{2}{*}{\textbf{\# Task}} 
        & \multirow{2}{*}{\textbf{Platform}} 
        & \multirow{2}{*}{\textbf{Source}} 
        & \multicolumn{3}{c}{\textbf{Query format}}  
        & \multirow{2}{*}{\parbox{1cm}{\centering \textbf{\# Avg.}\\ \textbf{Action}}}
        & \multicolumn{3}{c}{\textbf{Eval. dimension}} \\
        \cmidrule(lr){5-7} \cmidrule(lr){9-11}
        & & & & Text & Image & Video & & Task SR. & Hier. Plan. &  Action Exec. \\
        \midrule
    Mind2Web~\cite{mind2web} & 2350 & Web & Screenshot &  \cmark & &  & 7.3 &  \cmark & &  \cmark  \\
    PixelHelp~\cite{pixelhelp} & 187 & Android & Emulator & \cmark & & & 4.2 & \cmark &  & \cmark  \\
    AITW~\cite{aitw} & 30K & Android & Emulator & \cmark & & &  6.5 & \cmark &  & \cmark \\
    AssistGUI~\cite{assistgui} & 100 & Windows & Web Video & \cmark &  &  & $-$ &
    \cmark &  &   \\
    OSWorld~\cite{osworld} & 369 & Win.+Ubuntu & Emulator & \cmark & & & -- & \cmark & & \\
    \midrule
    V-WebArena~\cite{vwebarena} & 910 & Web & Screenshot & \cmark & \cmark &  & -- & \cmark & &  \\
    \multirow{2}{*}{\textbf{\our}} 
    ~\textsc{Subtask} & 463 & \multirow{2}{*}{Win. +Web} & \multirow{2}{*}{\makecell{\textbf{Video~$+$ Human} \\ \textbf{Demonstration}}} & \multirow{2}{*}{\cmark} & \multirow{2}{*}{\cmark} & \multirow{2}{*}{\cmark} & 5.6 & \multirow{2}{*}{\cmark} & \multirow{2}{*}{\cmark} & \multirow{2}{*}{\cmark} \\
    ~~~~~~~~~~~~~~~~~~~\textsc{Fulltask} & 86 &  &  &  &  &  & \textbf{22.7} &  & & \\
        \bottomrule
    \end{tabular}
    }
    \caption{\textbf{Comparison of \our~with existing GUI datasets.}
    \our~differs from existing benchmarks in:
    ($i$) sourcing from instructional videos with human demonstrations;
    ($ii$) featuring 86 challenging full tasks averaging 22.7 actions, and 463 subtasks;
    ($iii$) offering comprehensive evaluation with hierarchical planning and action categories.}
    \label{tab:benchmarks}
\end{table*}

\section{Related Works}
\textbf{Benchmarks GUI Tasks.}
In recent years, a range of works have focused on modeling GUI tasks and benchmarking agents, which include:
($i$) Web browsing~\cite{yao2022webshop}, where agents are developed to interact with web interfaces for navigation and to support a variety of tasks like online shopping.
($ii$) Mobile navigation~\cite{android_in_zoo}, aimed at improving accessibility within mobile GUI simulator environments, such as Android and iOS~\cite{androidenv}.
($iii$) Several efforts aimed at resolving issues with computer desktop software have emerged, such as grounding UI elements in offline settings like screenshots~\cite{seeclick}. Additionally, there has been development of executable simulated environments~\cite{GUIautonomous} for more interactive evaluation. AssistGUI~\cite{assistgui} is one project that utilizes video subtitles and metadata from instructional videos as reference, and evaluates the model by determining outcomes based on task success or failure.

Differing from these works, we focus on more complex and challenging GUI tasks that often require individuals to follow instructional videos to replicate long procedure operations and achieve goals. 
Specifically, We've developed a comprehensive evaluation framework that covers high-level task procedures, mid-level action decomposition, and atomic-level action execution. Our approach emphasizes UI visual-centric perception over textual understanding, focusing on identifying visual goals and transitions between states, which present significant challenges.

\textbf{Multi-Modal Agents.}
Recent studies have highlighted the promising potential of LLMs beyond language modeling. Notable advancements in Chain of Thought (CoT)~\cite{chain_of_thought}~and ReAct~\cite{react} strategies have demonstrated LLMs' capabilities as autonomous agents, capable of completing complex tasks through dynamic programming~\cite{mmreact, assistgpt}.
Motivated by these progresses, follow-up works connect LLMs with visual experts to enable multimodal applications, such as visual question answering~\cite{screenai}, or image editing applications~\cite{image_llm_edit}. 
In the realm of Embodied GUI tasks, the primary challenges involve understanding complex UI elements and planning to execute diverse tasks.
This has led to the development of approaches such as:
($i$) Training-free agent systems, which primarily consist of two stages: the first involves semantically understanding UI elements~\cite{actionbert,uibert,lexi}, either by transforming the GUI into HTML representations or language descriptions~\cite{empowering,gpt4vsom}, or using off-the-shelf visual models like OCR~\cite{pix2struct}. 
The second stage involves utilizing LLMs to integrate information and generate responses. 
This method heavily relies on closed-source LLMs, incurring significant costs. Additionally, it limits the model's UI visual perception abilities, such as demonstrating goals or state transitions visually rather than linguistically.
($ii$) Vision-Language-Action models~\cite{cogagent,you2024ferret}, which are pretrained on large-scale GUI vision-text corpus (\eg screenshots). This enables the LLMs to obtain more abilities such as element grounding and reasoning in unified responses~\cite{cogagent}. 
However, it remains unclear when and how to employ different types of GUI agents or tools. \our~provides a comprehensive suite for studying and benchmarking these models.



\section{VideoGUI Benchmarks}
\subsection{Data Construction}
\textbf{Data source.}
\our~consists of 11 software applications, categorized into:
(\textit{i}) media creation, featuring visual and animation tools like PowerPoint, Runway, and Stable Diffusion;
(\textit{ii}) media editing, including Adobe Photoshop, Premiere Pro, After Effects, CapCut, and DaVinci Resolve;
(\textit{iii}) media browsing, with platforms like YouTube, VLC Player, and Web Stock.

\begin{figure}[h]
	\centering
	\includegraphics[width=\linewidth]{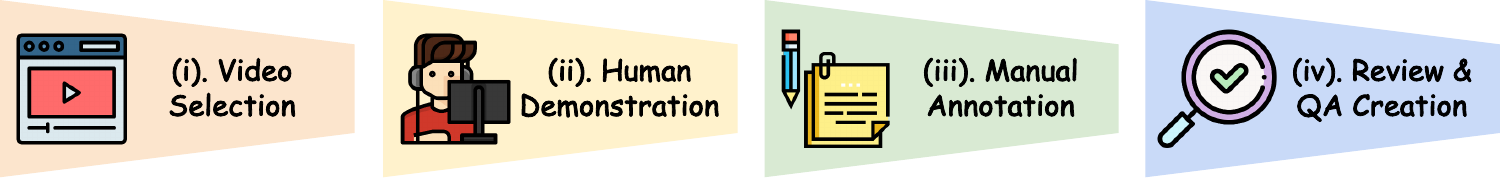}
\caption{
\textbf{Illustration of \our~creation pipeline}, encompassing four phases:
\textbf{(i)} High-quality instruction videos are manually selected,
\textbf{(ii)} Participants replicate skills demonstrated in videos,
\textbf{(iii)} Participants annotate task elements and procedures,
\textbf{(iv)} Annotated data is validated manually for \our~benchmarking use.
}
\label{fig:pipeline}
\end{figure}

\textbf{Pipeline.}
The \our~creation pipeline is illustrated in Fig.\ref{fig:pipeline}.
For each software, 
\textbf{(i)} we manually select instructional videos paired with high-quality  transcripts from YouTube, focusing on those teaching practical and novel usages. To collect the \textit{human manipulation trajectory}, we build a simulated environment to monitor user behaviors including \texttt{Click}, \texttt{Drag}, \texttt{Type/Press}, and \texttt{Scroll}.
\textbf{(ii)} We invite five participants who first watch the selected video and then try to reproduce the effects shown using our simulator, which records all cursor and keyboard activities (\eg~$[x,y]$ coordinates of a \texttt{RightClick}). Afterward, they provide a brief description of the overall goal 
(\ie~\textit{full-task's textual query}) and break down their operation sequence into several \textit{subtasks} and paired with descriptions, each focusing on a main functionality operation (\eg~inserting a figure).
\textbf{(iii)} We also instruct the annotators to identify the active elements (\eg~buttons `Insert') for each action, as they are not automatically identified and recorded by our simulator. 
After the demonstration, we retain all available files, including material, project files, and visual outcomes (the latter being our \textit{full-task's visual query}).
\textbf{(iv)} The participants cross-validate the annotations, and remove unclear/incorrect ones. 

\begin{figure}[h]
    \centering
    \begin{subfigure}[b]{0.3\linewidth}
        \includegraphics[width=0.9\linewidth]{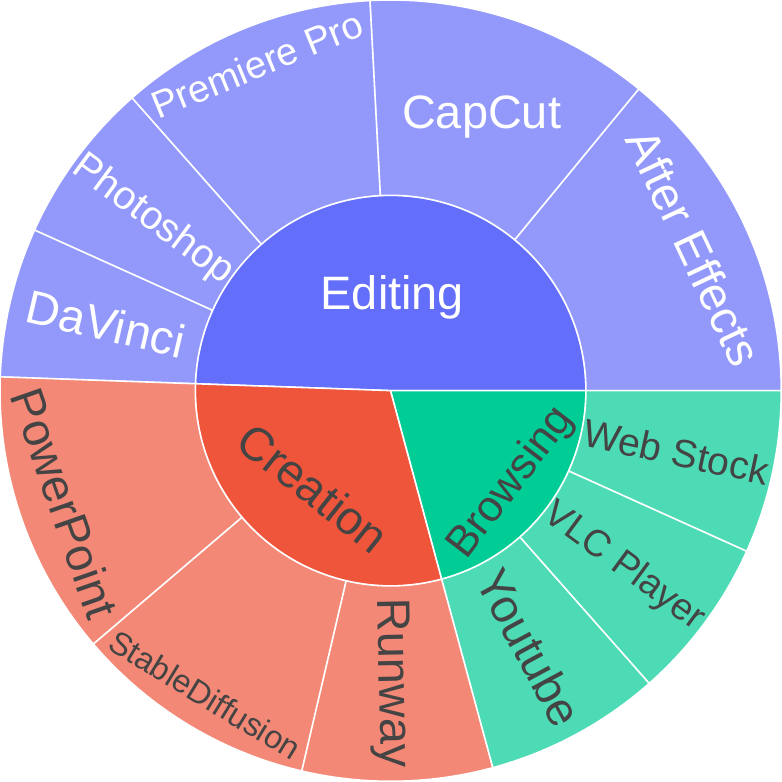}
        \caption{\textbf{Dist. of Software taxonomy.}}
        \label{fig:dist_app}
    \end{subfigure}
    \hfill
    \vspace{1em} 
    \begin{subfigure}[b]{0.3\linewidth}
        \includegraphics[width=\linewidth]{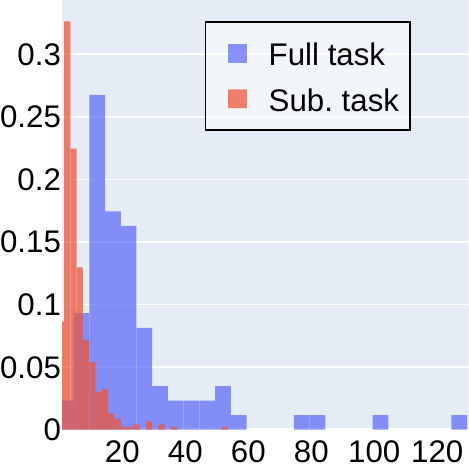}
        \caption{\textbf{Num. of Action per task.}}
        \label{fig:dist_act_num}
    \end{subfigure}
    \hfill
    \vspace{1.5em} 
    \begin{subfigure}[b]{0.32\linewidth}
        \includegraphics[width=1.0\linewidth]{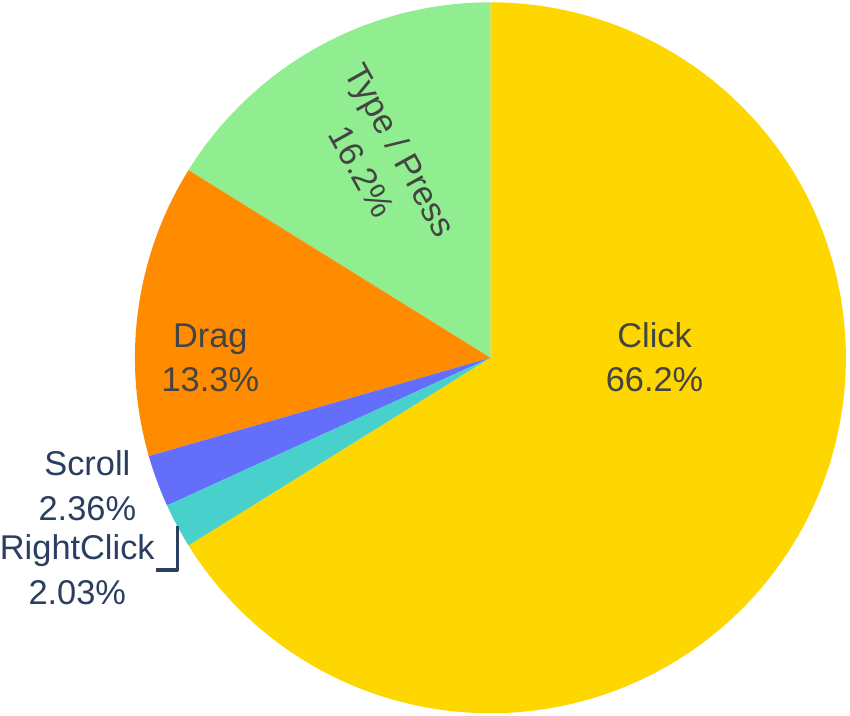}
        \caption{\textbf{Dist. of Atomic actions.}}
        \label{fig:dist_act}
    \end{subfigure}
    \vspace{-1.0em} 
    \caption{\textbf{Data statistics of \our.}
    }
    \label{fig:overall}
\end{figure}

\textbf{Data statistic.} 
Overall, \our~includes 178 tasks across 11 software applications (Fig.~\ref{fig:dist_app}) on Windows and Web browsers (Chrome, Edge, Firefox). It comprises 86 complex tasks (\ie~full task) and 92 simple tasks (\ie~subtask) that do not require high-level planning, where those 86 full tasks can be further divided into 371 subtasks, resulting in a total of 463 subtasks.
Fig.~\ref{fig:dist_act_num} shows the distribution of number of actions per task. In total, we collect 2,712 atomic manual actions. As shown in Fig.~\ref{fig:dist_act}, the most common action is LeftClick (66.2\%), while RightClick and Scroll are the least common actions (approximately 2\%).

\subsection{Evaluation and Metrics}
\label{sec:eval}
\textbf{Overview.} 
Imagine a human to complete the complex task illustrated in Tab.~\ref{fig:ps15}, we often first break down the full task into sub-tasks, and then sequentially perform the actions required to complete each subtask.
Existing GUI benchmarks~\cite{mind2web,pixelhelp,assistgui} predominantly use a boolean metric (i.e., Success Rate) to measure the success of completing a task. It may work okay for simpler tasks involving only a few actions, but is clearly not sufficient in providing feedback on where the models fall short, especially as the complexity of the task increases (\textit{e.g.}, a full task with over 100 actions), and nonetheless to say to guide future improvements in modeling for GUI navigation.   

To address this, we propose a \textit{hierarchical} decomposition of tasks into three key stages:
A. High-level Planning, which translates task instructions or reverse engineers final outcomes into several key milestones.
B. Middle-level Planning, which converts each milestone into detailed action narrations.
C. Atomic-level Execution, which focuses on accurately executing specific actions, such as clicking and typing, as dictated by the narration.
We discuss each part in detail subsequently.

\begin{table}[t]
    \centering
    \small
\begin{tabularx}{0.98\textwidth}{X|X|X|X}
        \toprule
        \multicolumn{4}{c}{\textbf{Visual preview}} \\
        \hline
        \multicolumn{4}{c}{\includegraphics[width=0.95\textwidth]{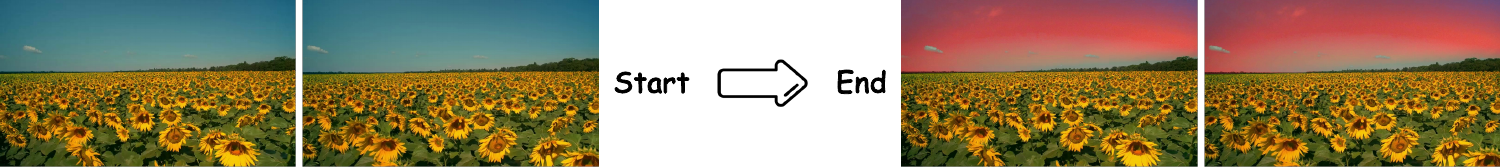}} \\
        \midrule
        \footnotesize{\textbf{Full task}} &
        \footnotesize{\dothh} &
        \footnotesize{\dotmm} & \footnotesize{\dotaa}\\  
        \hline
{\scriptsize
\textbf{Visual query:} How to transform from [start] to [end] in \pr?

\textcolor{gray}{\textbf{Textual query:} Change the blue sky in the background of the picture to the red color at sunset.}

}
&
{\scriptsize
\textcolor{darkyellow}{\textbf{a. Add ultra key effect to the video}}

\textbf{b.} Get the color of the background

\textbf{c.} Adjust the track order to the second track

\textbf{d.} Add the new background photo to the first video track;



}
&
{\scriptsize
\textcolor{darkgreen}{\textbf{a1. Click on Effects panel}}

\textbf{a2.} Click on search bar in Effects panel

\textbf{a3.} Key board Type `ultra'

\textbf{a4.} Click on `Ultra Key'

\textbf{a5.} Drag Ultry key effect from  effects panel to  the video. (Purpose:  add ultra key to the video)

}
&
{\scriptsize
\textcolor{darkblue}{a1. Click, [216, 996]}

\includegraphics[width=0.15\textwidth]{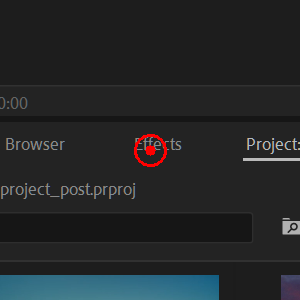}
}
\\
\bottomrule
\end{tabularx}
\vspace{0.2em}
    \caption{\textbf{\textit{Hierarchical} annotations in \our } 
    (\pr).
    The top row displays the video input and the desired task outcome as the visual query, with an optional textual query describing the video editing effect. The model is expected to "reverse-engineer" this outcome through a hierarchical process: first by planning high-level milestones (i.e., sub-tasks), then detailing each milestone into step-by-step narrations at the middle level, and finally translating these narrations into executable actions.
    }
    \vspace{-0.7cm}
    \label{fig:ps15}
\end{table}

\doth.
This method translates instructions or outcomes into key milestones (\textit{i.e.}, subtasks). Unlike previous approaches that start with explicit textual queries, practical scenarios often rely on final visual demonstrations like animations, requiring the reconstruction of past procedural tasks. Accordingly, we develop three categories based on different modal inputs:

\begin{itemize}[itemsep=-2pt, topsep=-5pt, leftmargin=*]
    \item \textit{Visual query} is our primary setting, with only a visual preview are provided. It could be a photo (produced by \ps), or a video transition (edited by \pr). 
    \item \textit{Textual query} explicitly defines the objectives using detailed descriptions.
    \item \textit{Visual query $+$ Textual query}, which provides the most complete information. 
\end{itemize}

\textbf{Metrics}: 
Planning involves open-ended question-answering with multiple correct approaches, making traditional metrics insufficient. To adaptively evaluate responses, we define a critic using \gptv~inspired by \cite{mmvet}, prompting LLM to focus on key elements and operations such as 3D shape and specific animation types. We score the model's generated procedure steps against the ground truth on a scale from $0$ (totally unrelated) to $5$ (almost correct).



\dotm.
Given a milestone task, the agent should perform appropriate UI operations based on its observation state (e.g., screenshots). This stage aims to generate a sequence of precise action narrations (\ie~desired action type with an accurate element) by combining textual milestones and visual observations. We devise three modes:

\begin{itemize}[itemsep=-2pt, topsep=-5pt, leftmargin=*]
\item \textit{Visual initial state $+$ Textual query:} Our main setting, as it accepts the output from the previous high-level planning, and the initial state (\ie~screenshot) can be straightforwardly obtained.
\item \textit{Textual query:} A common setting in most existing works.
\item \textit{Visual state transition (initial and end state):} the most challenging setting requiring the model to understand differences by screenshot transition and reverse its fine-grained actions.
\end{itemize}

\textbf{Metrics}: similar to the high-level planning phrase, we use the LLM as the critic for scoring.

\begin{figure}[t]
	\centering
	\includegraphics[width=0.95\linewidth]{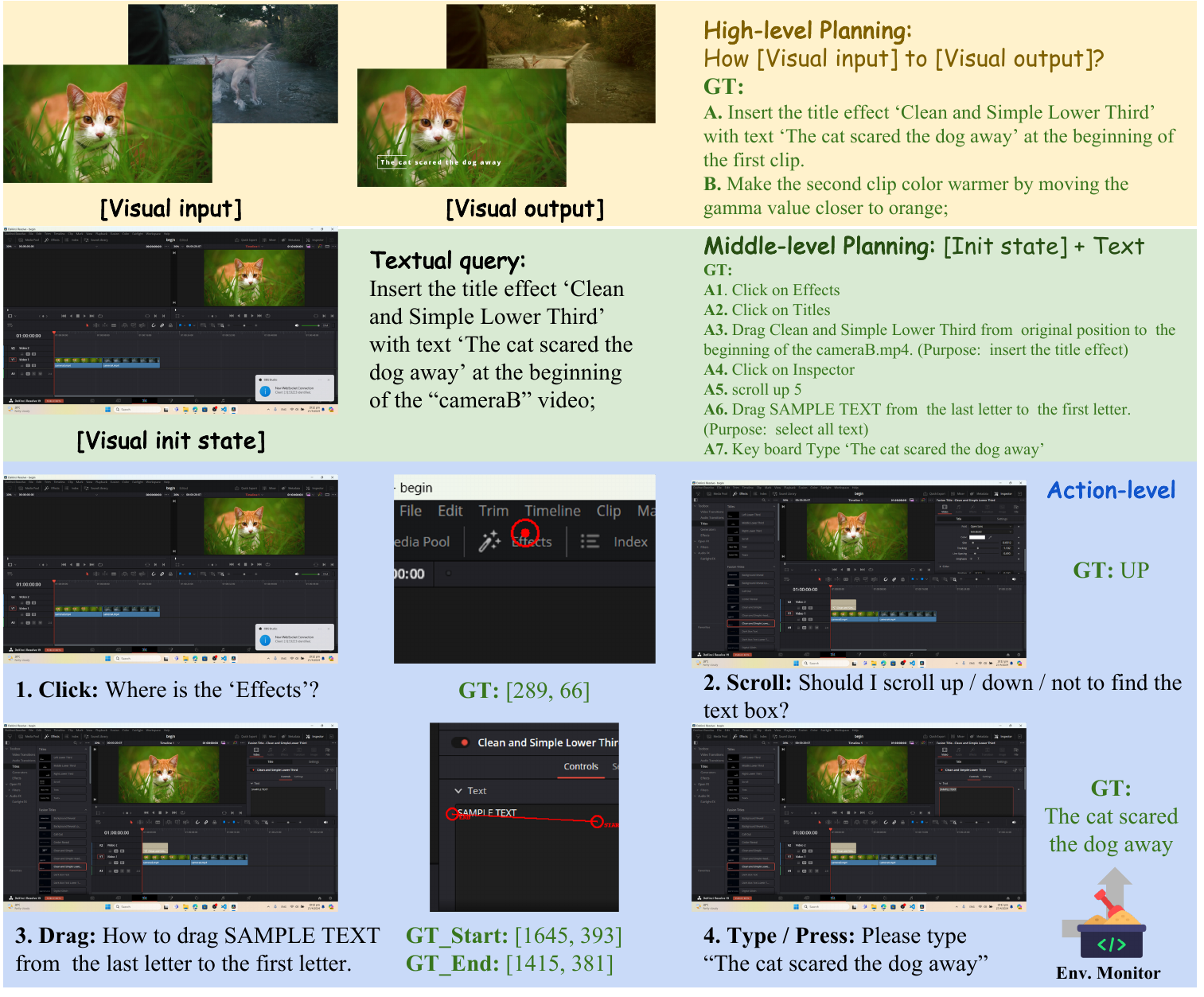}
    \caption{\textbf{Illustration of \our~hierarchical evaluation,} containing high-level planning (visual query), mid-level planning (visual state + text query), and action-level execution (with 4 types: click, scroll, drag, type/press). We use DaVinci software as an example.}
\label{fig:demo5}
\end{figure}

\dota.
After planning, the agent should respond to the action narrations with middle-level planning. We evaluate whether the model can accurately interpret narrations and perform the corresponding actions. Specifically, we consider the four most common action categories:

\begin{itemize}[itemsep=-2pt, topsep=-5pt, leftmargin=*]
    \item \texttt{Click:}  For a narration like \texttt{Click on the Insert button}, the model must accurately localize desired element on the screenshot by providing a bounding box or click position $[x,y]$. 

    \textbf{Metrics:} {($i$)} $\textup{Dist}\coloneqq\frac{\Delta}{L}$, where $\Delta$ is the pixel difference between the predicted location and the ground-truth coordinate.
    $L$ is the farthest distance from the ground-truth location to any of the four screenshot vertices for adaptive normalization.
    ($ii$) Recall@$d$: In practice, a click is usually valid when it falls within a very short distance, such as within a button area. We calculate the recall score with a threshold $d$, which we empirically set to 100.
    \item \texttt{Drag:} can be approximated as a combination of a Click with a short-phrase movement, where the model must infer both the initial and end coordinates. 

     \textbf{Metrics}: {($i$)} $\textup{Dist}\coloneqq\frac{1}{2}\left(\frac{\Delta s}{L s}+\frac{\Delta e}{L e}\right)$ by jointly measuring the start and end distance, then averaging them.
    ($ii$) Recall@$d$: This metric is more stricter than click, requiring both the start and end predictions to be within a distance $d$; otherwise, the score is 0.
    
    \item \texttt{Scroll:} The scroll action assesses whether the desired elements are visible in the current screenshot, determining if a scroll action is needed (e.g., scroll up, scroll down, or no need).
    
    \textbf{Metrics}: We frame this QA as a multiple-choice question: `Could you advise whether I need to scroll to see [target element]?' with options: [A] No need to scroll; [B] Scroll up; [C] Scroll down. To prevent bias, we shuffle the choices randomly and calculate the accuracy score.
    
    \item \texttt{Type \& Press:} For type actions such as \texttt{Type `Hello world!'}, the agent must accurately produce the string through keystrokes. For commands like \texttt{Ctrl+C}, it must execute multiple keystrokes and button presses. Most GUI agents utilize PyAutoGUI~\cite{pyautogui} for these operations, framing them as coding challenges that require verification for correctness.

    \textbf{Metrics}: We design a \textit{Sandbox} scheme by developing a mini simulator that executes the code produced by the agent. Additionally, we use a monitor to listen for the keys pressed or typed. We then compare the monitored results with the ground-truth results to check for matches. 
    This setting is evaluated using Recall (\ie~whether the GT is produced) and Precision (\ie~the count number of GT and actual outputs to study redundancy).
\end{itemize}
\section{Experiments}
\subsection{Baseline settings}
We evaluate leading Multi-modal Large Language Models (MLLMs) including \gptv, \gpto, \claude, \gemini, \qwen, and the open-source \cogagent. We also include text-only LLMs such as \chatgpt, \llama and \mixtral. 
Tab.~\ref{tab:mainres} summarizes all evaluated models and their supported modalities.

\subsection{Main Results on \our}
In Tab.~\ref{tab:mainres}, we provide a comprehensive evaluation of baseline models on \our. Scores are reported for high-level planning (visual query), middle-level planning (visual+text for MLLMs, or text only for LLMs), and atomic action (covering four categories), as well as an overall score summing these three.
The lowest scores in high-level planning across all models highlight the challenge posed by vision preview instructions. 
Overall, GPT-4o achieved the highest score of 39.4, excelling in all three tracks. 
We next explore procedural planning and action execution for deeper analysis.



\begin{table*}[h]

\tablestyle{7pt}{0.98} 
\centering
\resizebox{\textwidth}{!}
{
\begin{tabular}{l ccc llll}
\toprule
\multirow{2}{*}{\textbf{Model}} & \multicolumn{3}{c}{\textbf{Support Interleaved Instructions?}}& \multicolumn{4}{c}{\textbf{VideoGUI Evaluation (\%)}} \\
\cmidrule(lr){2-4} \cmidrule(lr){5-8}
& Text & Image (1f) & Media ($>1$f) & \cellcolor{high}{\textbf{High Plan}} & \cellcolor{middle}{\textbf{Mid. Plan}} & \cellcolor{action}{\textbf{Action}} & \textbf{Overall} \\
\midrule
\llama    & \cmark &                       &                       & \cellcolor{high}--  & \cellcolor{middle}40.5  & \cellcolor{action}20.3 & 20.3  \\
\mixtral & \cmark &                       &                       & \cellcolor{high}--  & \cellcolor{middle}36.0  & \cellcolor{action}19.6 & 18.6  \\
\chatgpt       & \cmark &                       &                       & \cellcolor{high}--  & \cellcolor{middle}49.1  & \cellcolor{action}22.3 & 23.8  \\
\cogagent      & \cmark & \cmark &  & \cellcolor{high}4.4  & \cellcolor{middle}21.8 & \cellcolor{action}7.4  & 11.2 \\
\qwen   & \cmark & \cmark & \cmark & \cellcolor{high}5.1  & \cellcolor{middle}35.7 & \cellcolor{action}28.9 & 23.2 \\
\gemini  & \cmark & \cmark & \cmark & \cellcolor{high}7.9  & \cellcolor{middle}28.6 & \cellcolor{action}23.8 & 20.1 \\
\claude & \cmark & \cmark & \cmark & \cellcolor{high}9.7  & \cellcolor{middle}45.6 & \cellcolor{action}39.4 & 31.6 \\
\gptv & \cmark & \cmark &    \cmark  & \cellcolor{high}14.3 & \cellcolor{middle}52.9 & \cellcolor{action}34.4 & 33.9 \\
\gpto        & \cmark & \cmark & \cmark & \cellcolor{high}\textbf{17.1} & \cellcolor{middle}\textbf{53.5} & \cellcolor{action}\textbf{47.6} & \textbf{39.4} \\
\midrule
GPT-4T + OCR & \cmark & \cmark & \cmark & \cellcolor{high}\textcolor{gray}{14.3} & \cellcolor{middle}\textcolor{gray}{52.9} & \cellcolor{action}{49.2} & {38.8} \\
GPT-4T + SoM~\cite{gpt4vsom} & \cmark & \cmark & \cmark  & \cellcolor{high}\textcolor{gray}{14.3} & \cellcolor{middle}\textcolor{gray}{52.9} & \cellcolor{action}44.2 & 37.1 \\
GPT-4o + OCR & \cmark & \cmark & \cmark & \cellcolor{high}\textcolor{gray}{17.1} & \cellcolor{middle}\textcolor{gray}{53.5} & \cellcolor{action}{\textbf{56.3}} & \textbf{42.3} \\
GPT-4o + SoM~\cite{gpt4vsom} & \cmark & \cmark & \cmark & \cellcolor{high}\textcolor{gray}{17.1} & \cellcolor{middle}\textcolor{gray}{53.5} & \cellcolor{action}{54.3} & {41.6} \\
\bottomrule
\end{tabular}
}
\caption{\textbf{Full evaluation on \our~with Baselines and their supported \textit{interleaved} instructions}, which might be a text query, an image (1 frame), or a media (more than 1 frame) such as two photos, one or two videos.}
\label{tab:mainres}
\end{table*}

\begin{table*}[h]
\small
\tablestyle{11pt}{0.98}
\centering
\resizebox{\textwidth}{!}
{
\begin{tabular}{l ccc ccc}
\toprule
\multirow{2}{*}{\textbf{Model}}  & \multicolumn{3}{c}{\textbf{High-level Planning} $\left(0-5\right)$} & \multicolumn{3}{c}{\textbf{Middle-level Planning} $\left(0-5\right)$} \\
\cmidrule(lr){2-4} \cmidrule(lr){5-7}  
&  \cellcolor{high}\textbf{Vision} & {Text} & {Vision \& Text} & {Vision} & {Text} & \cellcolor{middle}\textbf{Vision \& Text} \\
\midrule
\llama~ & \cellcolor{high}-- & {2.62} & -- & -- & {2.02} & \cellcolor{middle}-- \\
\mixtral & \cellcolor{high}-- & {2.43} & -- & -- & {1.80} & \cellcolor{middle}-- \\
\chatgpt & \cellcolor{high}-- & {2.67} & -- & --& {{2.46}} & \cellcolor{middle}-- \\
CogAgent~\cite{cogagent}  & \cellcolor{high}0.22 & {1.12} & {1.23} & -- & 1.32 & \cellcolor{middle} 1.09\\
Qwen-VL-Max~\cite{Qwen_technicalReport}  & \cellcolor{high}0.25 & {2.30} & {1.96} & 0.70 & 1.72 & \cellcolor{middle} 1.79 \\
Gemini-Pro-Vision~\cite{gemini} & \cellcolor{high}0.39 & {2.35} & {1.45} & {0.34} & 1.61 & \cellcolor{middle} 1.43 \\
Claude-3-Opus~\cite{claude} & \cellcolor{high}0.48 & {2.54} & {2.17} & 0.66 & 2.26 & \cellcolor{middle}2.28 \\
\gptv & \cellcolor{high}0.71 & {2.57} & {\textbf{2.55}} & {1.49} & \textbf{2.57} & \cellcolor{middle}2.65 \\
\gpto & \cellcolor{high}\textbf{0.86} & {\textbf{2.68}} & {2.46} & {\textbf{1.78}} & 2.45 & \cellcolor{middle}\textbf{2.68} \\
\cmidrule{1-7}
\textbf{Avg. by models} & \cellcolor{high}0.49 & {2.37} & {1.97} & {0.99} & 2.02 & \cellcolor{middle}1.98 \\
\bottomrule
\end{tabular}
} 
\caption{\textbf{Detailed evaluation on Procedural Planning}, including high-level and middle-level. 
Each level will be studied across three kinds of query formulation discussed in $\S$~\ref{sec:eval} (e.g., vision, text, and vision \& text). Colors are used to highlight the primary setting under each track. The maximum score is 5.}
\label{tab:plan}
\end{table*}

\textbf{Procedural planning.}
Tab.~\ref{tab:plan} studies the impact of different query formulations for planning. 
On both high and middle-level:
\textbf{(i)} The vision-only setting is significantly challenging (especially for high-level, {0.49} versus {2.37} for textual).
Among the models, GPT-4o demonstrates the strongest visual reasoning ability.
\textbf{(ii)} All models, except \cogagent with a small LLM~\cite{vicuna2023}, exhibit similar performance on textual-only inputs, as the textual query concretely indicates the key operations or effects type.
 This suggests that if we have clear and detailed textual instructions, a text LLM may be sufficient for this stage.
\textbf{(iii)} We do not observe a significant gain in the vision+text setting compared to text-only, which requires strong interleaved UI perception abilities.

\begin{table*}[h]
\tablestyle{4pt}{0.98} 
\centering
\resizebox{\textwidth}{!}
{
\begin{tabular}{l l ll ll ll l l}
\toprule
\multirow{2}{*}{\textbf{Model}}  & \multirow{2}{*}{\textbf{Grd.?}}  &\multicolumn{2}{c}{\textbf{1. Click}} & \multicolumn{2}{c}{\textbf{2. Drag}} & \multicolumn{2}{c}{\textbf{3. Type / Press}} & \multicolumn{1}{c}{\textbf{4. Scroll}} & {\cellcolor{action}\textbf{Action}}{\cellcolor{action}} \\
\cmidrule(lr){3-4} \cmidrule(lr){5-6} \cmidrule(lr){7-8}
& & Dist. $\downarrow$ & {Recall} $\uparrow$ 
& Dist. $\downarrow$ & {Recall} $\uparrow$ 
& Recall & Prec. & Acc. & \cellcolor{action}{Full} \\
\midrule
\color{gray}{Random} &\textcolor{gray}{--} & \textcolor{gray}{49.9} & \textcolor{gray}{0.7}  & \textcolor{gray}{47.2} & \textcolor{gray}{0.0} &\textcolor{gray}{--} & \textcolor{gray}{--} & \textcolor{gray}{31.3} & \cellcolor{action}\textcolor{gray}{8.0}\\
\midrule
\multicolumn{7}{c}{\textit{LLMs}} \\
LLama3-70B~\cite{llama3}        &    --  &    --   & --  &    --    & -- & {84.9} & {81.3} &  -- & \cellcolor{action}20.3 \\
Mixtral-8x22B~\cite{mixtral}     &   --  &    --  & --  &    --    & --  & {82.6} & {78.5} & --  & \cellcolor{action}19.6 \\
\chatgpt         &    --  &    --  & --  &    --    & -- & \textbf{93.1} & \textbf{89.5} & --  & \cellcolor{action}22.4 \\
\midrule
\multicolumn{8}{c}{\textit{Multi-modal LLMs}} \\
CogAgent~\cite{cogagent}   &    \cmark        & 30.9 & 3.4  & 44.7  & 0.0 & --                      & --                      & 26.6 & \cellcolor{action}7.5  \\
Qwen-VL-Max~\cite{Qwen_technicalReport}  &    \cmark     & 46.8 & 0.0  & 42.0 & 0.3 & 84.3                     & 73.0                     & 42.2 & \cellcolor{action}28.9 \\
Gemini-Pro-Vision~\cite{gemini} &  & 40.7 & 5.0  & 40.8 & 0.0 & 86.4                     & 82.2                     & 7.8 & \cellcolor{action}23.8 \\
Claude-3-Opus~\cite{claude}  &   & 30.7 & 7.0  & 30.6 & 1.7 & 92.5                     & 88.1                     & 60.9 & \cellcolor{action}39.4 \\
GPT-4-Turbo~\cite{gpt4report}  &            & 23.8 & 10.0 & 31.3 & 1.4 & 92.3                     & 88.8                     & 37.5 & \cellcolor{action}34.4 \\
GPT-4o~\cite{gpt4report}  &            & 16.6 & 17.7 & 21.9 & 2.5 & 92.3                     & 89.0                     & {81.3} & \cellcolor{action}47.6 \\
\midrule
\multicolumn{7}{c}{\textit{Modular methods: LLMs + Tools}} \\
GPT-3.5~$+$ OCR~\cite{chatgpt}  &    \cmark    & 16.8 & 48.7 & 36.4    & 5.5 & \textcolor{gray}{93.1}                     & \textcolor{gray}{89.5}                     & 56.3 & \cellcolor{action}50.0 \\
GPT-4T~~$+$ OCR~\cite{chatgpt}  &    \cmark & 14.8 & 55.1 & 26.6 & 12.2 & \textcolor{gray}{92.3} & \textcolor{gray}{88.8} &  40.6 & \cellcolor{action}{49.2}\\
GPT-4o~~$+$ OCR~\cite{chatgpt}  &    \cmark    & \textbf{12.0} & \textbf{60.1}~$_{(+42.4)}$ & 25.7     & \textbf{11.3}~$_{(+8.8)}$ & \textcolor{gray}{92.3}                     & \textcolor{gray}{89.0}                     & {82.8}~$_{(+1.5)}$ & \cellcolor{action}\textbf{56.3}~$_{(+8.7)}$ \\


GPT-4T~~$+$ SoM~\cite{chatgpt}  &  \cmark  & 19.1 & 30.6 & 25.3 & 4.1 & \textcolor{gray}{92.3} & \textcolor{gray}{88.8} & 53.1 & \cellcolor{action}{44.2} \\
GPT-4o~~$+$ SoM~\cite{gpt4vsom}   &  \cmark     & 15.7 & 35.9~$_{(+18.2)}$ & \textbf{22.9} & {3.0}~$_{(+0.5)}$ & \textcolor{gray}{92.3}                     & \textcolor{gray}{89.0}                     & \textbf{89.0}~$_{(+7.7)}$ & \cellcolor{action}{54.3}~$_{(+6.7)}$ \\
\bottomrule
\end{tabular}
}
\caption{\textbf{Detailed evaluation on Actions Executions.}
We study four types of atomic action execution, where the full score is the sum of Click recall, Drag recall, Type precision, and Scroll accuracy.
\textbf{Grd.} indicates whether the model has emplicit grounding ability such as output element's coordinates.
In the bottom half, we equip LLMs with tools like SoM~\cite{gpt4vsom} and OCR~\cite{azureocr}.
}
\label{tab:act}
\end{table*}

\textbf{Action executions.}
Tab.~\ref{tab:act} examines the impact of different atomic actions on model performance. 
We summarize our findings as below.
\textbf{(i) Click:} 
We prompt multi-modal LLMs to output coordinates by providing screenshots with its {resolutions}, and we found that they can have a proper estimation, with meaningful improvement over random score but with poor recall.
Notably, closed-source LLMs demonstrate better grounding abilities than grounding-based models such as CogAgent; 
Enhancing LLMs with tools such as OCR~\cite{azureocr} or SoM~\cite{gpt4vsom} significantly improves model performance. Notably, for the text-based GPT-3.5 with OCR, it achieves a 48.7 recall.
\textbf{(ii) Drag:} To perform Drag, it requires models to accurately localize the movement at both the start and end points. The best model, GPT4-o with OCR, yields only 11.3 recall.
For LLMs with tools, OCR brings 8.8 recall gain over the base model, which is even more helpful than SoM as it helps to precisely localize text for the button, while SoM often suffers from poor segmentation results.
\textbf{(iii) Type / Press:}
Regarding keyboard activity, most models achieve good scores, as large-scale instruction-tuned LLMs generally is decent at coding, making the LLMs even more competent for this task. 
\textbf{(iv) Scroll:} 
For Scroll, models must infer not only whether an element appears but also its order relative to other elements.
GPT-4o is the top-performing model, while Gemini scores extremely low, often preferring outputs without scrolling.
 

\begin{wrapfigure}{r}{0.26\textwidth}
    \vspace{-1em}
    \begin{minipage}{0.26\textwidth}
        \centering
        \includegraphics[width=\textwidth, clip, trim={1mm 1mm 0cm 0cm}]{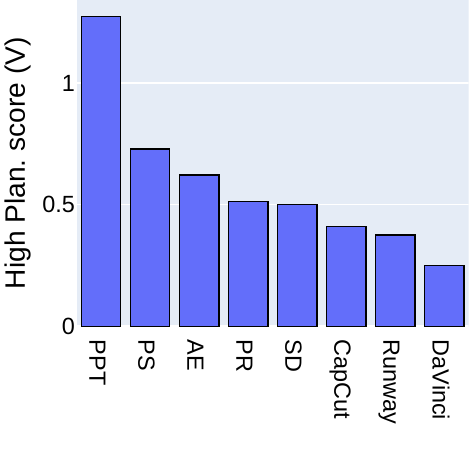}
        \label{fig:plan_by_app}
    \end{minipage}
    
    
    \begin{minipage}{0.26\textwidth}
        \centering
        \includegraphics[width=\textwidth, clip, trim={1mm 1mm 0cm 0cm}]{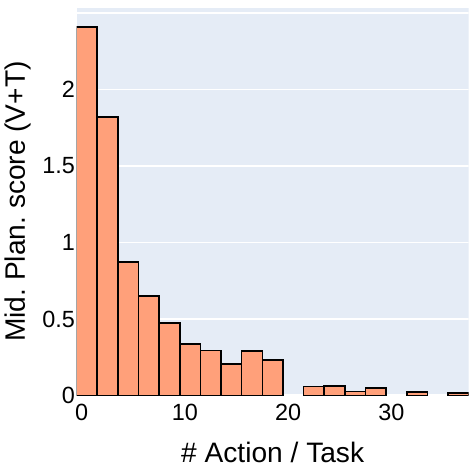}
        \label{fig:plan_by_act_num}
    \end{minipage}
    \vspace{-10pt}
    \caption{\textbf{Top:} High plan. score (V) by diff. software; \textbf{Bottom:} Mid. plan score (V+T) by action number. 
    }
    \label{fig:plan_performance}
    \vspace{-3em}
\end{wrapfigure}

\subsection{Performance by Task Difficulty.}

\textbf{High planning by different software.} 
Fig.~\ref{fig:plan_performance} (top) shows mid-level plan scores (visual query) across different software. Models perform highest on \ppt, which is more commonly used than others.
On \ae~and \ps, model performance drops significantly as they are professional software. It is worth mentioning that being web-based, Runway and Stable Diffusion remains challenging because these novel applications are relatively new to the MLLMs.

\textbf{Middle planning by action number.} 
Fig.~\ref{fig:plan_performance} (bottom) shows the mid-level planning scores (visual + text query) by the number of actions per task. Scores tend to decrease as the number of actions increases,
demonstrating the difficulty of long procedural GUI tasks.



\definecolor{mygreen}{RGB}{34, 139, 34}

\begin{figure}[!ht]
    \centering
    \begin{subfigure}{1\textwidth}
        \centering
        \includegraphics[width=\textwidth, clip, trim={0cm 1mm 0cm 0cm}]{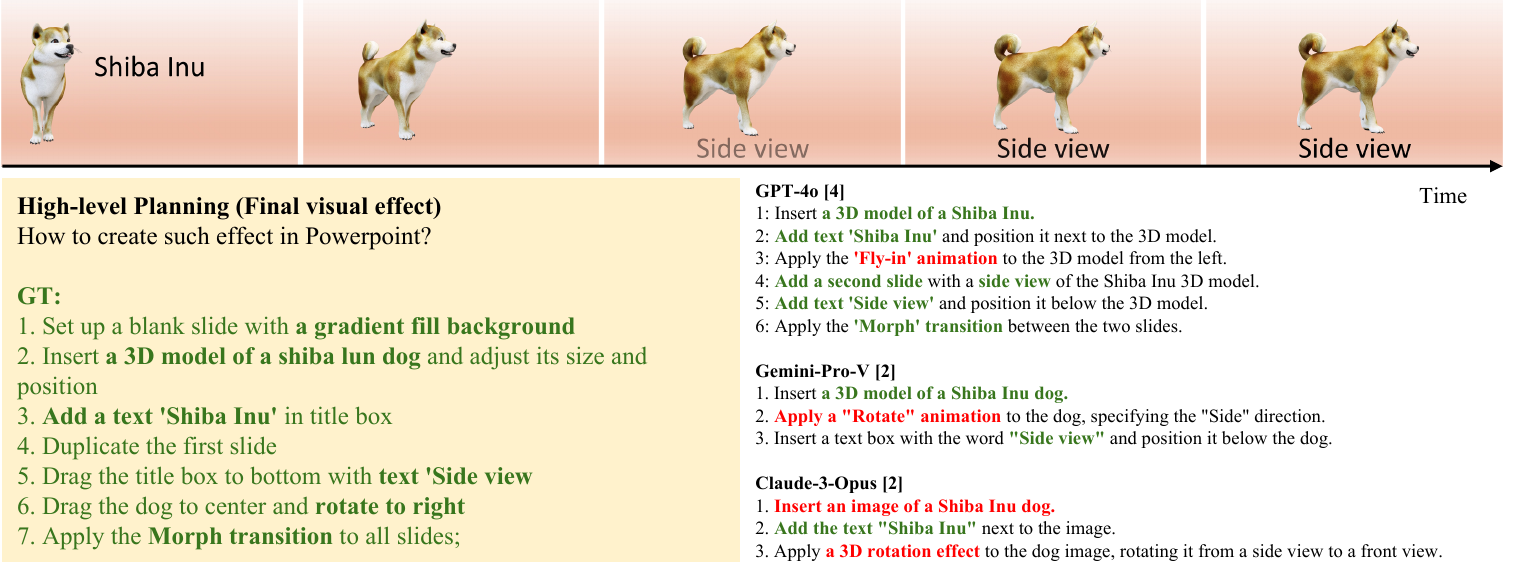}
        
        \caption{\textbf{High-level Planning}}
        \vspace{15pt}
    \end{subfigure}
    \hfill
    \begin{subfigure}{1\textwidth}
        \centering
        \includegraphics[width=1\textwidth, clip, trim={0cm 3mm 0cm 0cm}]{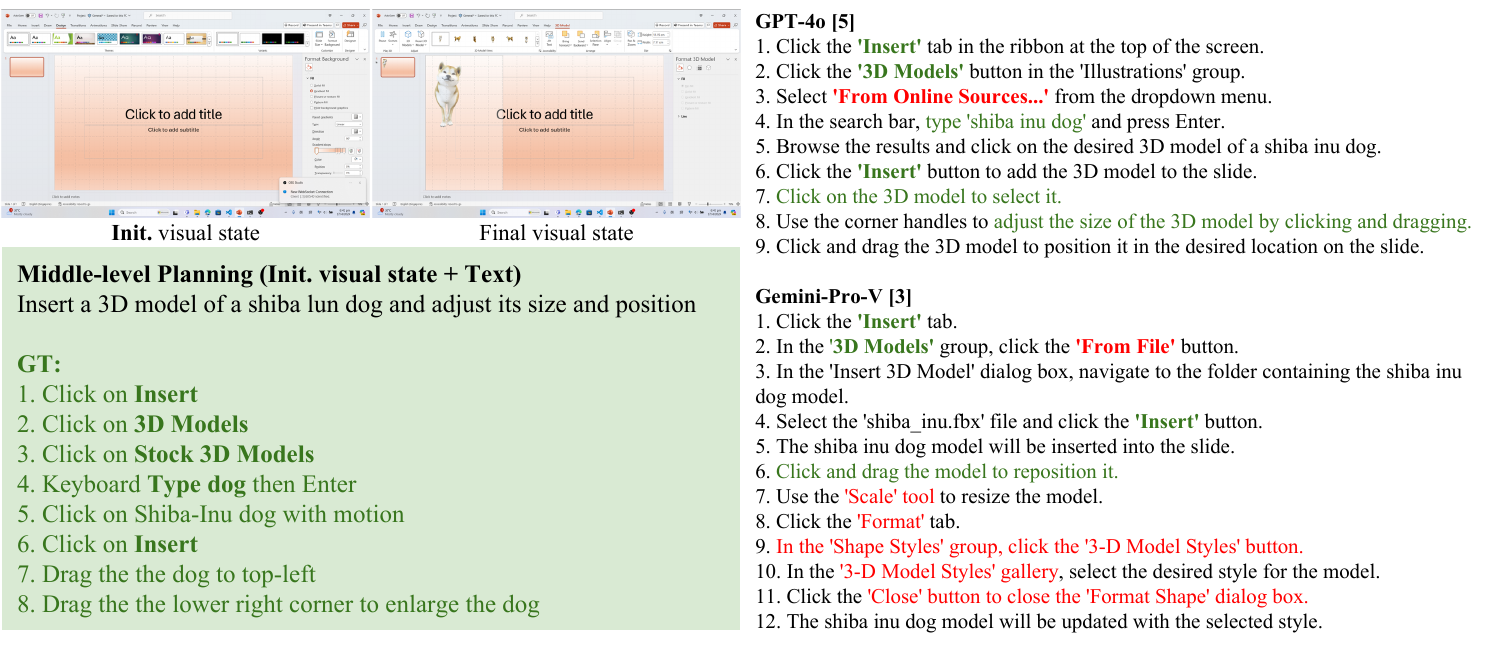}
        \caption{\textbf{Middle-level Planning}}
        \vspace{15pt}
    \end{subfigure}
    \hfill
    \begin{subfigure}{1\textwidth}
        \centering
        \includegraphics[width=1\textwidth, clip, trim={0cm 2mm 0cm 0cm}]{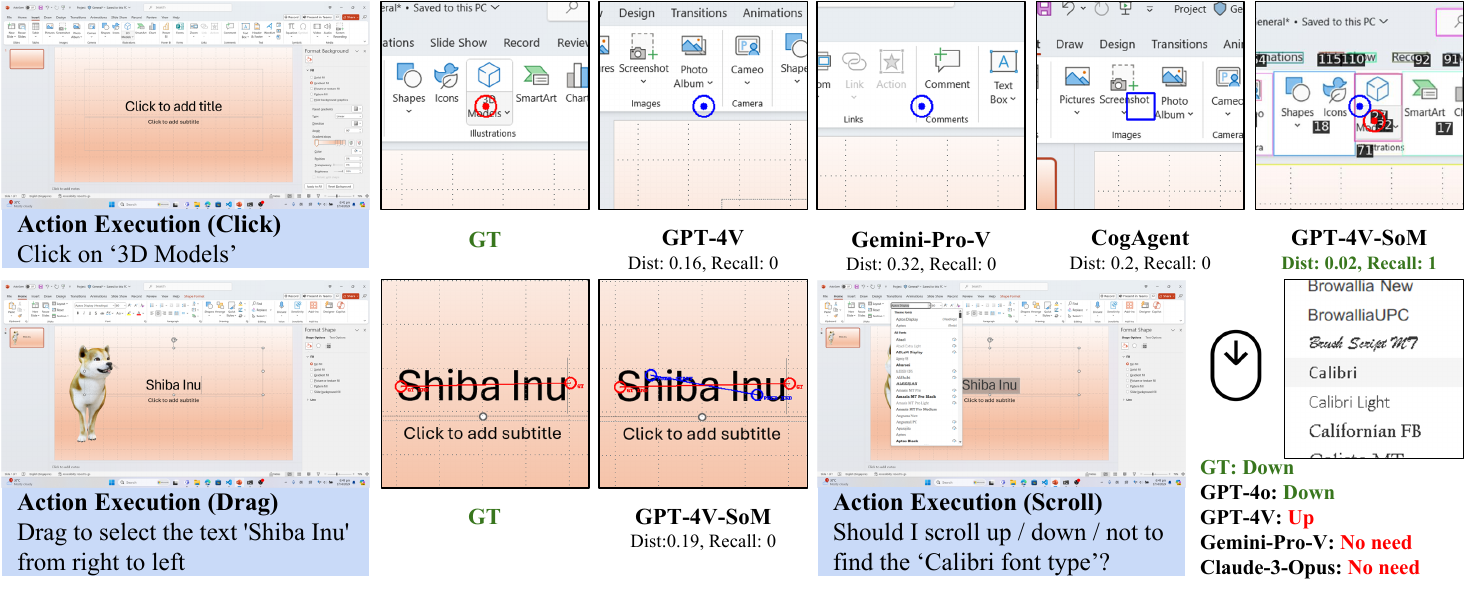}
        \caption{\textbf{Atomic Execution} with Click, Drag and Scroll.}
        \vspace{15pt}
    \end{subfigure}
    \caption{\textbf{Qualitative Results on \our}, which perform top-down evaluation across high-level, middle-level and atomic execution. 
    The color \textcolor{mygreen}{\textbf{green}} indicates the human reference (GT), while \textcolor{red}{\textbf{red}} indicates model's mistakes.
    The example is from Powerpoint software.}
    \label{fig:example}
\end{figure}

\subsection{Qualitative Results}
In Fig.~\ref{fig:example}, we visualize model performance and failure cases.
In (a) High-level planning, GPT-4o and Gemini-Pro-V successfully predict the sub-tasks for the slide with the 3D model. GPT-4o also accurately identifies the Morph transition effect, achieving the best score.
In (b) Mid-level planning, both models inserted and adjusted the 3D Shiba Inu model.
However, Gemini-Pro-V introduces unnecessary operations, such as shape styles and formatting, leading to discrepancies in positioning and scaling.
In (c) Atomic execution, models are assessed on precise actions. In Drag, GPT-4V selects part of the letters, but as the pixel distance is larger than the threshold, it still receives a recall of 0.
\section{Conclusion}
\vspace{-0.2cm}

In this work, we introduced \our, a multi-modal benchmark for advanced GUI tasks sourced from high-quality instructional videos targeting professional and novel software. \our, with its long procedural tasks, hierarchical manual annotations, and well-established evaluation metrics, provides clear signals for existing limitations and areas for improvement.
By comparing state-of-the-art models, we highlight the challenges of visual-oriented GUI automation and the potential of instructional videos for advancing GUI task automation.

\clearpage
\appendix
\section{Experimental Settings}
\subsection{Data Collection Settings}
We use OBS Studio~\cite{obs} software to record the demonstration videos and capture the user's screenshots. Notably, in the screenshots, the user's cursor is not recorded, which is beneficial as the screenshots can be used directly without revealing the target coordinates. 
We use \texttt{pynput} to monitor detailed user activity metadata, such as click location $[x,y]$, typed content, and scroll distance.

\begin{figure}[h]
\centering
\includegraphics[width=\linewidth]{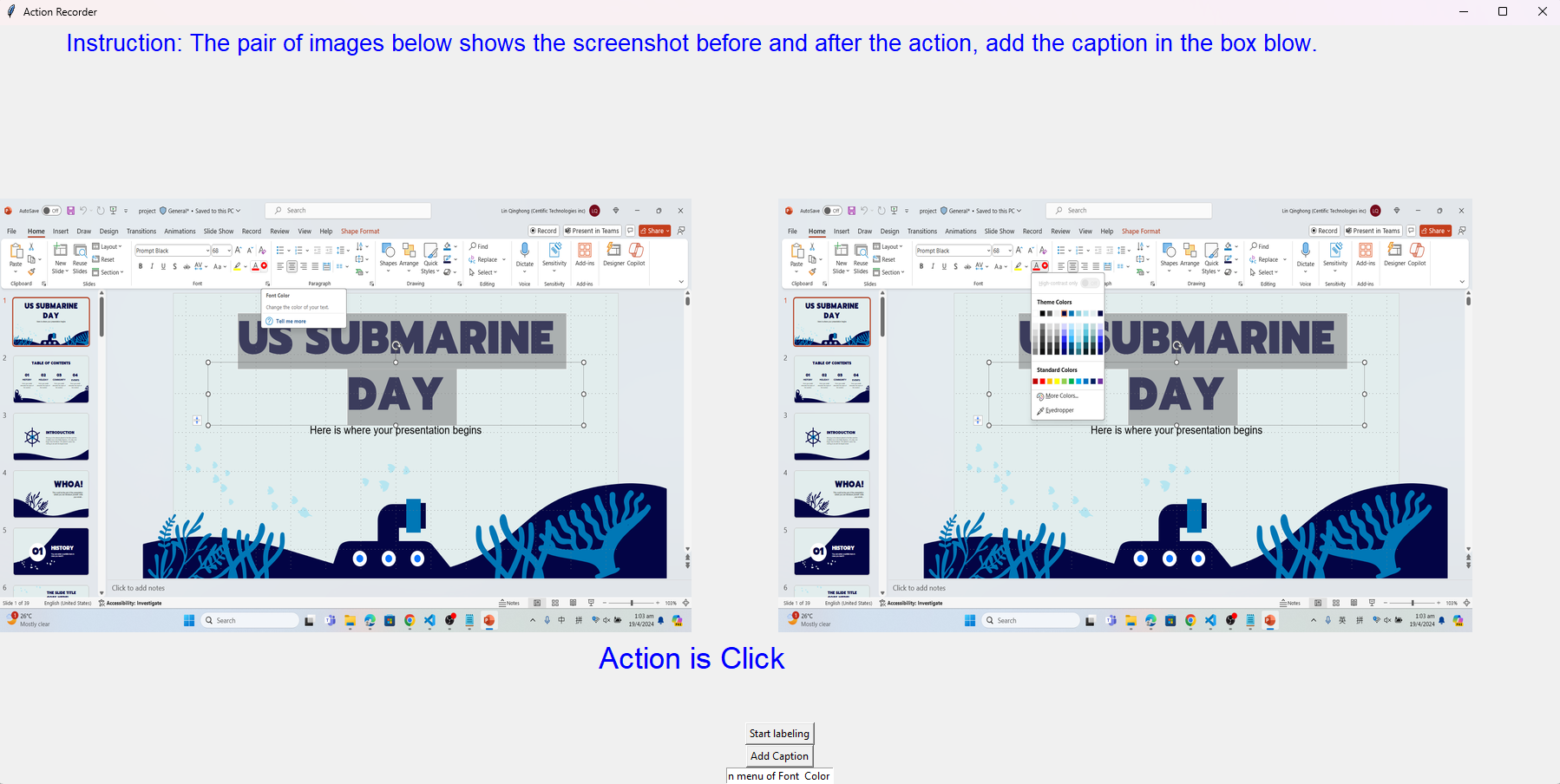}
\caption{Illustration of Manual annotation tools. The user are asked to watch their keyframe in their recording, and prompt to provide the element name regarding action.}
\label{supp:label_gui}
\end{figure}

In Fig.~\ref{supp:label_gui}, we display our manually labeled interface. Here, the annotator watches their key recording screenshots, with active regions such as the cursor coordinates highlighted in red. The annotators are then asked to enter the element name (e.g., "Drop-down menu of font color").

\subsection{Baseline Details}

\begin{tabular}{lll}
\toprule
Model & Ref. link & Version (\eg~model id)  \\
\midrule 
\llama & \href{https://deepinfra.com/}{\texttt{deepinfra}} & \texttt{meta-llama/Meta-Llama-3-70B-Instruct}  \\
\mixtral & \href{https://deepinfra.com/}{\texttt{deepinfra}} & \texttt{mistralai/Mixtral-8x22B-Instruct-v0.1} \\
\chatgpt & \href{https://platform.openai.com/docs/models/gpt-3-5-turbo}{\texttt{OpenAI}} & \texttt{gpt-3.5-turbo}  \\
\cogagent & \href{https://github.com/THUDM/CogVLM}{\texttt{CogAgent}} & \texttt{CogAgent-18B}  \\
\qwen & \href{https://help.aliyun.com/zh/dashscope/developer-reference/tongyi-qianwen-vl-plus-api?spm=a2c4g.11186623.0.0.645b7794Zi8mEy}{\texttt{Aliyun}} & \texttt{qwen-vl-max}  \\
\claude & \href{https://docs.anthropic.com/en/docs/models-overview}{\texttt{Anthropic}} & \texttt{claude-3-opus-20240229}  \\
\gemini & \href{https://deepmind.google/technologies/gemini/pro/}{\texttt{Google}} & \texttt{gemini-pro-vision}  \\
\gptv & \href{https://platform.openai.com/docs/models/gpt-4-turbo-and-gpt-4}{\texttt{OpenAI}} & \texttt{gpt-4-turbo}  \\
\gpto & \href{https://platform.openai.com/docs/models/gpt-4o}{\texttt{OpenAI}} & \texttt{gpt-4o}  \\
\bottomrule
\end{tabular}

\subsection{Evaluation Settings}
\textbf{Click.}
We detail how we calculate the distance metric. Assume we have a ground-truth point \([x_o, y_o]\) while the screenshot size is $H\times W$.

\begin{figure}[!h]
  \centering
  \begin{minipage}[b]{0.48\textwidth}
    \centering
    \includegraphics[width=\textwidth]{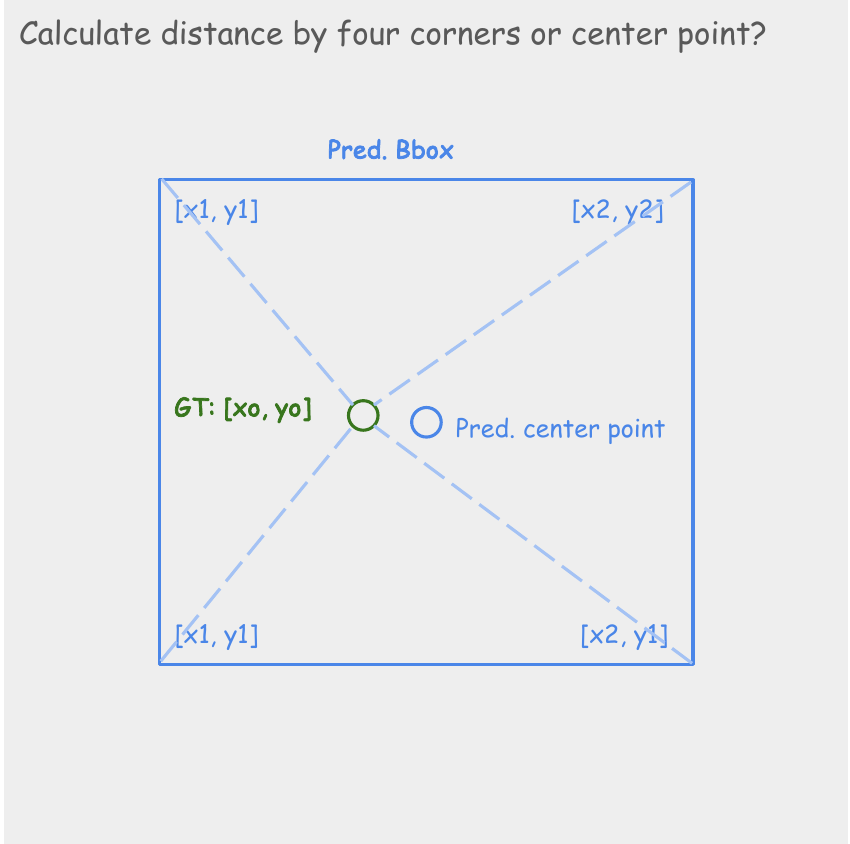}
    \label{fig:image1}
  \end{minipage}
  \hfill
  \begin{minipage}[b]{0.48\textwidth}
    \centering
    \includegraphics[width=\textwidth]{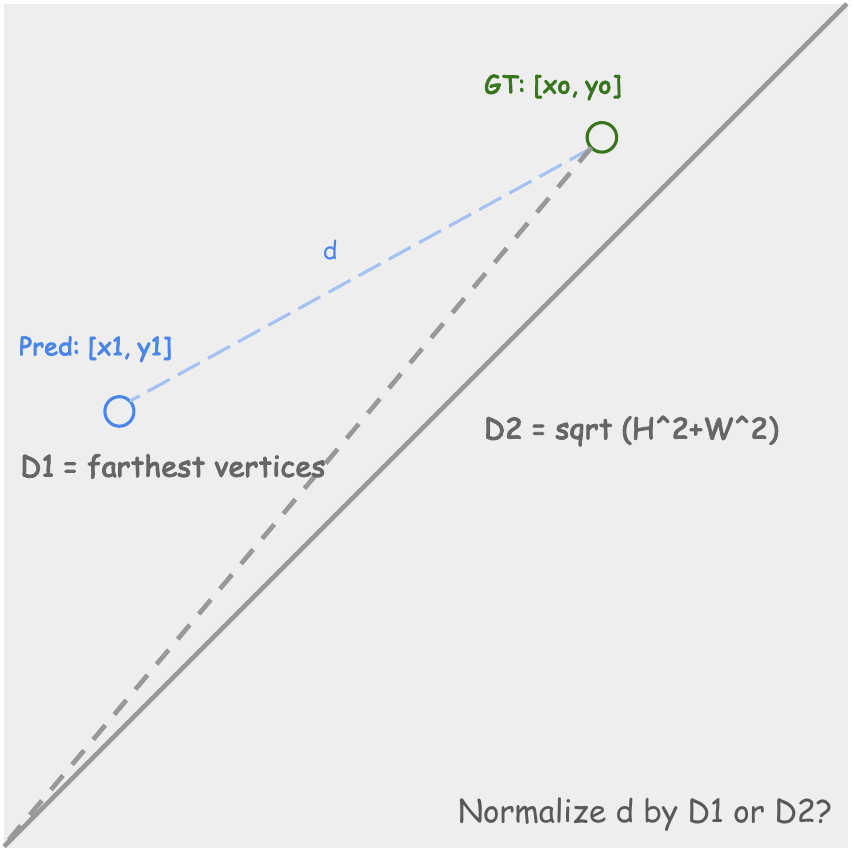}
    \label{fig:image2}
  \end{minipage}
  \caption{
  \textbf{(a) Illustration of why taking the distance btween the center point of a bounding box and groundtruth is not a proper measure of model performance on click.} As shown, the predicted bounding box center point is quite close to the ground-truth point, but the predicted bounding box area is large.
  \textbf{(b) Illustration of distance normalization.} To normalize the distance $d$ to $0 - 1$, a more proper term should be $D1$ (farthest vertices) rather than $D2$.
  }
  \label{fig:click:metric}
\end{figure}

$\bullet$ If the model prediction is a bounding box \([x_1, y_1, x_2, y_2]\) (\eg CogAgent \cite{cogagent} or Qwen-VL-Max \cite{Qwen_technicalReport}):

We cannot only take the center of the bounding box as the click target for evaluation because it does not account for the area of the bounding box. 
As illustrated in Fig.~\ref{fig:click:metric} (a), if the center point is very close to the ground truth but the bounding box cover a large area, the distance between the center point and the groundtruth would be small. 
Therefore, we design our metric to penalize for the area of the bounding box. Specifically, we calculate the distance between the ground truth and the four corners of the bounding box and then take the average. 
For the predicted bounding box, the average distance $d$ is calculated as follows: 
\begin{align*}
d &= \frac{1}{4} \left( \sqrt{(x_o - x_1)^2 + (y_o - y_1)^2} + \sqrt{(x_o - x_1)^2 + (y_o - y_2)^2} \right. \\
&\quad \left. + \sqrt{(x_o - x_2)^2 + (y_o - y_1)^2} + \sqrt{(x_o - x_2)^2 + (y_o - y_2)^2} \right)
\end{align*}
$\bullet$ If the model prediction is a coordinate \([x_1, y_1]\) (\eg as in GPT4V+SoM \cite{gpt4vsom}): 

We directly adopt the distance \(d\) calculated by:
    \[
    d = \sqrt{(x_o - x_1)^2 + (y_o - y_1)^2}
    \]

To normalize the pixel-level distance $d$ to $0-1$, a simple way is to divide $d$ by the maximum length in the screenshot, such as $\sqrt{H^2+W^2}$. But in practice, the maximum length should be the distance between the ground-truth point and the farthest vertices, so we use that for normalization. The comparison between the two normalization methods is illustrated in Fig.~\ref{fig:click:metric} (b).

\textbf{Drag.}
Drag is a combination of Clicks, so we simply adopt the click metric for the start and end point of drag, and take the average. The score is calculated as
$\textup{Dist}\coloneqq\frac{1}{2}\left(\frac{d_s}{D_s}+\frac{d_e}{D_e}\right)$ 
where $d_s$ is the pixel difference between predict start and GT start, while $D_s$ is the farthest vertices for the GT start; $d_e$ is the pixel difference between predict end and GT end, while $D_e$ is the farthest vertices for the GT end;

For Recall, it is calculated by:
\[ \textup{Recall}\left(\textup{start, end}\right) = 
\begin{cases} 
1 & {if}\quad\textup{Recall}\left(\textup{start}\right) \&~ \textup{Recall}\left(\textup{end}\right) \\ 
0 & {otherwise}
\end{cases}
\]

\begin{figure}[h]
\centering
\includegraphics[width=\linewidth]{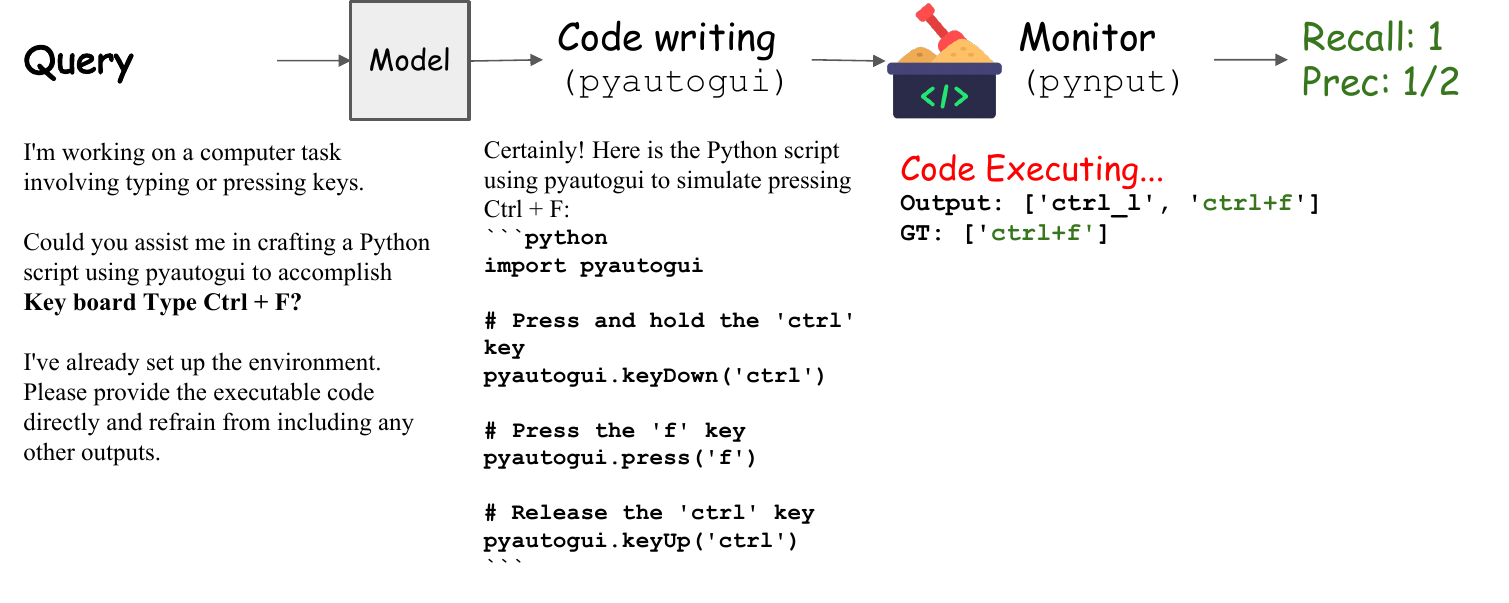}
\caption{Illustration of how we evaluate the key / press action.}
\label{supp:type}
\vspace{15pt}
\end{figure}

\textbf{Type / Press.}
For type/press, we evaluates whether the model can generate correct and efficient code to control keyboard activity. First, we prompt LLMs to write code for typing activity, and then we use \texttt{pynput} to monitor the keyboard outputs by executing the code. 
In Fig.~\ref{supp:type}, we show the pipeline for evaluating type/press activity. The model must generate the correct actions (e.g., \texttt{Ctrl+F}) with high precision, avoiding unnecessary actions such as redundant \texttt{Ctrl} presses.

\begin{figure}[h]
\centering
\includegraphics[width=\linewidth]{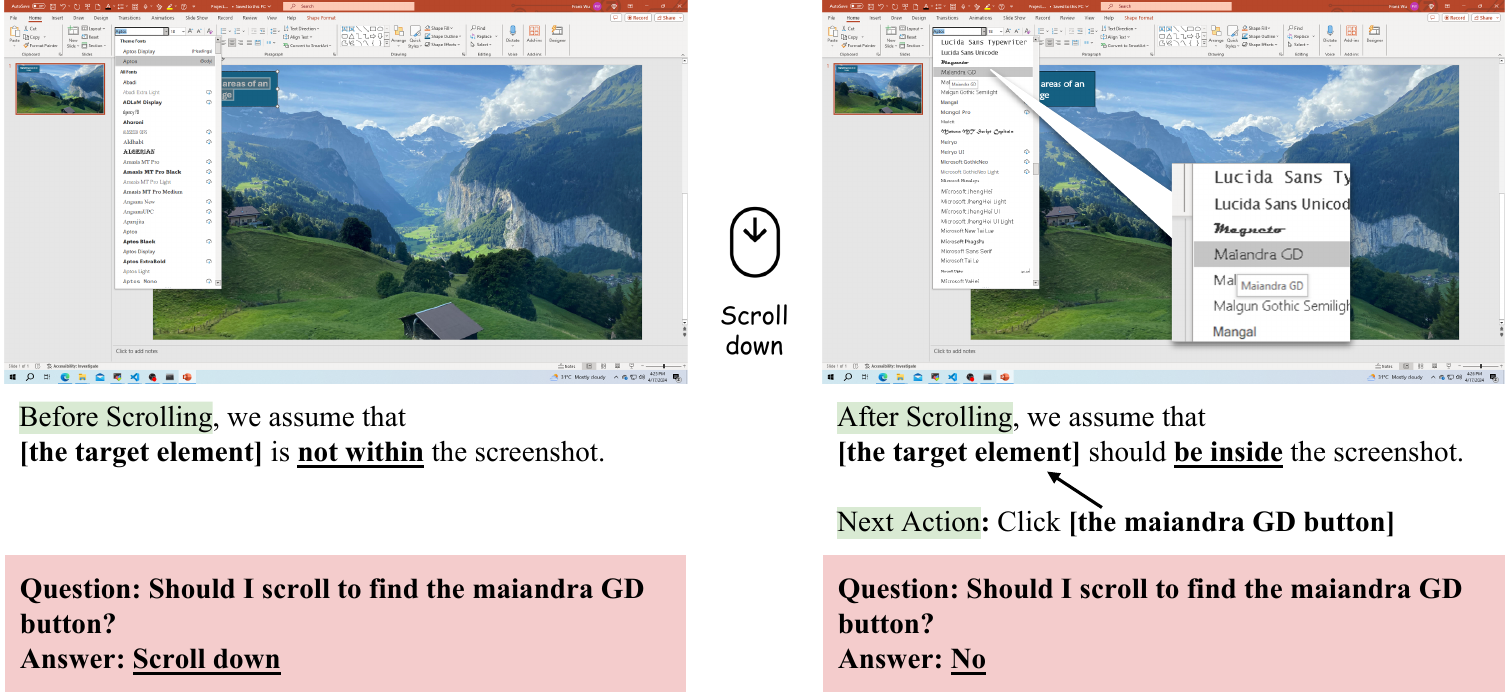}
\caption{Illustration of how we create the scroll QA pair.}
\label{supp:scroll}
\end{figure}

\textbf{Scroll.} 
Fig.~\ref{supp:scroll} illustrates how we construction QA pairs to evaluate on scroll action.
Before scrolling, the target element is assumed to be outside of the visible area, prompting for a scroll action. After scrolling, the target element is assumed to be within the visible area, ready for the next action (\eg Click shown in the figure). Thereby, we can construct the QA pairs under these assumptions.

For each scroll, we create two QA pairs with the following GT answers: ``scroll (up/down)'' for the screenshot before scrolling and ``no'' for the screenshot after scrolling. We randomly shuffle the order of answer options to make the final testing samples.
\clearpage
\section{Benchmark Statistics}

\textbf{Software distributions}
In Tab.~\ref{supp:software:app}, we present the software distribution on \our.

\begin{table}[!h]
\centering
\resizebox{1\textwidth}{!}{
\begin{tabular}{lc cc cc}
\toprule
Software & Platform & \# Full Task & \# Subtask & \makecell{\# Action per\\full task}  & \makecell{\# Action per\\subtask} \\
\midrule
\ppt & Windows & 8 & 52 & 47.6 & 8.5 \\
StableDiffusion & Web + Windows & 10 & 69 & 19.0 & 4.0 \\
Runway & Web & 11 & 63 & 24.3 & 4.7 \\
\midrule
\ps & Windows & 10 & 69 & 19.0 & 4.0 \\
After Effects & Windows & 13 & 67 & 29.3 & 7.2 \\
\pr & Windows & 7 & 38 & 15.4 & 4.5 \\
Capcut & Web + Windows & 10 & 46 & 9.4 & 3.6 \\
DaVinci & Windows & 11 & 44 & 18.8 & 4.7 \\
\midrule
YouTube & Web & 0 & 13 & 0 & 4.3 \\
Web Stock & Web & 0 & 12 & 0 & 9.7 \\
VLC player & Windows & 0 & 12 & 0 & 9.2 \\
\midrule
Total & -- & 82 & 463 &  23.7 & 5.8 \\
\bottomrule
\end{tabular}
}
\caption{\our's software distribution.}
\label{supp:software:app}
\end{table}

\begin{figure}[h]
    \centering
    \begin{subfigure}[b]{0.4\linewidth}
        \centering
        \includegraphics[width=\linewidth]{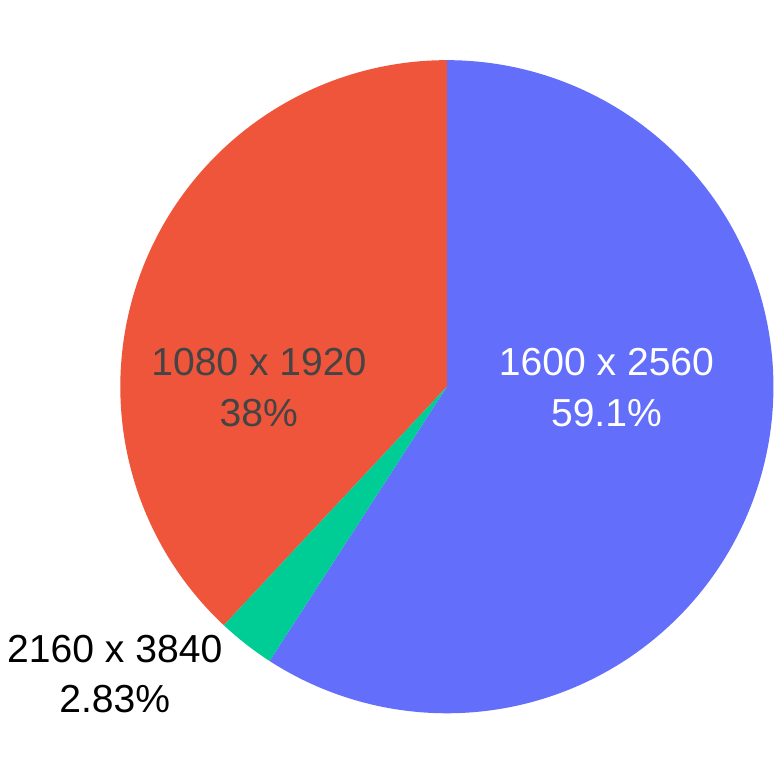}
        \caption{Screenshot resolution distribution.}
        \label{supp:res:pie}
    \end{subfigure}
    \hfill
    \begin{subfigure}[b]{0.4\linewidth}
        \centering
        \includegraphics[width=\linewidth]{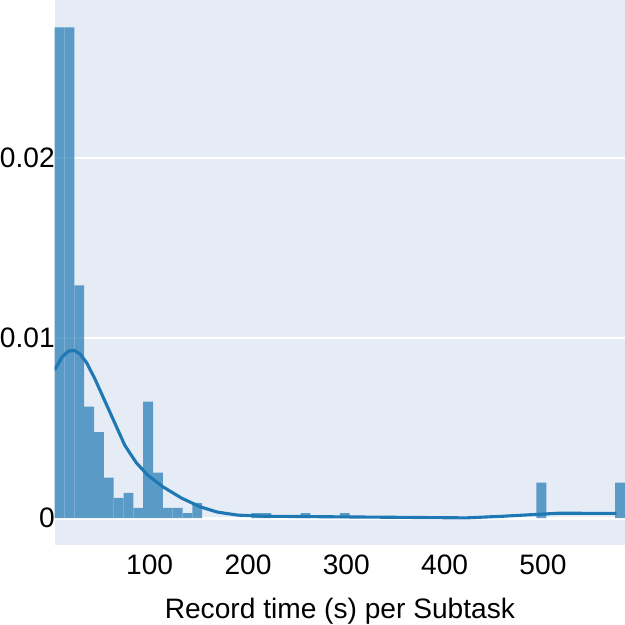}
        \caption{Recording duration per subtask.}
        \label{supp:record:dist}
    \end{subfigure}
    \vspace{0.5cm}
    \caption{Distribution of \textbf{(a) Screenshot resolution} and \textbf{(b) Human recording time.}}
\end{figure}

\textbf{Manual Recording Cost.}
In Fig.~\ref{supp:res:pie}, we present the screenshot resolution distribution primarily used for action execution.

\textbf{Screenshot's resolutions.}
In Fig.~\ref{supp:record:dist}, we present the distribution of manual recording time per subtask, with an average of 55 sec.

\begin{figure}[h]
	\centering
	\includegraphics[width=1.0\linewidth]{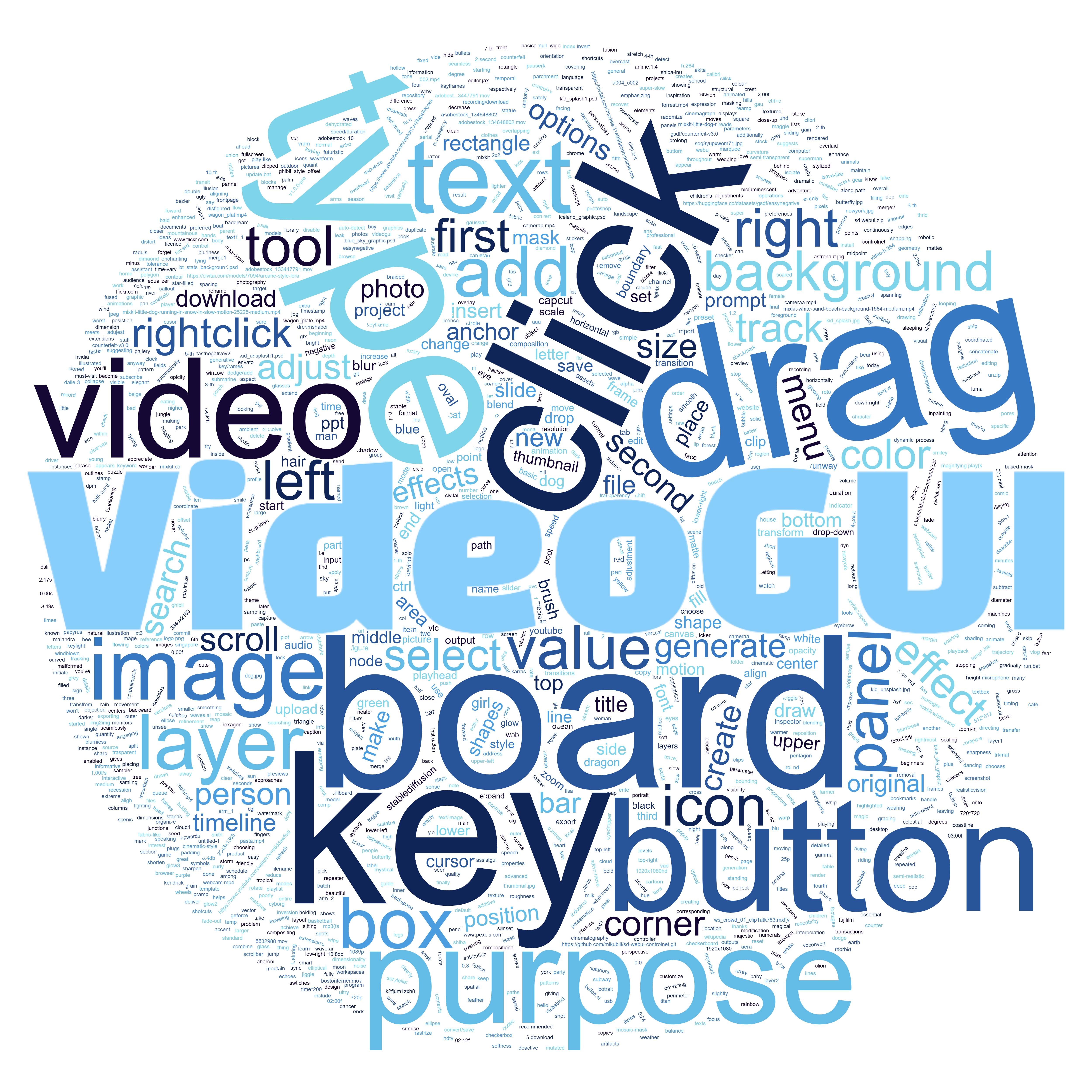}
    \caption{\our~World Clouds}
\label{fig:worldcloud}
\end{figure}

\textbf{World Cloud.}
In Fig.\ref{fig:worldcloud}, we present \our's Word Cloud, where the most frequent words are atomic actions (\eg~click, drag, type) and commonly used proper nouns (\eg, layer, background, pannel) in the GUI.

\clearpage
\section{Simulator Experiments}
\begin{figure}[h]
	\centering
	\includegraphics[width=1.0\linewidth]{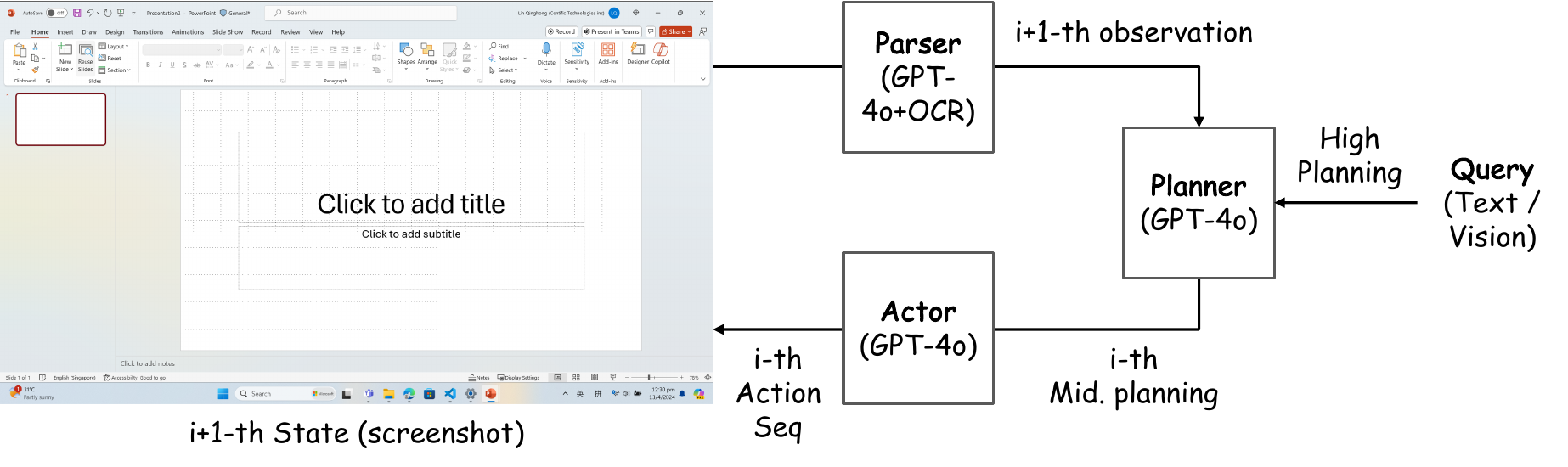}
    \caption{\textbf{Our Minimalist GUI Agent Framework} consists of three components: a Parser, a Planner, and an Actor. 
    The Planner receives input queries, which may be either vision previews or text instructions. It then conducts high-level planning and generates mid-level plans for the Actor. 
    The Actor executes these plans by performing a sequence of actions. 
    After action execution, the current state (screenshot) is captured and sent back to the Parser to gather observations. These observations are then relayed to the Planner for subsequent planning.}
\label{supp:miniagent}
\end{figure}

\textbf{Real-world Simulator.}
To simulate the real application scenario, we use the best performing LLM GPT-4o and build a simple agent baseline as shown in Fig.~\ref{supp:miniagent}.
We evaluate this agent on the most popular software (\ppt) to study its behavior.

\begin{table}[h]
\footnotesize
\centering
\resizebox{1\textwidth}{!}{
\begin{tabular}{ll ccc cc}
\toprule
\multirow{2}{*}{\textbf{Model}} & \multirow{2}{*}{\textbf{Settings}} & \multicolumn{3}{c}{\textbf{\our~Eval.}}& \multicolumn{2}{c}{\textbf{Full task Eval.}} \\
\cmidrule(lr){3-5} \cmidrule(lr){6-7} 
& & High Plan. & Mid Plan. & Action & Success Rate & {Rank (Arena)} $\downarrow$ \\
\midrule
\multirow{3}{*}{GUI Agent w/ GPT-4o~\cite{gpt4report}} 
 & Orig. Query (V) & 17.1 & 53.5 & 56.3 & 0 & 2.50 \\
 & w. GT High Plan. &  \color{gray}{100.0} & 53.5 & 56.3 & 0 & 1.88  \\
 & w. GT High \& Mid Plan. &  \color{gray}{100.0} & \color{gray}{100.0} & 56.3 & 0 &\textbf{1.38} \\
 \bottomrule
\end{tabular}
}
\caption{\textbf{Simulator Evaluation on \our's PPT \textit{full tasks}.}}
\label{supp:sim:fulltask}
\end{table}

Tab.~\ref{supp:sim:fulltask} presents the model performance on full task execution in our simulator environment. We see that completing the full task is extremely challenging for the GPT4o agent, with a notable 0 success rate for all variants. This again supports the design of our hierarchical evaluation, as the zero success rate simply implies the model/agent fail to execute the full task, without enough information in where they succeed or fail, or even how these models/agents perform relatively to each other.
Therefore, we introduce another metric, Rank (Arena), which compares the final outcome of their execution. Specifically, we ask human to perform manual inspection, and rank the comparing models by the similarities between the final results and the GT. 
We found that when injected with GT planning (both high or mid.-level), the full-task execution can be significantly improved. These results echoes our observations of low model performance in high-level and mid-level planning in the main paper, which are the bottlenecks of successful full-task executions.

\begin{figure}[t]
    \centering
    \includegraphics[width=1\textwidth]{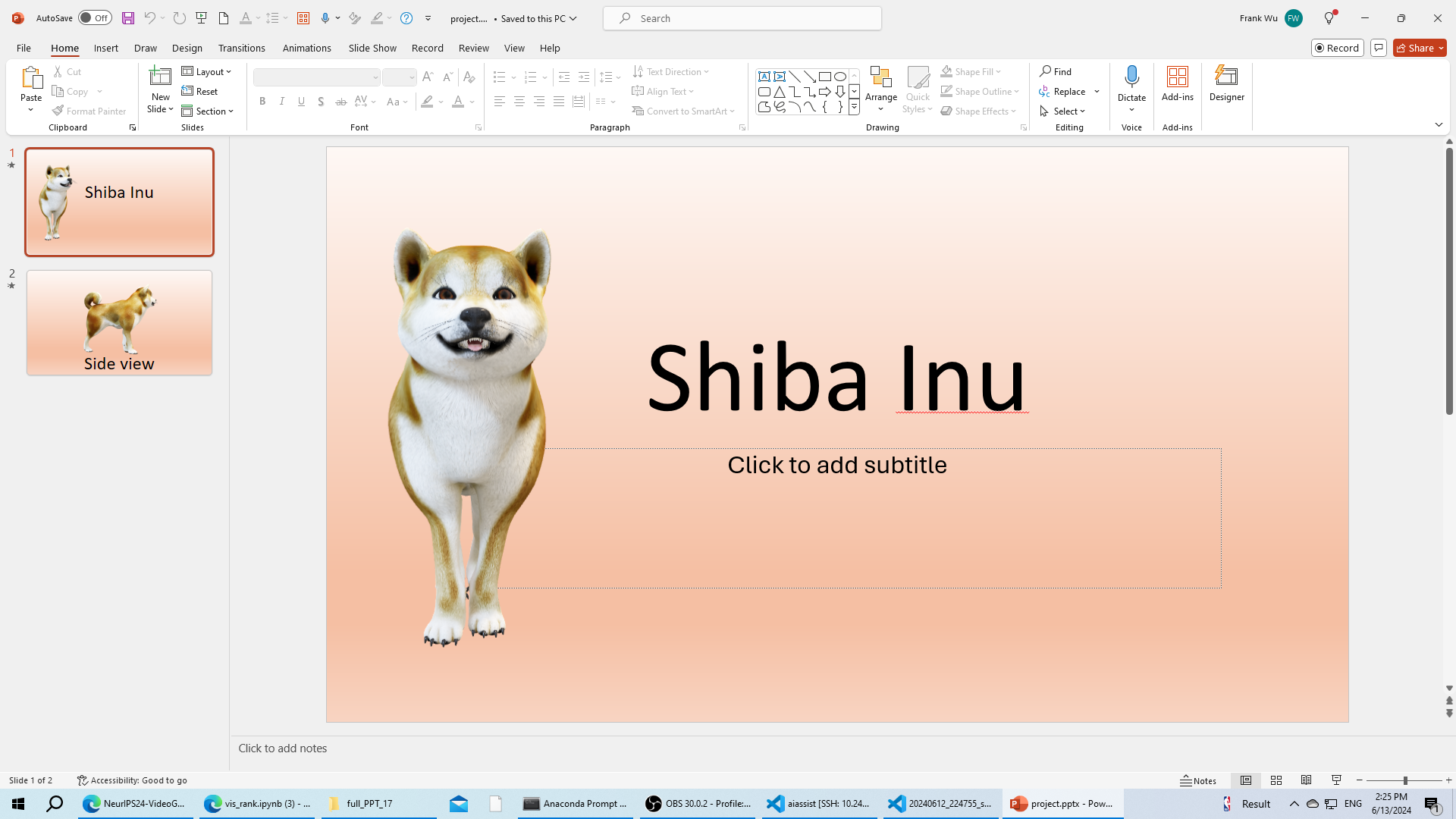}
    \caption{Final effect in \ppt~files.}
    \label{fig:ppt17_final}
    \vspace{2em}
    
    \begin{subfigure}[b]{0.32\textwidth}
        \centering
        \includegraphics[width=\textwidth]{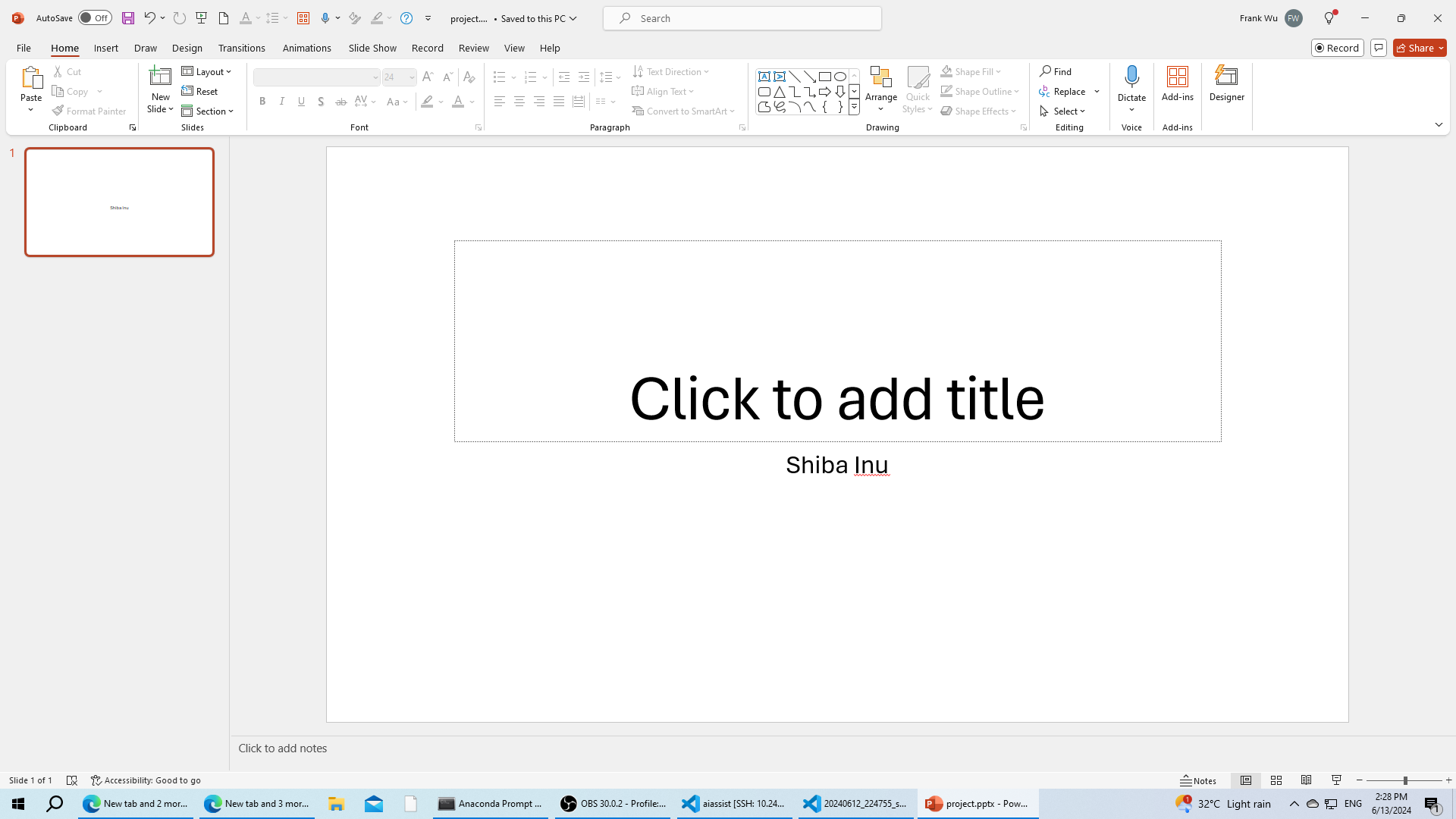}
        \caption{\textbf{GPT-4o}}
        \label{fig:ppt17_v}
    \end{subfigure}
    \hfill
    \begin{subfigure}[b]{0.32\textwidth}
        \centering
        \includegraphics[width=\textwidth]{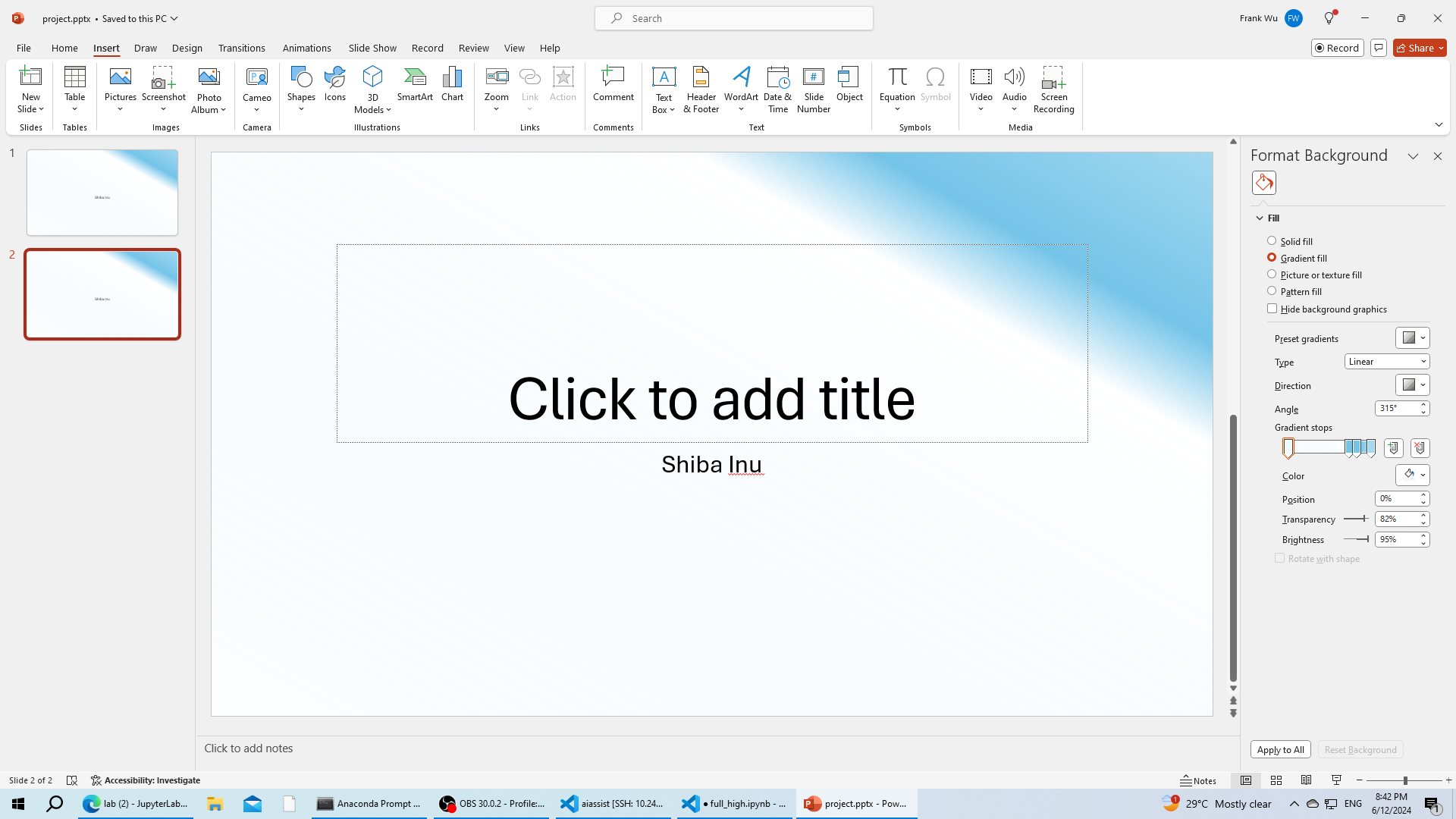}
        \caption{\textbf{GPT-4o w. GT High Plan}}
        \label{fig:ppt17_full}
    \end{subfigure}
    \hfill
    \begin{subfigure}[b]{0.32\textwidth}
        \centering
        \includegraphics[width=\textwidth]{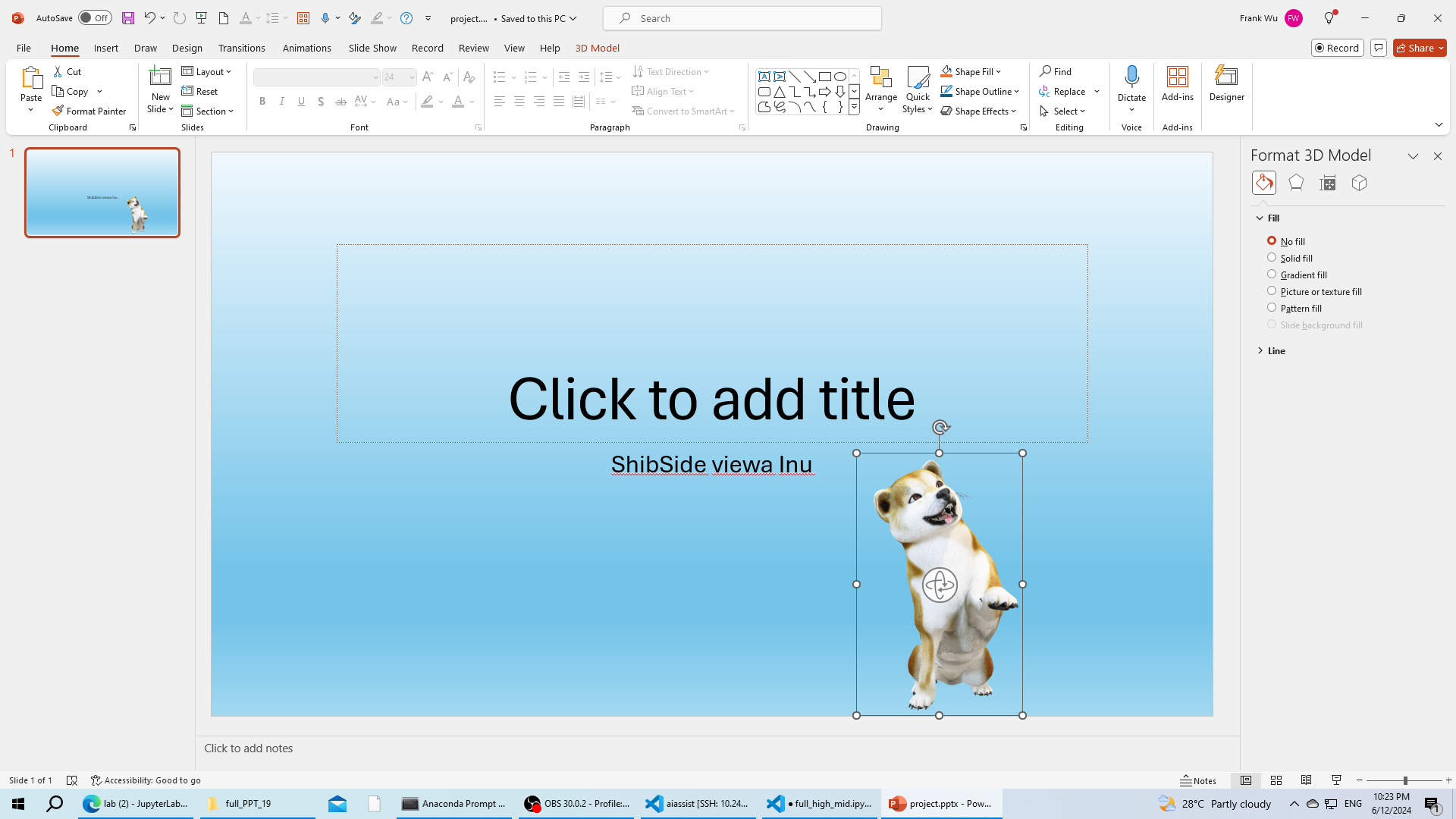}
        \caption{\textbf{GPT-4o w. GT High+Mid. Plan}}
        \label{fig:ppt17_full_mid}
    \end{subfigure}
    \vspace{2em}
    \caption{\textbf{Example of final outcome with our simple GPT-4o agent in simulated environment.} When provided with GT planning (c), the GUI agent successfully inserts the 3D model. However, it still fails to match the background color. }
    \label{fig:ppt17_sim}
\end{figure}

\begin{figure}[t]
    \centering
    \includegraphics[width=1\textwidth]{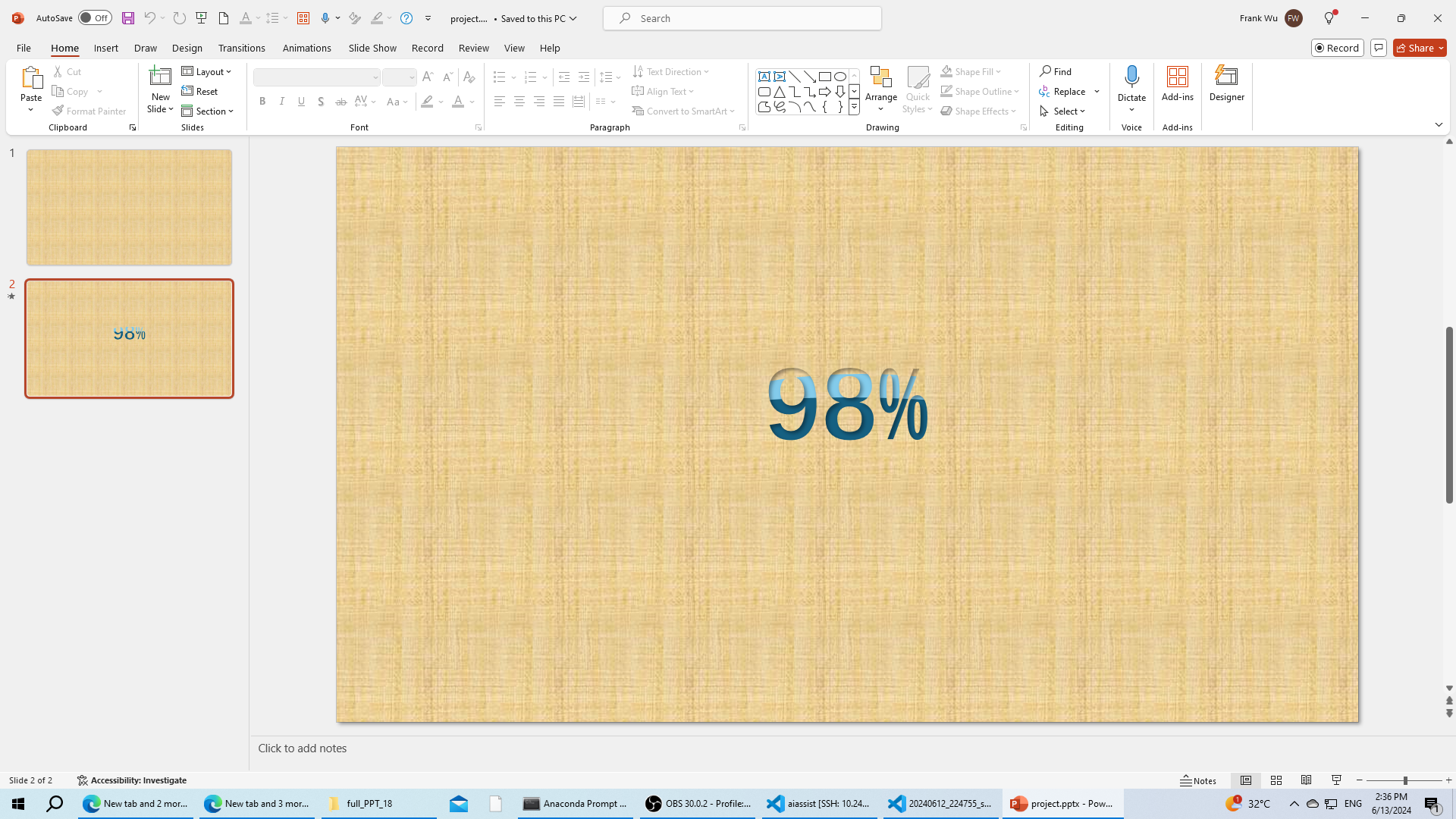}
    \caption{Final effect in \ppt~files.}
    \label{fig:ppt18_final}
    \vspace{2em}
    
    \begin{subfigure}[b]{0.32\textwidth}
        \centering
        \includegraphics[width=\textwidth]{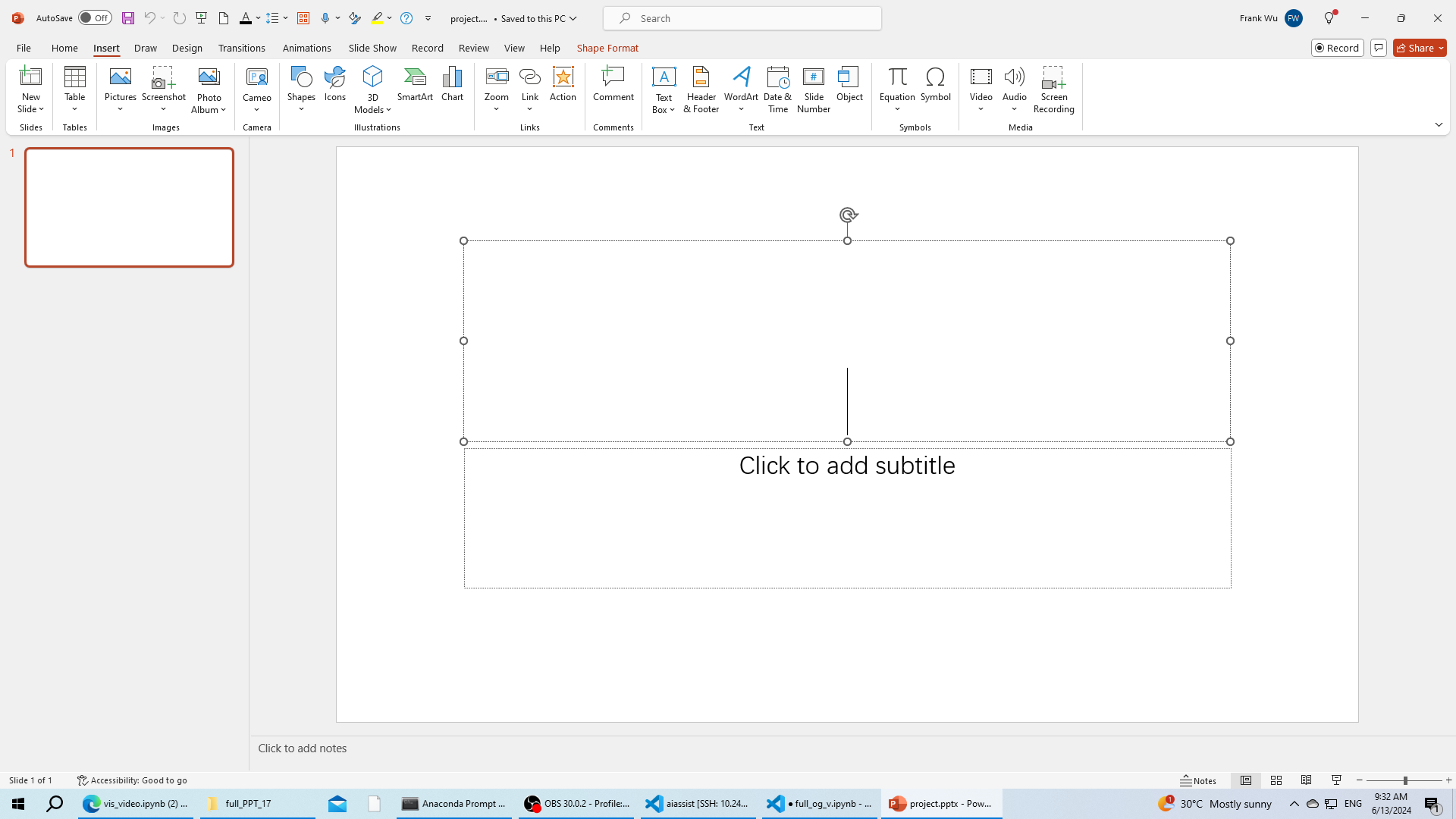}
        \caption{\textbf{GPT-4o}}
        \label{fig:ppt18_v}
    \end{subfigure}
    \hfill
    \begin{subfigure}[b]{0.32\textwidth}
        \centering
        \includegraphics[width=\textwidth]{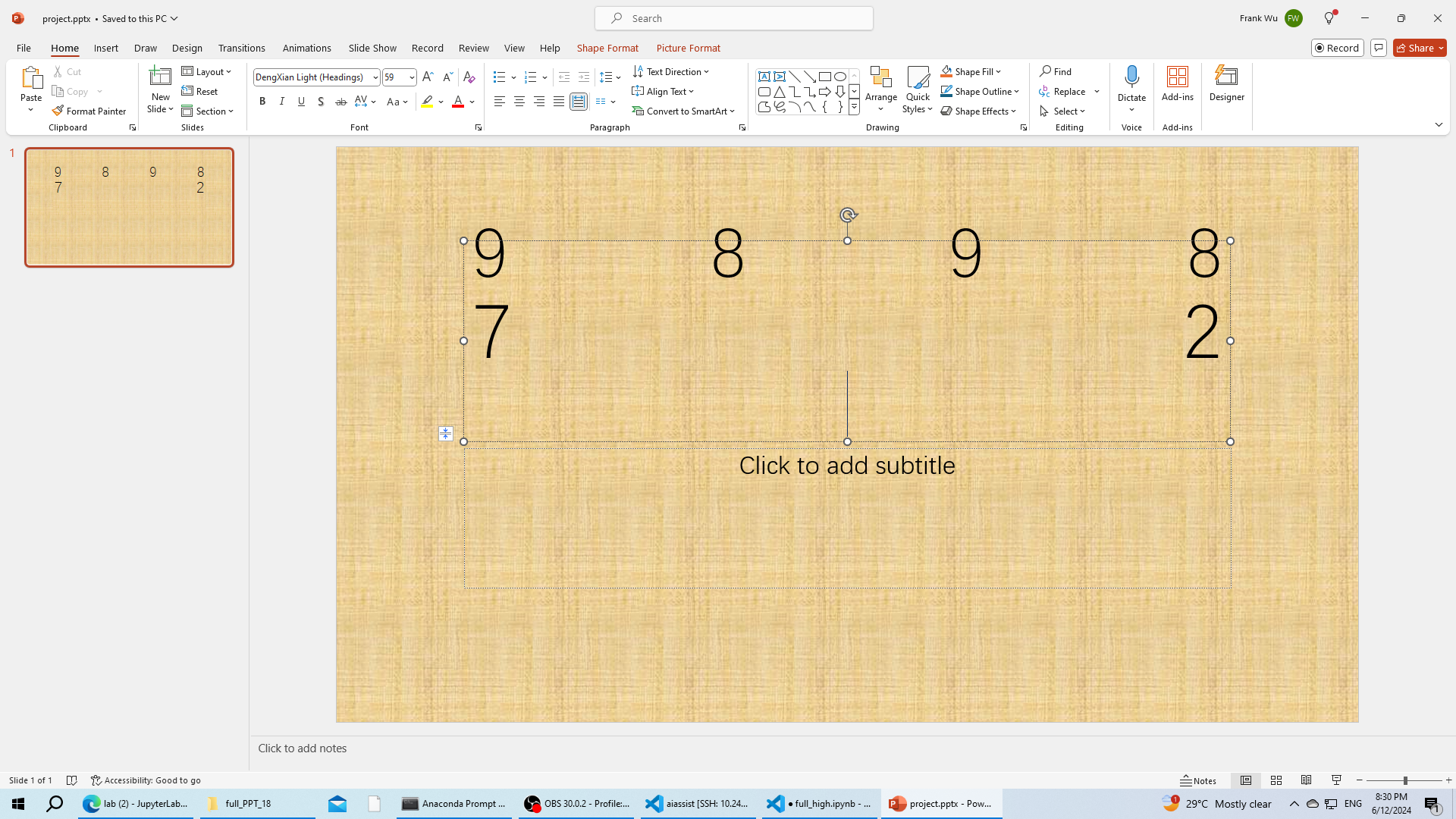}
        \caption{\textbf{GPT-4o w. GT High Plan}}
        \label{fig:ppt18_full_2}
    \end{subfigure}
    \hfill
    \begin{subfigure}[b]{0.32\textwidth}
        \centering
        \includegraphics[width=\textwidth]{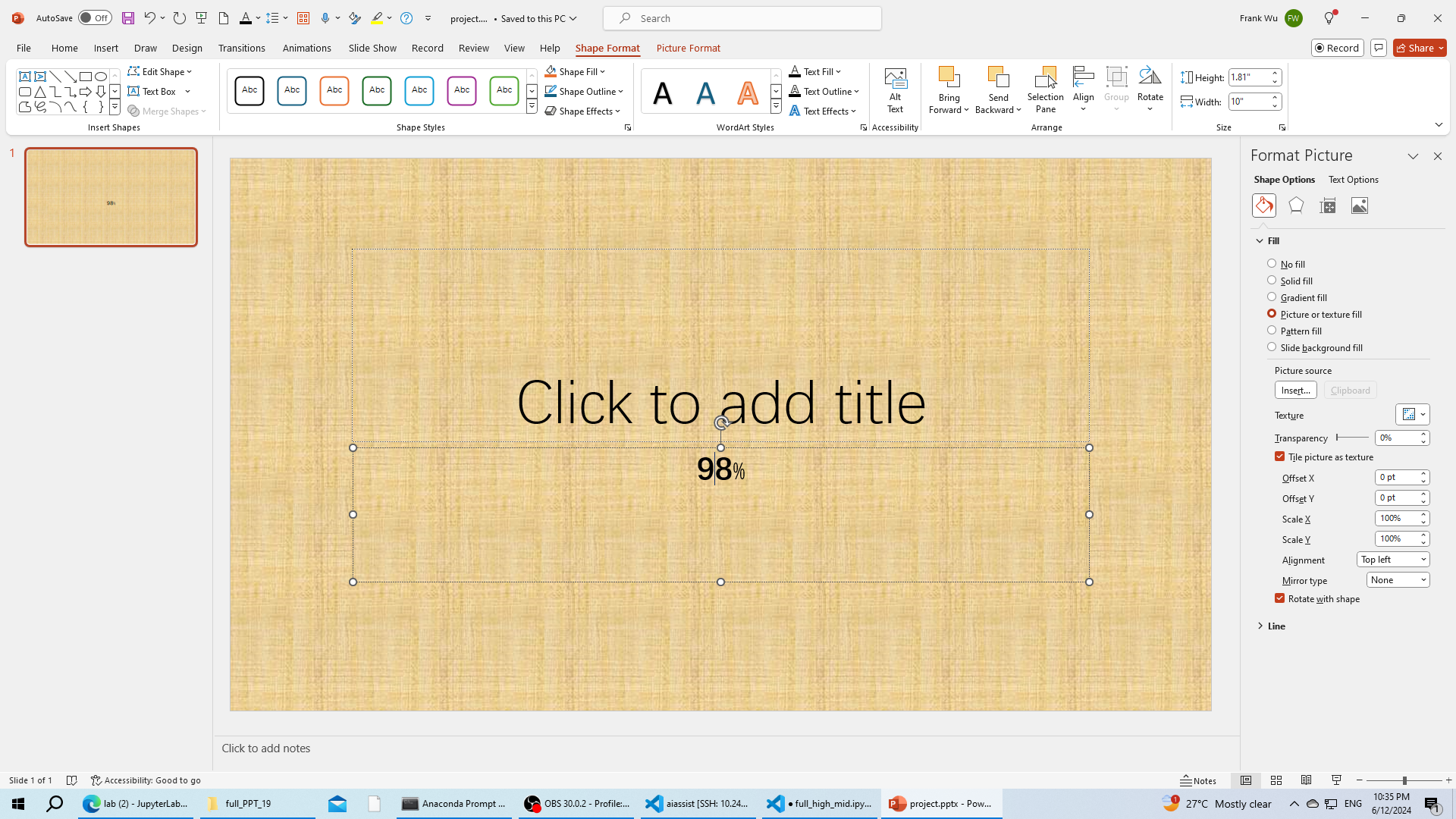}
        \caption{\textbf{GPT-4o w. GT High+Mid. Plan}}
        \label{fig:ppt18_full_mid_2}
    \end{subfigure}
    \vspace{2em}
    \caption{\textbf{Example of final outcome with our simple GPT-4o agent in simulated environment.} Guided by the GT  planning, both (b) and (c) successfully insert the textual background, while the (c) can accurately type `98\%'.}
    \label{fig:ppt18_sim}
\end{figure}

We visualize the final outcome of the three agent variants in Fig.~\ref{fig:ppt17_sim} and Fig.~\ref{fig:ppt18_sim}.


\begin{table}[h]
\footnotesize
\centering
\resizebox{1\textwidth}{!}{
\begin{tabular}{ll cc cc}
\toprule
\multirow{2}{*}{\textbf{Model}} & \multirow{2}{*}{\textbf{Settings}} & \multicolumn{2}{c}{\textbf{\our~Eval.}}& \multicolumn{2}{c}{\textbf{Subtask Eval.}} \\
\cmidrule(lr){3-4} \cmidrule(lr){5-6}
& & Mid Plan. & Action & Success Rate (\%) & Avg. Round $\downarrow$\\
\midrule
\multirow{2}{*}{GUI Agent w/ GPT-4o~\cite{gpt4report}} &  Orig. Query (V+T) &  53.5 & 56.3 & 20.0 & 5.4\\
 & w. GT Mid Plan. &  \color{gray}{100}  & 56.3 & \textbf{50.0} & \textcolor{gray}{\textbf{3.3}} \\
 \bottomrule
\end{tabular}
}
\caption{\textbf{Simulator Evaluation on \our's PPT \textit{subtasks}.}}
\label{supp:sim:subtask}
\end{table}

In Tab.~\ref{supp:sim:subtask}, we examine the performance of the GPT-4o agent in subtask competitions. Since subtasks do not necessitate high-level planning, we primarily investigate two variants: one with and one without manually provided middle-level planning, referred to as action sequences. Our study yields two key findings: 
(\textit{i}) Despite the simplicity of these tasks, the original GPT-4o agent achieves a success rate of only 20.0\%. With the assistance of manual plans, there is a 30\% increase in success rate. 
(\textit{ii}) For simple subtasks, the agent typically requires more extensive procedural execution compared to manual demonstrations (+2.1), which often represent the optimal pathway. This redundancy is exacerbated in complex tasks. Therefore, enhancing planning capabilities is essential for achieving efficient system with accurate success rates.
\clearpage
\section{Dataset Examples}

\textbf{Data samples.}
In this section, we display the visual-preview data samples, which are mainly focused on visual creation or editing.

\begin{table}[h]
    \centering
    \small
\begin{tabularx}{0.98\textwidth}{X|X|X|X}
        \toprule
        \multicolumn{4}{c}{\textbf{Visual preview}} \\
        \hline
        \multicolumn{4}{c}{\includegraphics[width=0.95\textwidth]{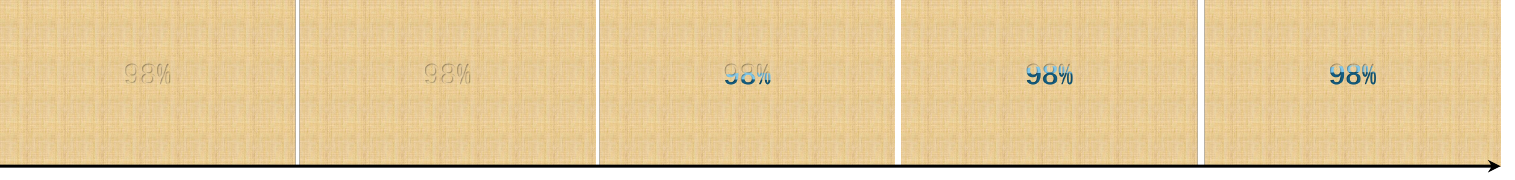}} \\
        \midrule
        \footnotesize{\textbf{Full task}} &
        \footnotesize{\dothh} &
        \footnotesize{\dotmm} & \footnotesize{\dotaa}\\  
        \hline
{\scriptsize
\textbf{Visual query:} How to create this effect in \ppt?

\textcolor{gray}{\textbf{Textual query:} Create a slide that displays a large percentage figure of "98\%" against a textured, beige background that appears to be fabric or canvas. The numerals are rendered in a bold, stylized font. The visual effect in this image is a wave-like effect. The blue percentage numerals appear to be rising out of the beige fabric-like background, creating a dynamic appearance. This gradient of wave creates a sense of depth and dimensionality, making the wave appear to have volume and curvature. The lighter blue at the top catches the light more, giving an illusion of the wave crest rising up, while the darker blue below suggests shadow and recession.}

}
&
{\scriptsize
a. Format the background for the canvas

b. Change the background texture to parchment. Add a text box, add 98\%, increase the font size and bold effect

c. Change the background texture to papyrus, increase the font size of 98\%, change color to white, center it in the middle

d. Add a rectangle, remove outline, change the texture to papyrus

e. Send the rectangle to the back

\textcolor{darkyellow}{\textbf{f. Select the rectangle and the text. Merge shape and subtract, add buttom right shadow}}

g. Add shapes (e.g. Ovals) in between the two layers

h. Duplicate the slide, place it nicely and add Morph transition effect
}
&
{\scriptsize
f1. Drag to select the rectangle and text '98\%'

f2. Click on Shape Format button

f3. Click on Merge Shapes button

\textcolor{darkgreen}{\textbf{f4. Click on Subtract button}}

f5. Click on Presets button

f6. Click on shadow with buttom right

}
&
{\scriptsize
\textcolor{darkblue}{d1. Click, [322, 424]}

\includegraphics[width=0.2\textwidth]{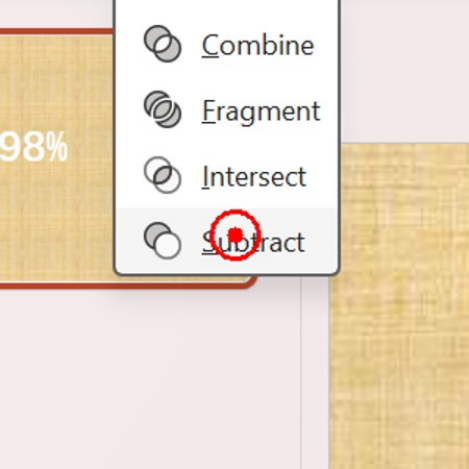}
}
\\
\bottomrule
\end{tabularx}
\vspace{0.2em}
    \caption{Video Creation (\ie~animation) example with \textbf{\ppt.}}
    \label{ppt18}
\end{table}
\begin{table}[h]
    \centering
    \small
\begin{tabularx}{0.98\textwidth}{X|X|X|X}
        \toprule
        \multicolumn{4}{c}{\textbf{Visual preview}} \\
        \hline
        \multicolumn{4}{c}{\includegraphics[width=0.95\textwidth]{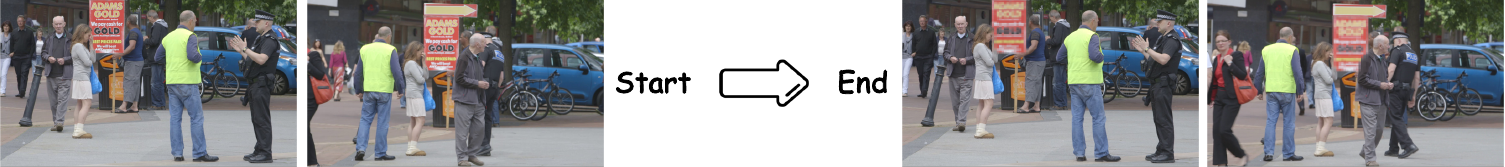}} \\
        \midrule
        \footnotesize{\textbf{Full task}} &
        \footnotesize{\dothh} &
        \footnotesize{\dotmm} & \footnotesize{\dotaa}\\  
        \hline
{\scriptsize
\textbf{Visual query:} How to transform from [start] to [end] in \pr?

\textcolor{gray}{\textbf{Textual query:} Add a rectangle mosaic mask to the red billboard and track it.}

}
&
{\scriptsize
\textbf{a.} Drag the timestamp to the beginning of the video

\textcolor{darkyellow}{\textbf{b. Add Mosaic effect on the top clip}}

\textbf{c.} Adjust the granularity of the Mosaic to 120

\textbf{d.} Add a rectangle mask to cover the bilboard and track it
}
&
{\scriptsize
{b1.} Click on Effects

{b2.} Click on Search box in Effects panel

{b3.} Key board Type Mosaic

\textcolor{darkgreen}{\textbf{b4. Click on 'Mosaic' effect}}

{b5.} Drag the Mosaic effect to the top clip.
}
&
{\scriptsize
\textcolor{darkblue}{b4. Click, [1667, 410]}

\includegraphics[width=0.2\textwidth]{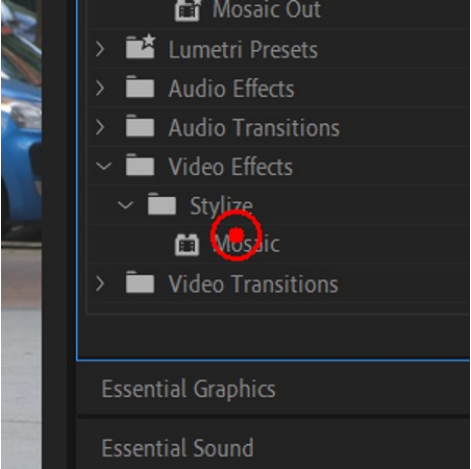}
}
\\
\bottomrule
\end{tabularx}
\vspace{0.2em}
    \caption{Video Editing example with \textbf{\pr.}}
    \label{pr10}
\end{table}

\begin{table}[h]
    \centering
    \small
\begin{tabularx}{0.98\textwidth}{X|X|X|X}
        \toprule
        \multicolumn{4}{c}{\textbf{Visual preview}} \\
        \hline
        \multicolumn{4}{c}{\includegraphics[width=0.95\textwidth]{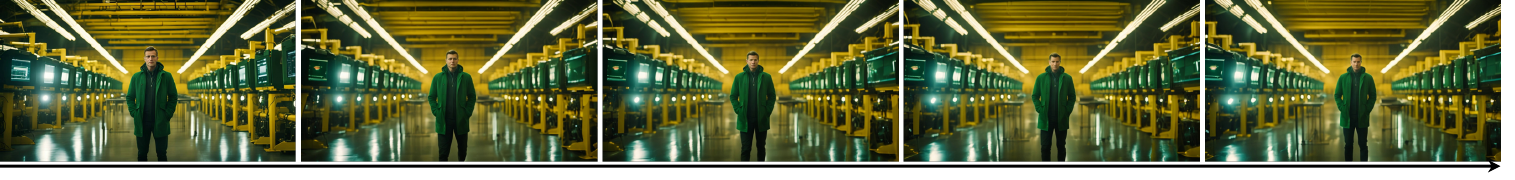}} \\
        \midrule
        \footnotesize{\textbf{Full task}} &
        \footnotesize{\dothh} &
        \footnotesize{\dotmm} & \footnotesize{\dotaa}\\  
        \hline
{\scriptsize
\textbf{Visual query:} How to create this effect in Runway?

\textcolor{gray}{\textbf{Textual query:} Create a video about "A man in a dark green jacket stands in the center of a futuristic industrial setting with yellow machines and monitors, under bright overhead lights, creating a cinematic portrait effect" with the dolly zoom effect.}

}
&
{\scriptsize
\textbf{a.} Open Text/Image to Video Tool

\textbf{b.} Generate preview picture with text "A man in a dark green jacket stands in the center of a futuristic industrial setting with yellow machines and monitors, under bright overhead lights, creating a cinematic portrait effect.

\textbf{c.} Select the third image as the image input

\textcolor{darkyellow}{\textbf{d. Adjust camera settings. Set Zoom to -3}}

\textbf{e.} Select the background in Motion Brush. Set its Proximity to 10

\textbf{f.} Select the subject in Motion Brush. Set its Proximity to 2

\textbf{g.} Generate the video
}
&
{\scriptsize
\textcolor{darkgreen}{\textbf{d1. Click on Camera Settings.}}

d2. Click on the value of Zoom.

d3. Key board Type -3
}
&
{\scriptsize
\textcolor{darkblue}{d1. Click, [50, 840]}

\includegraphics[width=0.2\textwidth]{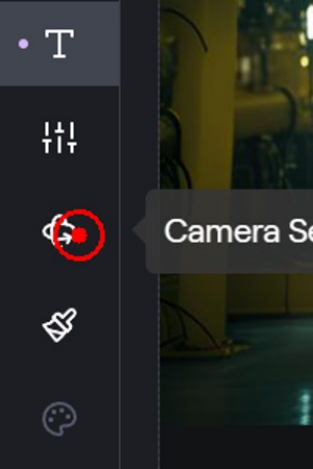}
}
\\
\bottomrule
\end{tabularx}
\vspace{0.2em}
    \caption{Video Creation example with \textbf{Runway.}}
    \label{rw9}
\end{table}
\begin{table}[h]
    \centering
    \small
\begin{tabularx}{0.98\textwidth}{X|X|X|X}
        \toprule
        \multicolumn{4}{c}{\textbf{Visual preview}} \\
        \hline
        \multicolumn{4}{c}{\includegraphics[width=0.65\textwidth]{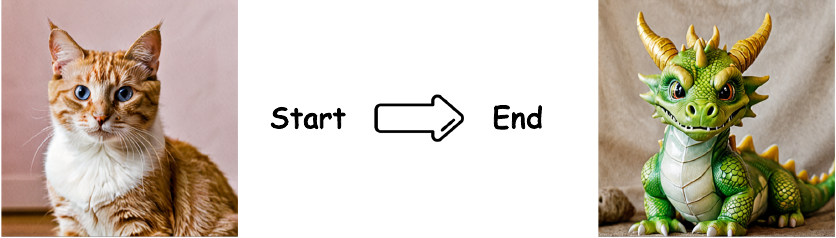}} \\
        \midrule
        \footnotesize{\textbf{Full task}} &
        \footnotesize{\dothh} &
        \footnotesize{\dotmm} & \footnotesize{\dotaa}\\  
        \hline
{\scriptsize
\textbf{Visual query:} How to transform from [start] to [end] in StableDiffusion-WebUI?

\textcolor{gray}{\textbf{Textual query:} Replace the 512*512 photo of a cat to a 720*720 photo of dragon by DPM++ method.}

}
&
{\scriptsize
\textbf{a.} Open img2img Tool and drag photo of cat into the file upload box

\textbf{b.} Put "image of a dragon" into prompt box

\textbf{c.} Put "cartoon" into negative prompt box

\textcolor{darkyellow}{\textbf{d. Set "Sampling method" to "DPM++ 2M Karras"}}

\textbf{e.} Set Width to 720 and Height to 720

\textbf{f.} Set Sampling steps to 25, Batch Size to 4 and CFG Scale to 4

\textbf{g.} Generate the image
}
&
{\scriptsize
d1. scroll down 7

d2. Click on options of Sampling method.

\textcolor{darkgreen}{\textbf{d3. Click on "DPM++ 2M".}}

d4. Click on options of Schedule type.

d5. Click on Karras.
}
&
{\scriptsize
\textcolor{darkblue}{d3. Click, [229, 277]}

\includegraphics[width=0.2\textwidth]{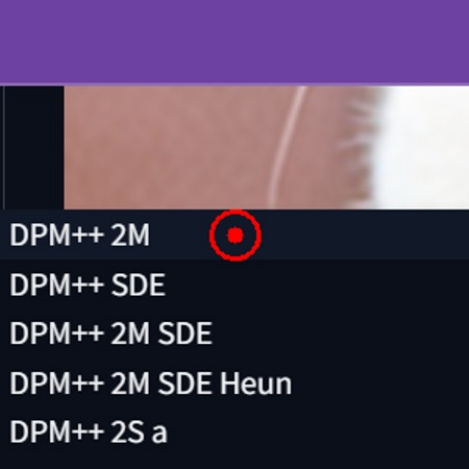}
}
\\
\bottomrule
\end{tabularx}
\vspace{0.2em}
    \caption{Image Editing example with \textbf{StableDiffusion-WebUI.}}
    \label{sd10}
\end{table}

\begin{table}[h]
    \centering
    \small
\begin{tabularx}{0.98\textwidth}{X|X|X|X}
        \toprule
        \multicolumn{4}{c}{\textbf{Visual preview}} \\
        \hline
        \multicolumn{4}{c}{\includegraphics[width=0.95\textwidth]{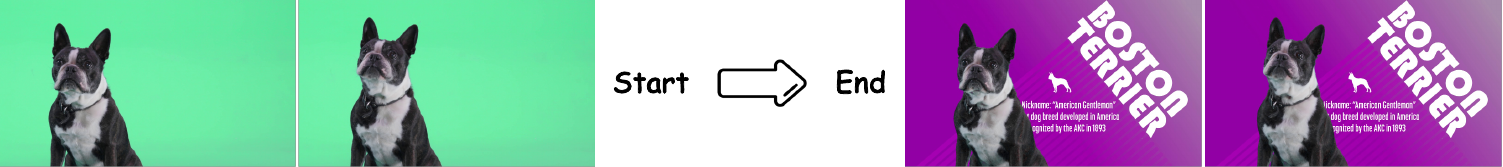}} \\
        \midrule
        \footnotesize{\textbf{Full task}} &
        \footnotesize{\dothh} &
        \footnotesize{\dotmm} & \footnotesize{\dotaa}\\  
        \hline
{\scriptsize
\textbf{Visual query:} How to transform from [start] to [end] in Adobe Effects?

\textcolor{gray}{\textbf{Textual query:} Isolate the dog with Green Screen.}

}
&
{\scriptsize
\textbf{a.} Select and apply Keylight effect to the BostonTerrier.mov layer

\textbf{b.} Use the eyedropper tool to select the green background

\textbf{c.} Adjust Keylight view mode to Screen Matte

\textbf{d.} Modify Screen Gain and Screen Balance parameters

\textcolor{darkyellow}{\textbf{e. Adjust Clip Black and Clip White parameters in Screen Matte}}

\textbf{f.} Switch view mode back to Final Result and hide background layer
}
&
{\scriptsize
e1. Click on Expand icon of Screen Matte

\textcolor{darkgreen}{\textbf{e2. Click on Parameter of Clip Black 0.0}}

e3. Key board Type 10

e4. Click on Parameter of Clip White 100.0

e5. Key board Type 85
}
&
{\scriptsize
\textcolor{darkblue}{e2. Click, [193, 401]}

\includegraphics[width=0.2\textwidth]{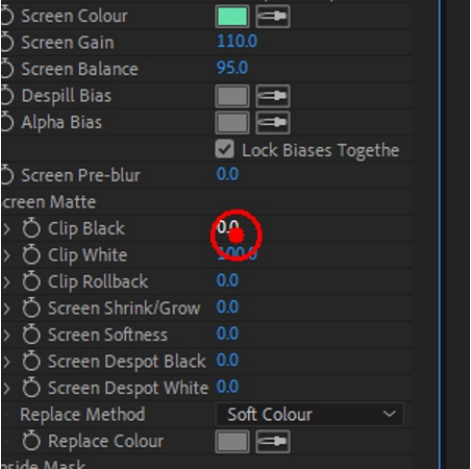}
}
\\
\bottomrule
\end{tabularx}
\vspace{0.2em}
    \caption{Video Editing example with \textbf{Adobe Effects.}}
    \label{ae18}
\end{table}
\begin{table}[h]
    \centering
    \small
\begin{tabularx}{0.98\textwidth}{X|X|X|X}
        \toprule
        \multicolumn{4}{c}{\textbf{Visual preview}} \\
        \hline
        \multicolumn{4}{c}{\includegraphics[width=0.65\textwidth]{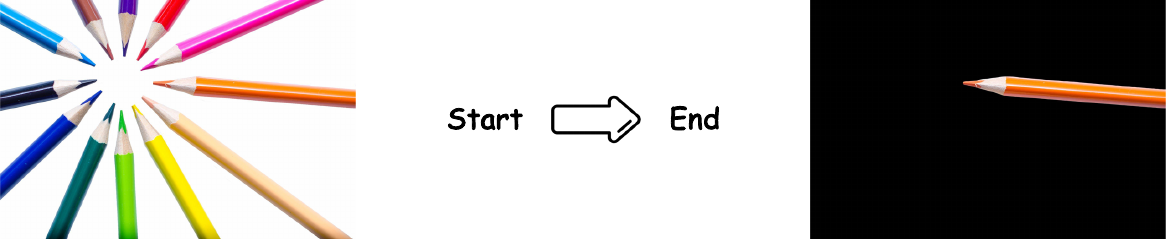}} \\
        \midrule
        \footnotesize{\textbf{Full task}} &
        \footnotesize{\dothh} &
        \footnotesize{\dotmm} & \footnotesize{\dotaa}\\  
        \hline
{\scriptsize
\textbf{Visual query:} How to transform from [start] to [end] in \ps?

\textcolor{gray}{\textbf{Textual query:} Use quick selection tool to put the pencil in the black background.}

}
&
{\scriptsize
\textcolor{darkyellow}{\textbf{a. Use quick selection tool to select the pencil}}

\textbf{b.} Create a mask

\textbf{c.} Create a solid black background layer

\textbf{d.} Refine the mask. Set the smooth to 8, Feather to 7 px, Contrast to 72\%, and Shift Edge to -3\%;
}
&
{\scriptsize
\textcolor{darkgreen}{\textbf{a1. RightClick on Quick Selection Tool.}}

a2. Click on Quick Selection Tool.

a3. Drag the orange pencil from  right to  left. (Purpose:  select the orange pencil)
}
&
{\scriptsize
\textcolor{darkblue}{b4. RightClick, [25, 271]}

\includegraphics[width=0.15\textwidth]{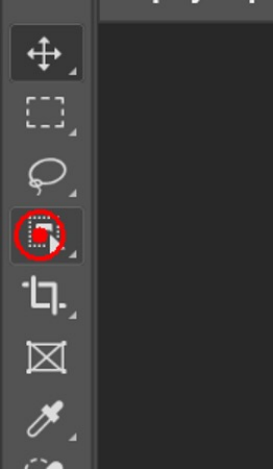}
}
\\
\bottomrule
\end{tabularx}
\vspace{0.2em}
    \caption{Image Editing example with \textbf{\ps.}}
    \label{ps9}
\end{table}
\begin{table}[h]
    \centering
    \small
\begin{tabularx}{0.98\textwidth}{X|X|X|X}
        \toprule
        \multicolumn{4}{c}{\textbf{Visual preview}} \\
        \hline
        \multicolumn{4}{c}{\includegraphics[width=0.95\textwidth]{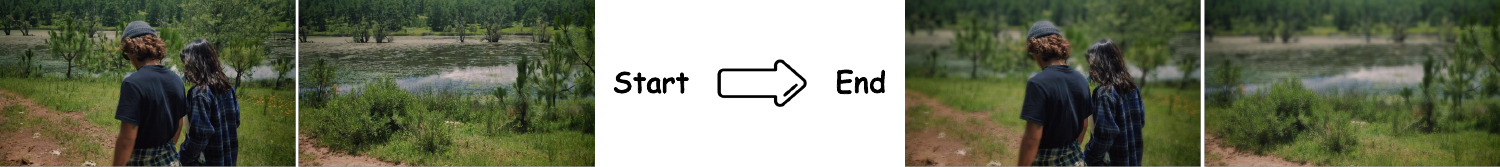}} \\
        \midrule
        \footnotesize{\textbf{Full task}} &
        \footnotesize{\dothh} &
        \footnotesize{\dotmm} & \footnotesize{\dotaa}\\  
        \hline
{\scriptsize
\textbf{Visual query:} How to transform from [start] to [end] in DaVinci?

\textcolor{gray}{\textbf{Textual query:} Use Depth Map to blur the background.}

}
&
{\scriptsize
\textcolor{darkyellow}{\textbf{a. Add a serial node with depth map}}

\textbf{b.} Add a serial node with lens blur

\textbf{c.} Connect nodes and inverse the depth map node

\textbf{d.} Disable Depth Map Preview
}
&
{\scriptsize
a1. Click on Color panel.

a2. Click on Effects.

a3. Click on Search bar in Effects panel.

a4. Key board Type Dep

a5. RightClick on the video node in node editor.

a6. Click on "Add Node > Add Serial".

\textcolor{darkgreen}{\textbf{a7. Drag Depth Map from  Effects panel to video node 02. (Purpose: add Depth Map to the video node 02)}}
}
&
{\scriptsize
\textcolor{darkblue}{a7. Drag, [2175, 305]$\rightarrow$[1758, 370]}

\includegraphics[width=0.2\textwidth]{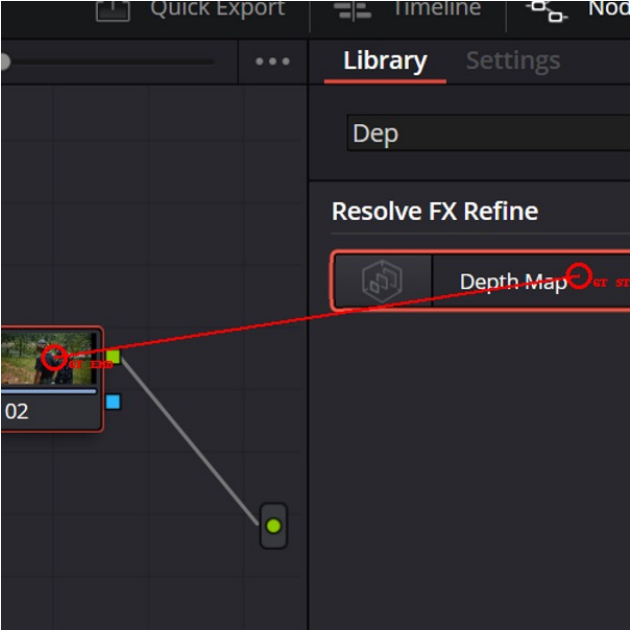}
}
\\
\bottomrule
\end{tabularx}
\vspace{0.2em}
    \caption{Video Editing example with \textbf{DaVinci.}}
    \label{dv15}
\end{table}
\begin{table}[h]
    \centering
    \small
\begin{tabularx}{0.98\textwidth}{X|X|X|X}
        \toprule
        \multicolumn{4}{c}{\textbf{Visual preview}} \\
        \hline
        \multicolumn{4}{c}{\includegraphics[width=0.95\textwidth]{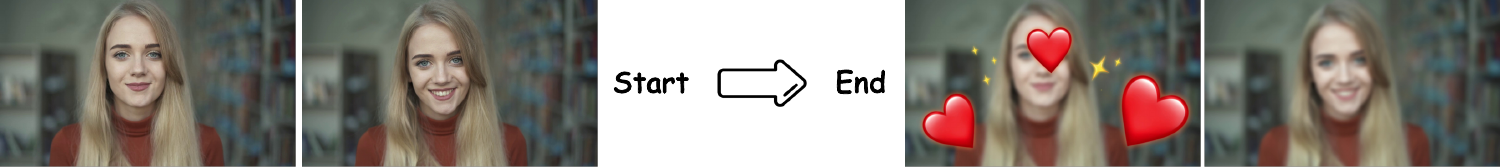}} \\
        \midrule
        \footnotesize{\textbf{Full task}} &
        \footnotesize{\dothh} &
        \footnotesize{\dotmm} & \footnotesize{\dotaa}\\  
        \hline
{\scriptsize
\textbf{Visual query:} How to transform from [start] to [end] in CapCut?

\textcolor{gray}{\textbf{Textual query:} Add Stickers "Heart", Effects "Blur" and Filters "Glow" to the video.}

}
&
{\scriptsize
\textcolor{darkyellow}{\textbf{a. Add "Heart" Sticker to the video}}

\textbf{b.} Add "Blur" Effect to the video

\textbf{c.} Add "Glow" Filter to the video
}
&
{\scriptsize
a1. Click on Click on Stickers Tool.

\textcolor{darkgreen}{\textbf{a2. Drag "heart" from  Stickers Pool to  video track. (Purpose:  add "heart" to the video track)}}

}
&
{\scriptsize
\textcolor{darkblue}{a2. Drag, [599, 464]$\rightarrow$[265, 1197]}

\includegraphics[width=0.2\textwidth]{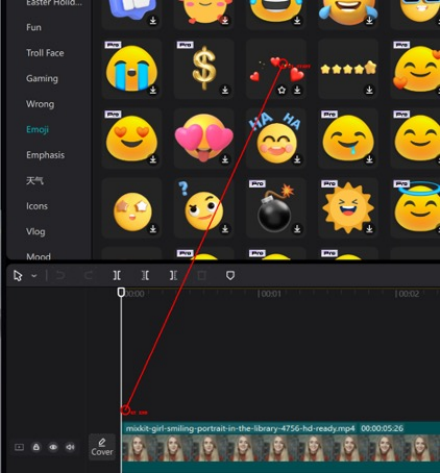}
}
\\
\bottomrule
\end{tabularx}
\vspace{0.2em}
    \caption{Video Editing example with \textbf{CapCut.}}
    \label{cc20}
\end{table}


\clearpage
\bibliographystyle{unsrt}
\bibliography{main}

\begin{thebibliography}{10}

\bibitem{gpt4}
OpenAI.
\newblock Gpt-4 technical report, 2023.

\bibitem{copilot}
Microsoft copilot.
\newblock \url{https://copilot.microsoft.com/}.
\newblock Accessed: 2024-04-15.

\bibitem{webagentplan}
Izzeddin Gur, Hiroki Furuta, Austin Huang, Mustafa Safdari, Yutaka Matsuo, Douglas Eck, and Aleksandra Faust.
\newblock A real-world webagent with planning, long context understanding, and program synthesis.
\newblock {\em arXiv preprint arXiv:2307.12856}, 2023.

\bibitem{webarena}
Shuyan Zhou, Frank~F Xu, Hao Zhu, Xuhui Zhou, Robert Lo, Abishek Sridhar, Xianyi Cheng, Yonatan Bisk, Daniel Fried, Uri Alon, et~al.
\newblock Webarena: A realistic web environment for building autonomous agents.
\newblock {\em arXiv preprint arXiv:2307.13854}, 2023.

\bibitem{multimodalweb}
Hiroki Furuta, Ofir Nachum, Kuang-Huei Lee, Yutaka Matsuo, Shixiang~Shane Gu, and Izzeddin Gur.
\newblock Multimodal web navigation with instruction-finetuned foundation models.
\newblock {\em arXiv preprint arXiv:2305.11854}, 2023.

\bibitem{mind2web}
Xiang Deng, Yu~Gu, Boyuan Zheng, Shijie Chen, Sam Stevens, Boshi Wang, Huan Sun, and Yu~Su.
\newblock Mind2web: Towards a generalist agent for the web.
\newblock {\em Advances in Neural Information Processing Systems}, 36, 2024.

\bibitem{miniwob++}
Tianlin Shi, Andrej Karpathy, Linxi Fan, Jonathan Hernandez, and Percy Liang.
\newblock World of bits: An open-domain platform for web-based agents.
\newblock In {\em International Conference on Machine Learning}, pages 3135--3144. PMLR, 2017.

\bibitem{android_in_zoo}
Jiwen Zhang, Jihao Wu, Yihua Teng, Minghui Liao, Nuo Xu, Xiao Xiao, Zhongyu Wei, and Duyu Tang.
\newblock Android in the zoo: Chain-of-action-thought for gui agents.
\newblock {\em arXiv preprint arXiv:2403.02713}, 2024.

\bibitem{comprehensive_smartphone_agent}
Xinbei Ma, Zhuosheng Zhang, and Hai Zhao.
\newblock Comprehensive cognitive llm agent for smartphone gui automation.
\newblock {\em arXiv preprint arXiv:2402.11941}, 2024.

\bibitem{aitw}
Christopher Rawles, Alice Li, Daniel Rodriguez, Oriana Riva, and Timothy Lillicrap.
\newblock Android in the wild: A large-scale dataset for android device control.
\newblock {\em arXiv preprint arXiv:2307.10088}, 2023.

\bibitem{empowering}
Hao Wen, Yuanchun Li, Guohong Liu, Shanhui Zhao, Tao Yu, Toby Jia-Jun Li, Shiqi Jiang, Yunhao Liu, Yaqin Zhang, and Yunxin Liu.
\newblock Empowering llm to use smartphone for intelligent task automation.
\newblock {\em arXiv preprint arXiv:2308.15272}, 2023.

\bibitem{openagents}
Tianbao Xie, Fan Zhou, Zhoujun Cheng, Peng Shi, Luoxuan Weng, Yitao Liu, Toh~Jing Hua, Junning Zhao, Qian Liu, Che Liu, et~al.
\newblock Openagents: An open platform for language agents in the wild.
\newblock {\em arXiv preprint arXiv:2310.10634}, 2023.

\bibitem{agentstudio}
Longtao Zheng, Zhiyuan Huang, Zhenghai Xue, Xinrun Wang, Bo~An, and Shuicheng Yan.
\newblock Agentstudio: A toolkit for building general virtual agents.
\newblock {\em arXiv preprint arXiv:2403.17918}, 2024.

\bibitem{pixel2act}
Peter Shaw, Mandar Joshi, James Cohan, Jonathan Berant, Panupong Pasupat, Hexiang Hu, Urvashi Khandelwal, Kenton Lee, and Kristina~N Toutanova.
\newblock From pixels to ui actions: Learning to follow instructions via graphical user interfaces.
\newblock {\em Advances in Neural Information Processing Systems}, 36, 2024.

\bibitem{yao2022webshop}
Shunyu Yao, Howard Chen, John Yang, and Karthik Narasimhan.
\newblock Webshop: Towards scalable real-world web interaction with grounded language agents.
\newblock {\em Advances in Neural Information Processing Systems}, 35:20744--20757, 2022.

\bibitem{seeclick}
Kanzhi Cheng, Qiushi Sun, Yougang Chu, Fangzhi Xu, Yantao Li, Jianbing Zhang, and Zhiyong Wu.
\newblock Seeclick: Harnessing gui grounding for advanced visual gui agents.
\newblock {\em arXiv preprint arXiv:2401.10935}, 2024.

\bibitem{pixelhelp}
Yang Li, Jiacong He, Xin Zhou, Yuan Zhang, and Jason Baldridge.
\newblock Mapping natural language instructions to mobile ui action sequences.
\newblock {\em arXiv preprint arXiv:2005.03776}, 2020.

\bibitem{assistgui}
Difei Gao, Lei Ji, Zechen Bai, Mingyu Ouyang, Peiran Li, Dongxing Mao, Qinchen Wu, Weichen Zhang, Peiyi Wang, Xiangwu Guo, et~al.
\newblock Assistgui: Task-oriented desktop graphical user interface automation.
\newblock {\em arXiv preprint arXiv:2312.13108}, 2023.

\bibitem{osworld}
Tianbao Xie, Danyang Zhang, Jixuan Chen, Xiaochuan Li, Siheng Zhao, Ruisheng Cao, Toh~Jing Hua, Zhoujun Cheng, Dongchan Shin, Fangyu Lei, et~al.
\newblock Osworld: Benchmarking multimodal agents for open-ended tasks in real computer environments.
\newblock {\em arXiv preprint arXiv:2404.07972}, 2024.

\bibitem{vwebarena}
Jing~Yu Koh, Robert Lo, Lawrence Jang, Vikram Duvvur, Ming~Chong Lim, Po-Yu Huang, Graham Neubig, Shuyan Zhou, Ruslan Salakhutdinov, and Daniel Fried.
\newblock Visualwebarena: Evaluating multimodal agents on realistic visual web tasks.
\newblock {\em arXiv preprint arXiv:2401.13649}, 2024.

\bibitem{androidenv}
Daniel Toyama, Philippe Hamel, Anita Gergely, Gheorghe Comanici, Amelia Glaese, Zafarali Ahmed, Tyler Jackson, Shibl Mourad, and Doina Precup.
\newblock Androidenv: A reinforcement learning platform for android.
\newblock {\em arXiv preprint arXiv:2105.13231}, 2021.

\bibitem{GUIautonomous}
Juyeon Yoon, Robert Feldt, and Shin Yoo.
\newblock Autonomous large language model agents enabling intent-driven mobile gui testing.
\newblock {\em arXiv preprint arXiv:2311.08649}, 2023.

\bibitem{chain_of_thought}
Jason Wei, Xuezhi Wang, Dale Schuurmans, Maarten Bosma, Fei Xia, Ed~Chi, Quoc~V Le, Denny Zhou, et~al.
\newblock Chain-of-thought prompting elicits reasoning in large language models.
\newblock {\em Advances in neural information processing systems}, 35:24824--24837, 2022.

\bibitem{react}
Shunyu Yao, Jeffrey Zhao, Dian Yu, Nan Du, Izhak Shafran, Karthik Narasimhan, and Yuan Cao.
\newblock React: Synergizing reasoning and acting in language models.
\newblock In {\em International Conference on Learning Representations (ICLR)}, 2023.

\bibitem{mmreact}
Zhengyuan Yang, Linjie Li, Jianfeng Wang, Kevin Lin, Ehsan Azarnasab, Faisal Ahmed, Zicheng Liu, Ce~Liu, Michael Zeng, and Lijuan Wang.
\newblock Mm-react: Prompting chatgpt for multimodal reasoning and action.
\newblock {\em arXiv preprint arXiv:2303.11381}, 2023.

\bibitem{assistgpt}
Difei Gao, Lei Ji, Luowei Zhou, Kevin~Qinghong Lin, Joya Chen, Zihan Fan, and Mike~Zheng Shou.
\newblock Assistgpt: A general multi-modal assistant that can plan, execute, inspect, and learn.
\newblock {\em arXiv preprint arXiv:2306.08640}, 2023.

\bibitem{screenai}
Gilles Baechler, Srinivas Sunkara, Maria Wang, Fedir Zubach, Hassan Mansoor, Vincent Etter, Victor C{\u{a}}rbune, Jason Lin, Jindong Chen, and Abhanshu Sharma.
\newblock Screenai: A vision-language model for ui and infographics understanding.
\newblock {\em arXiv preprint arXiv:2402.04615}, 2024.

\bibitem{image_llm_edit}
Xinyu Zhang, Mengxue Kang, Fei Wei, Shuang Xu, Yuhe Liu, and Lin Ma.
\newblock Tie: Revolutionizing text-based image editing for complex-prompt following and high-fidelity editing.
\newblock {\em arXiv preprint arXiv:2405.16803}, 2024.

\bibitem{actionbert}
Zecheng He, Srinivas Sunkara, Xiaoxue Zang, Ying Xu, Lijuan Liu, Nevan Wichers, Gabriel Schubiner, Ruby Lee, and Jindong Chen.
\newblock Actionbert: Leveraging user actions for semantic understanding of user interfaces.
\newblock In {\em Proceedings of the AAAI Conference on Artificial Intelligence}, volume~35, pages 5931--5938, 2021.

\bibitem{uibert}
Chongyang Bai, Xiaoxue Zang, Ying Xu, Srinivas Sunkara, Abhinav Rastogi, Jindong Chen, et~al.
\newblock Uibert: Learning generic multimodal representations for ui understanding.
\newblock {\em arXiv preprint arXiv:2107.13731}, 2021.

\bibitem{lexi}
Pratyay Banerjee, Shweti Mahajan, Kushal Arora, Chitta Baral, and Oriana Riva.
\newblock Lexi: Self-supervised learning of the ui language.
\newblock {\em arXiv preprint arXiv:2301.10165}, 2023.

\bibitem{gpt4vsom}
Jianwei Yang, Hao Zhang, Feng Li, Xueyan Zou, Chunyuan Li, and Jianfeng Gao.
\newblock Set-of-mark prompting unleashes extraordinary visual grounding in gpt-4v, 2023.

\bibitem{pix2struct}
Kenton Lee, Mandar Joshi, Iulia~Raluca Turc, Hexiang Hu, Fangyu Liu, Julian~Martin Eisenschlos, Urvashi Khandelwal, Peter Shaw, Ming-Wei Chang, and Kristina Toutanova.
\newblock Pix2struct: Screenshot parsing as pretraining for visual language understanding.
\newblock In {\em International Conference on Machine Learning}, pages 18893--18912. PMLR, 2023.

\bibitem{cogagent}
Wenyi Hong, Weihan Wang, Qingsong Lv, Jiazheng Xu, Wenmeng Yu, Junhui Ji, Yan Wang, Zihan Wang, Yuxiao Dong, Ming Ding, et~al.
\newblock Cogagent: A visual language model for gui agents.
\newblock {\em arXiv preprint arXiv:2312.08914}, 2023.

\bibitem{you2024ferret}
Keen You, Haotian Zhang, Eldon Schoop, Floris Weers, Amanda Swearngin, Jeffrey Nichols, Yinfei Yang, and Zhe Gan.
\newblock Ferret-ui: Grounded mobile ui understanding with multimodal llms.
\newblock {\em arXiv preprint arXiv:2404.05719}, 2024.

\bibitem{gpt4report}
Josh Achiam, Steven Adler, Sandhini Agarwal, Lama Ahmad, Ilge Akkaya, Florencia~Leoni Aleman, Diogo Almeida, Janko Altenschmidt, Sam Altman, Shyamal Anadkat, et~al.
\newblock Gpt-4 technical report.
\newblock {\em arXiv preprint arXiv:2303.08774}, 2023.

\bibitem{mmvet}
Weihao Yu, Zhengyuan Yang, Linjie Li, Jianfeng Wang, Kevin Lin, Zicheng Liu, Xinchao Wang, and Lijuan Wang.
\newblock Mm-vet: Evaluating large multimodal models for integrated capabilities.
\newblock {\em arXiv preprint arXiv:2308.02490}, 2023.

\bibitem{pyautogui}
PyAutoGUI.
\newblock Pyautogui.
\newblock 2024.
\newblock \url{https://pyautogui.readthedocs.io/en/latest/}.

\bibitem{claude}
AI~Anthropic.
\newblock The claude 3 model family: Opus, sonnet, haiku.
\newblock {\em Claude-3 Model Card}, 2024.

\bibitem{gemini}
Gemini Team, Rohan Anil, Sebastian Borgeaud, Yonghui Wu, Jean-Baptiste Alayrac, Jiahui Yu, Radu Soricut, Johan Schalkwyk, Andrew~M Dai, Anja Hauth, et~al.
\newblock Gemini: a family of highly capable multimodal models.
\newblock {\em arXiv preprint arXiv:2312.11805}, 2023.

\bibitem{Qwen_technicalReport}
Jinze Bai, Shuai Bai, Yunfei Chu, Zeyu Cui, Kai Dang, Xiaodong Deng, Yang Fan, Wenbin Ge, Yu~Han, Fei Huang, et~al.
\newblock Qwen technical report.
\newblock {\em arXiv preprint arXiv:2309.16609}, 2023.

\bibitem{chatgpt}
OpenAI.
\newblock Introducing chatgpt.
\newblock OpenAI Blog, 09 2021.

\bibitem{llama3}
{Meta}.
\newblock Introducing meta llama 3: The most capable openly available llm to date, 2024.
\newblock Accessed: 2024-04-18.

\bibitem{mixtral}
Albert~Q Jiang, Alexandre Sablayrolles, Antoine Roux, Arthur Mensch, Blanche Savary, Chris Bamford, Devendra~Singh Chaplot, Diego de~las Casas, Emma~Bou Hanna, Florian Bressand, et~al.
\newblock Mixtral of experts.
\newblock {\em arXiv preprint arXiv:2401.04088}, 2024.

\bibitem{vicuna2023}
Wei-Lin Chiang, Zhuohan Li, Zi~Lin, Ying Sheng, Zhanghao Wu, Hao Zhang, Lianmin Zheng, Siyuan Zhuang, Yonghao Zhuang, Joseph~E. Gonzalez, Ion Stoica, and Eric~P. Xing.
\newblock Vicuna: An open-source chatbot impressing gpt-4 with 90\%* chatgpt quality, March 2023.

\bibitem{azureocr}
Azure OCR.
\newblock Azure ocr.
\newblock 2024.
\newblock \url{https://azure.microsoft.com/en-us/products/ai-services/ai-vision}.

\bibitem{obs}
OBS Studio.
\newblock Obs studio.
\newblock 2024.
\newblock \url{https://obsproject.com/}.

\end{thebibliography}

\end{document}


\maketitle
\tableofcontents
\section{Experimental Settings}
\subsection{Data Collection Settings}
We use OBS Studio~\cite{obs} software to record the demonstration videos and capture the user's screenshots. Notably, in the screenshots, the user's cursor is not recorded, which is beneficial as the screenshots can be used directly without revealing the target coordinates. 
We use \texttt{pynput} to monitor detailed user activity metadata, such as click location $[x,y]$, typed content, and scroll distance.

\begin{figure}[h]
\centering
\includegraphics[width=\linewidth]{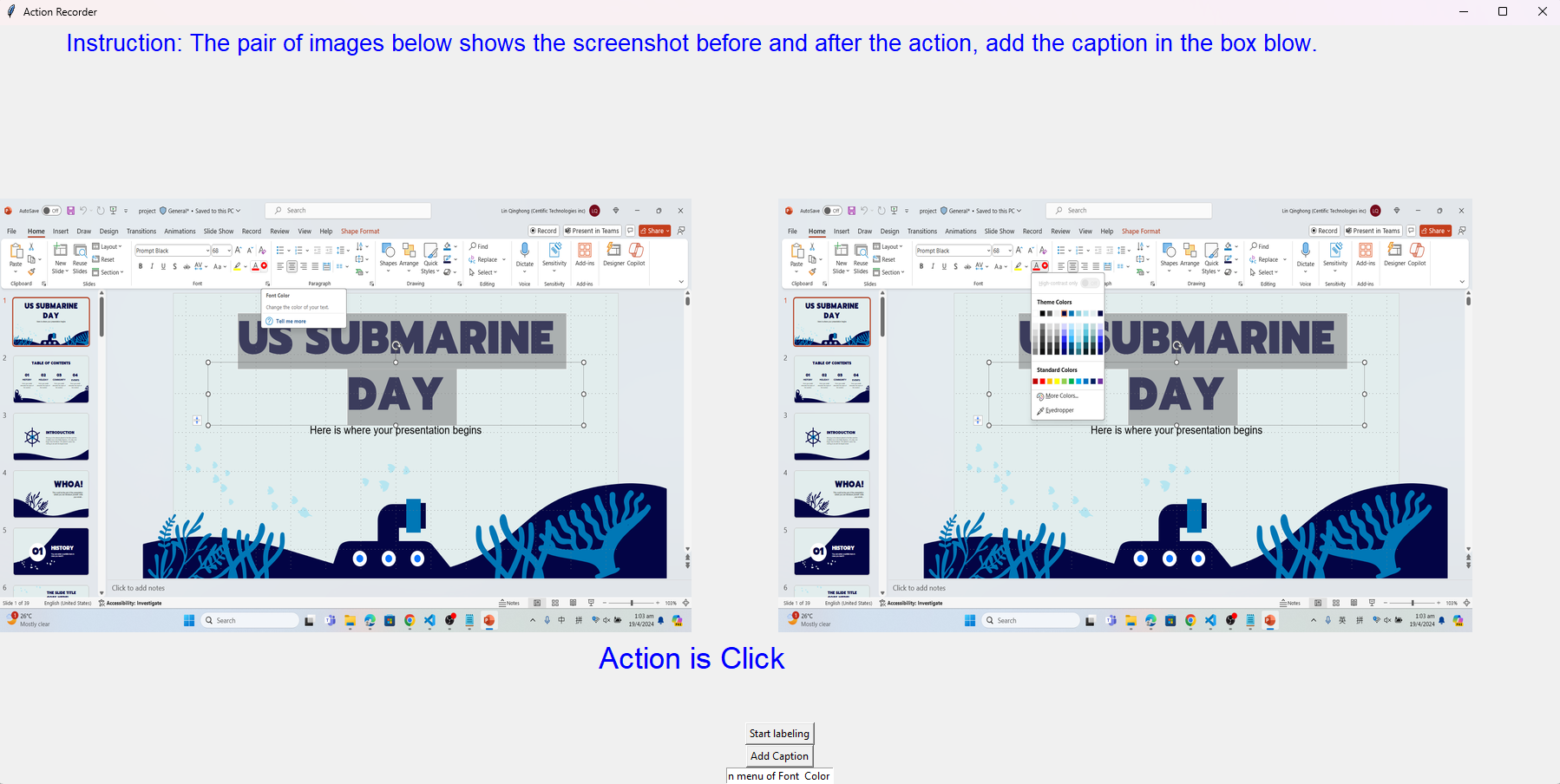}
\caption{Illustration of Manual annotation tools. The user are asked to watch their keyframe in their recording, and prompt to provide the element name regarding action.}
\label{supp:label_gui}
\end{figure}

In Fig.~\ref{supp:label_gui}, we display our manually labeled interface. Here, the annotator watches their key recording screenshots, with active regions such as the cursor coordinates highlighted in red. The annotators are then asked to enter the element name (e.g., "Drop-down menu of font color").

\subsection{Baseline Details}

\begin{tabular}{lll}
\toprule
Model & Ref. link & Version (\eg~model id)  \\
\midrule 
\llama & \href{https://deepinfra.com/}{\texttt{deepinfra}} & \texttt{meta-llama/Meta-Llama-3-70B-Instruct}  \\
\mixtral & \href{https://deepinfra.com/}{\texttt{deepinfra}} & \texttt{mistralai/Mixtral-8x22B-Instruct-v0.1} \\
\chatgpt & \href{https://platform.openai.com/docs/models/gpt-3-5-turbo}{\texttt{OpenAI}} & \texttt{gpt-3.5-turbo}  \\
\cogagent & \href{https://github.com/THUDM/CogVLM}{\texttt{CogAgent}} & \texttt{CogAgent-18B}  \\
\qwen & \href{https://help.aliyun.com/zh/dashscope/developer-reference/tongyi-qianwen-vl-plus-api?spm=a2c4g.11186623.0.0.645b7794Zi8mEy}{\texttt{Aliyun}} & \texttt{qwen-vl-max}  \\
\claude & \href{https://docs.anthropic.com/en/docs/models-overview}{\texttt{Anthropic}} & \texttt{claude-3-opus-20240229}  \\
\gemini & \href{https://deepmind.google/technologies/gemini/pro/}{\texttt{Google}} & \texttt{gemini-pro-vision}  \\
\gptv & \href{https://platform.openai.com/docs/models/gpt-4-turbo-and-gpt-4}{\texttt{OpenAI}} & \texttt{gpt-4-turbo}  \\
\gpto & \href{https://platform.openai.com/docs/models/gpt-4o}{\texttt{OpenAI}} & \texttt{gpt-4o}  \\
\bottomrule
\end{tabular}

\subsection{Evaluation Settings}
\textbf{Click.}
We detail how we calculate the distance metric. Assume we have a ground-truth point \([x_o, y_o]\) while the screenshot size is $H\times W$.

\begin{figure}[!h]
  \centering
  \begin{minipage}[b]{0.48\textwidth}
    \centering
    \includegraphics[width=\textwidth]{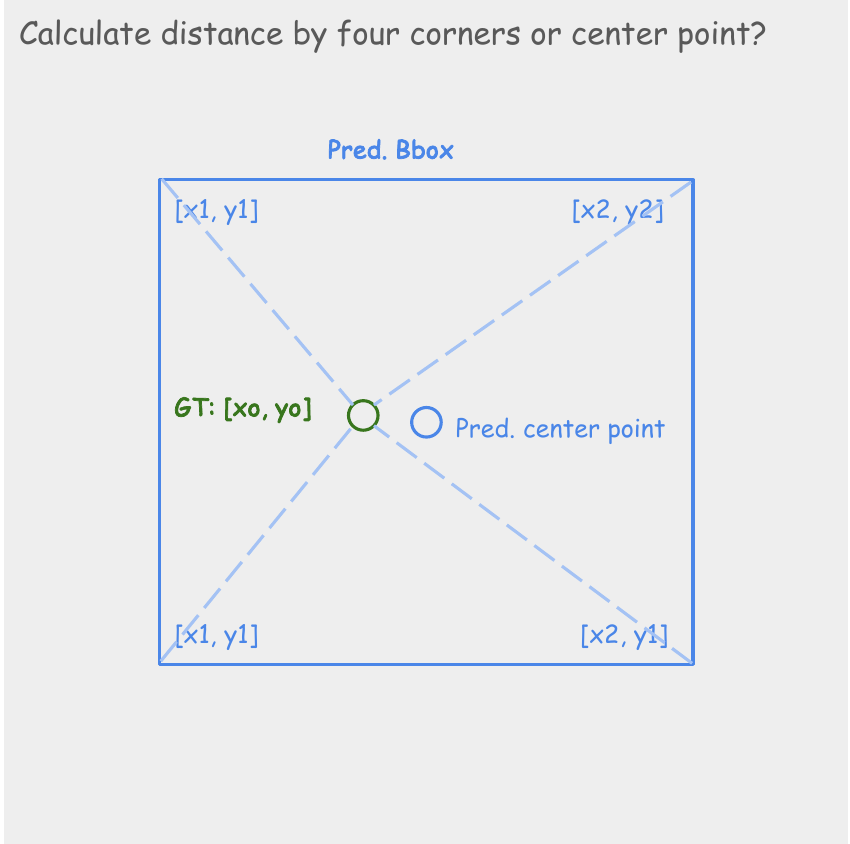}
    \label{fig:image1}
  \end{minipage}
  \hfill
  \begin{minipage}[b]{0.48\textwidth}
    \centering
    \includegraphics[width=\textwidth]{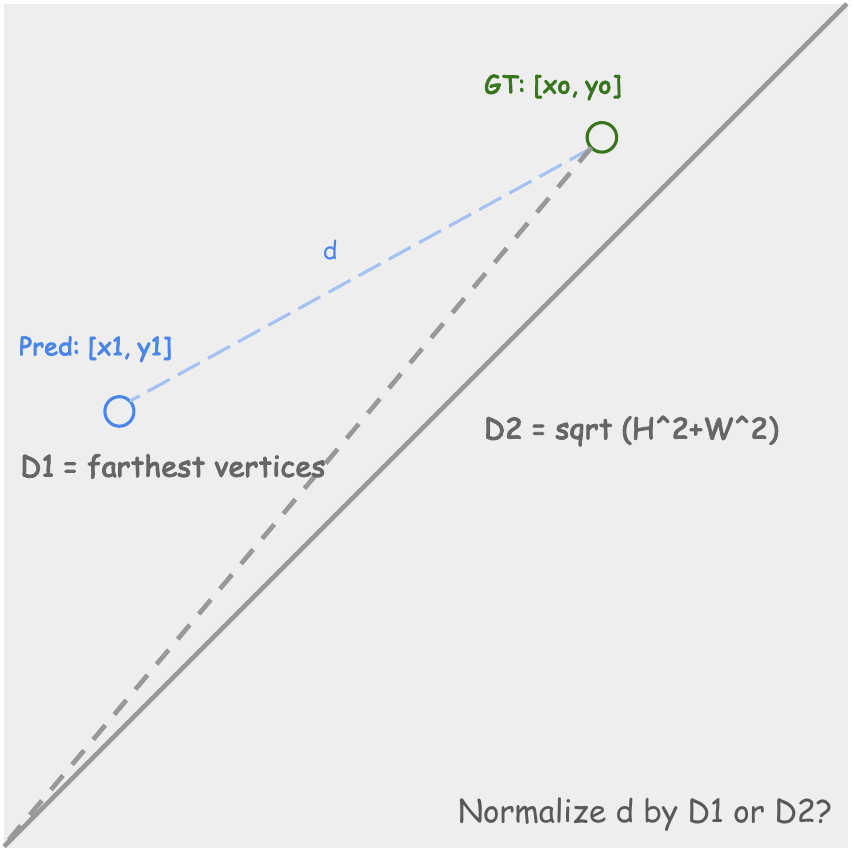}
    \label{fig:image2}
  \end{minipage}
  \caption{
  \textbf{(a) Illustration of why taking the distance btween the center point of a bounding box and groundtruth is not a proper measure of model performance on click.} As shown, the predicted bounding box center point is quite close to the ground-truth point, but the predicted bounding box area is large.
  \textbf{(b) Illustration of distance normalization.} To normalize the distance $d$ to $0 - 1$, a more proper term should be $D1$ (farthest vertices) rather than $D2$.
  }
  \label{fig:click:metric}
\end{figure}

$\bullet$ If the model prediction is a bounding box \([x_1, y_1, x_2, y_2]\) (\eg CogAgent \cite{cogagent} or Qwen-VL-Max \cite{Qwen_technicalReport}):

We cannot only take the center of the bounding box as the click target for evaluation because it does not account for the area of the bounding box. 
As illustrated in Fig.~\ref{fig:click:metric} (a), if the center point is very close to the ground truth but the bounding box cover a large area, the distance between the center point and the groundtruth would be small. 
Therefore, we design our metric to penalize for the area of the bounding box. Specifically, we calculate the distance between the ground truth and the four corners of the bounding box and then take the average. 
For the predicted bounding box, the average distance $d$ is calculated as follows: 
\begin{align*}
d &= \frac{1}{4} \left( \sqrt{(x_o - x_1)^2 + (y_o - y_1)^2} + \sqrt{(x_o - x_1)^2 + (y_o - y_2)^2} \right. \\
&\quad \left. + \sqrt{(x_o - x_2)^2 + (y_o - y_1)^2} + \sqrt{(x_o - x_2)^2 + (y_o - y_2)^2} \right)
\end{align*}
$\bullet$ If the model prediction is a coordinate \([x_1, y_1]\) (\eg as in GPT4V+SoM \cite{gpt4vsom}): 

We directly adopt the distance \(d\) calculated by:
    \[
    d = \sqrt{(x_o - x_1)^2 + (y_o - y_1)^2}
    \]

To normalize the pixel-level distance $d$ to $0-1$, a simple way is to divide $d$ by the maximum length in the screenshot, such as $\sqrt{H^2+W^2}$. But in practice, the maximum length should be the distance between the ground-truth point and the farthest vertices, so we use that for normalization. The comparison between the two normalization methods is illustrated in Fig.~\ref{fig:click:metric} (b).

\textbf{Drag.}
Drag is a combination of Clicks, so we simply adopt the click metric for the start and end point of drag, and take the average. The score is calculated as
$\textup{Dist}\coloneqq\frac{1}{2}\left(\frac{d_s}{D_s}+\frac{d_e}{D_e}\right)$ 
where $d_s$ is the pixel difference between predict start and GT start, while $D_s$ is the farthest vertices for the GT start; $d_e$ is the pixel difference between predict end and GT end, while $D_e$ is the farthest vertices for the GT end;

For Recall, it is calculated by:
\[ \textup{Recall}\left(\textup{start, end}\right) = 
\begin{cases} 
1 & {if}\quad\textup{Recall}\left(\textup{start}\right) \&~ \textup{Recall}\left(\textup{end}\right) \\ 
0 & {otherwise}
\end{cases}
\]

\begin{figure}[h]
\centering
\includegraphics[width=\linewidth]{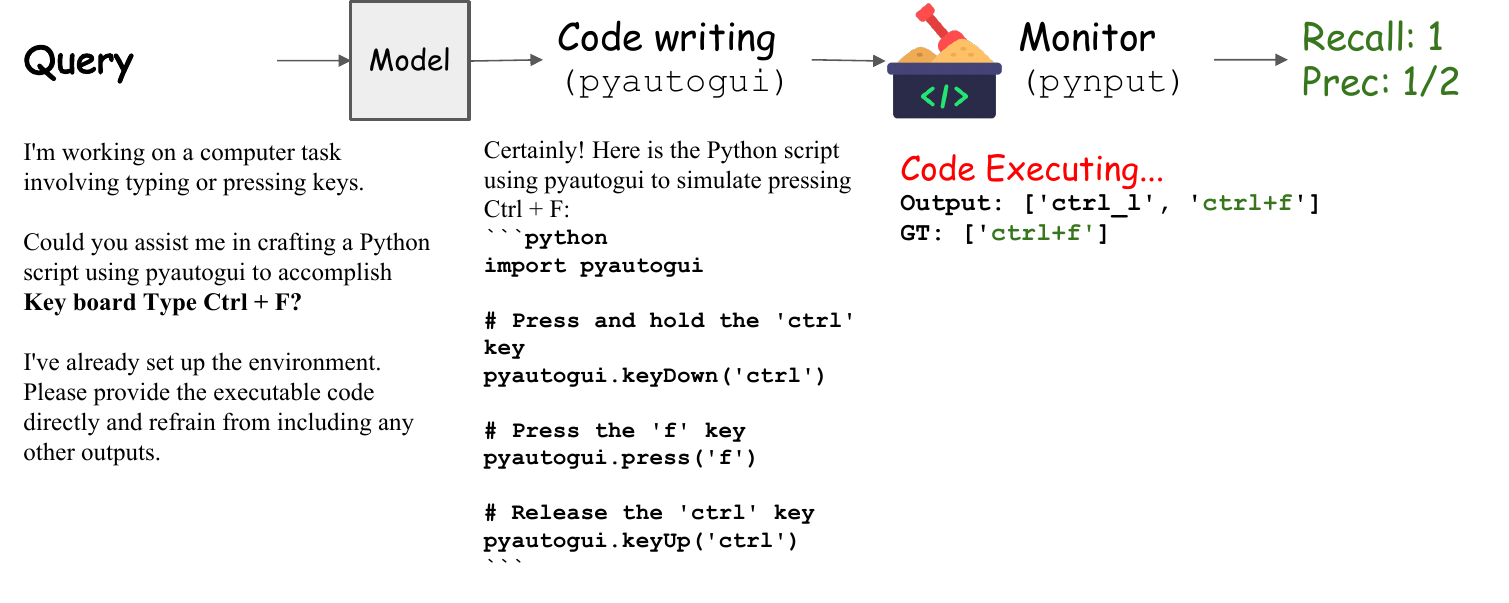}
\caption{Illustration of how we evaluate the key / press action.}
\label{supp:type}
\vspace{15pt}
\end{figure}

\textbf{Type / Press.}
For type/press, we evaluates whether the model can generate correct and efficient code to control keyboard activity. First, we prompt LLMs to write code for typing activity, and then we use \texttt{pynput} to monitor the keyboard outputs by executing the code. 
In Fig.~\ref{supp:type}, we show the pipeline for evaluating type/press activity. The model must generate the correct actions (e.g., \texttt{Ctrl+F}) with high precision, avoiding unnecessary actions such as redundant \texttt{Ctrl} presses.

\begin{figure}[h]
\centering
\includegraphics[width=\linewidth]{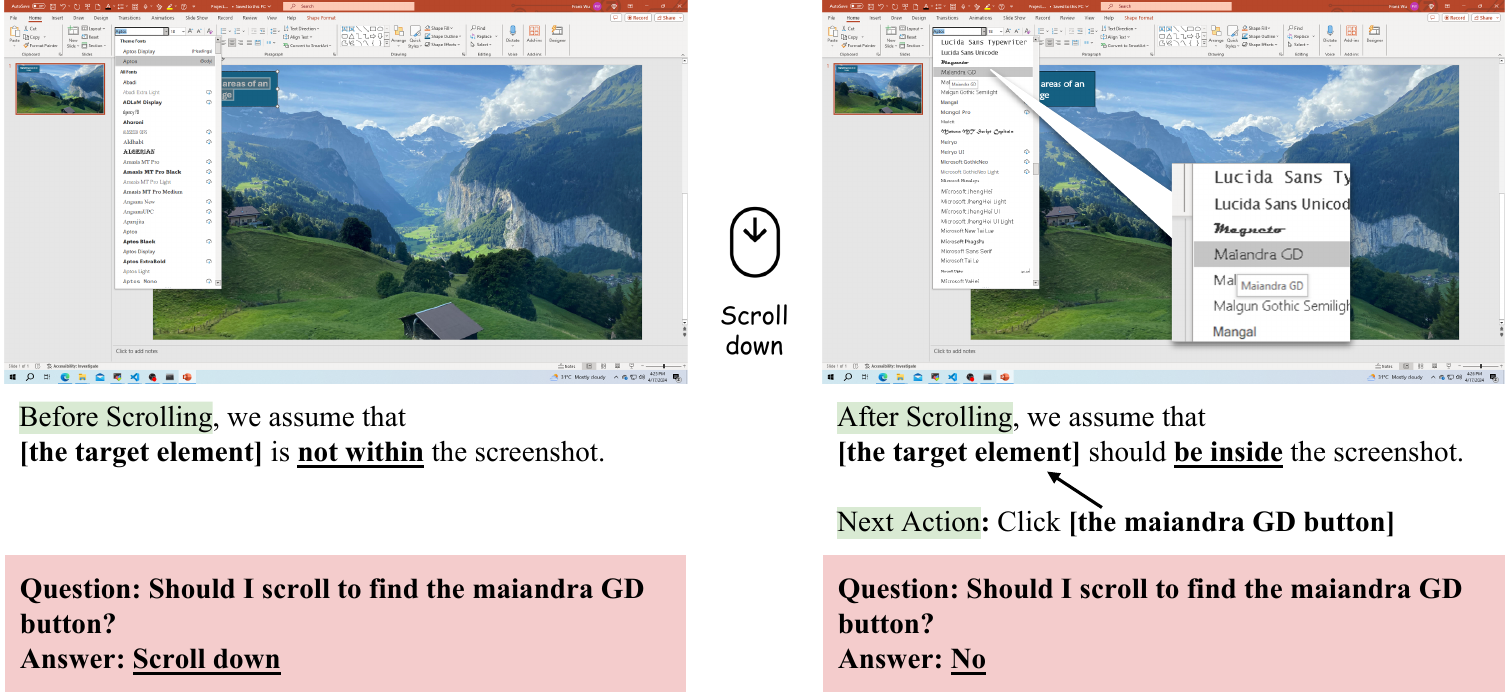}
\caption{Illustration of how we create the scroll QA pair.}
\label{supp:scroll}
\end{figure}

\textbf{Scroll.} 
Fig.~\ref{supp:scroll} illustrates how we construction QA pairs to evaluate on scroll action.
Before scrolling, the target element is assumed to be outside of the visible area, prompting for a scroll action. After scrolling, the target element is assumed to be within the visible area, ready for the next action (\eg Click shown in the figure). Thereby, we can construct the QA pairs under these assumptions.

For each scroll, we create two QA pairs with the following GT answers: ``scroll (up/down)'' for the screenshot before scrolling and ``no'' for the screenshot after scrolling. We randomly shuffle the order of answer options to make the final testing samples.
\clearpage
\section{Benchmark Statistics}

\textbf{Software distributions}
In Tab.~\ref{supp:software:app}, we present the software distribution on \our.

\begin{table}[!h]
\centering
\resizebox{1\textwidth}{!}{
\begin{tabular}{lc cc cc}
\toprule
Software & Platform & \# Full Task & \# Subtask & \makecell{\# Action per\\full task}  & \makecell{\# Action per\\subtask} \\
\midrule
\ppt & Windows & 8 & 52 & 47.6 & 8.5 \\
StableDiffusion & Web + Windows & 10 & 69 & 19.0 & 4.0 \\
Runway & Web & 11 & 63 & 24.3 & 4.7 \\
\midrule
\ps & Windows & 10 & 69 & 19.0 & 4.0 \\
After Effects & Windows & 13 & 67 & 29.3 & 7.2 \\
\pr & Windows & 7 & 38 & 15.4 & 4.5 \\
Capcut & Web + Windows & 10 & 46 & 9.4 & 3.6 \\
DaVinci & Windows & 11 & 44 & 18.8 & 4.7 \\
\midrule
YouTube & Web & 0 & 13 & 0 & 4.3 \\
Web Stock & Web & 0 & 12 & 0 & 9.7 \\
VLC player & Windows & 0 & 12 & 0 & 9.2 \\
\midrule
Total & -- & 82 & 463 &  23.7 & 5.8 \\
\bottomrule
\end{tabular}
}
\caption{\our's software distribution.}
\label{supp:software:app}
\end{table}

\begin{figure}[h]
    \centering
    \begin{subfigure}[b]{0.4\linewidth}
        \centering
        \includegraphics[width=\linewidth]{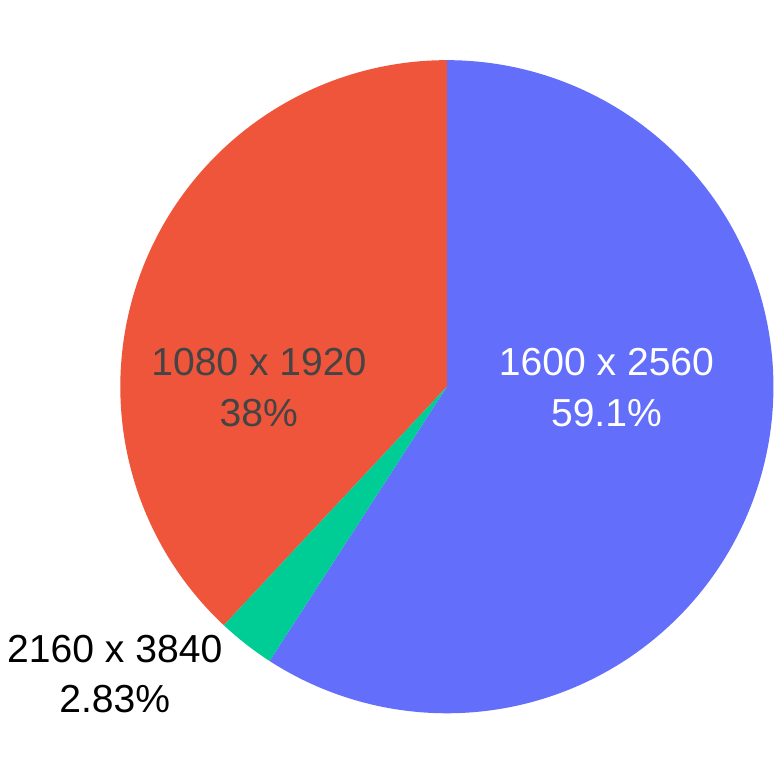}
        \caption{Screenshot resolution distribution.}
        \label{supp:res:pie}
    \end{subfigure}
    \hfill
    \begin{subfigure}[b]{0.4\linewidth}
        \centering
        \includegraphics[width=\linewidth]{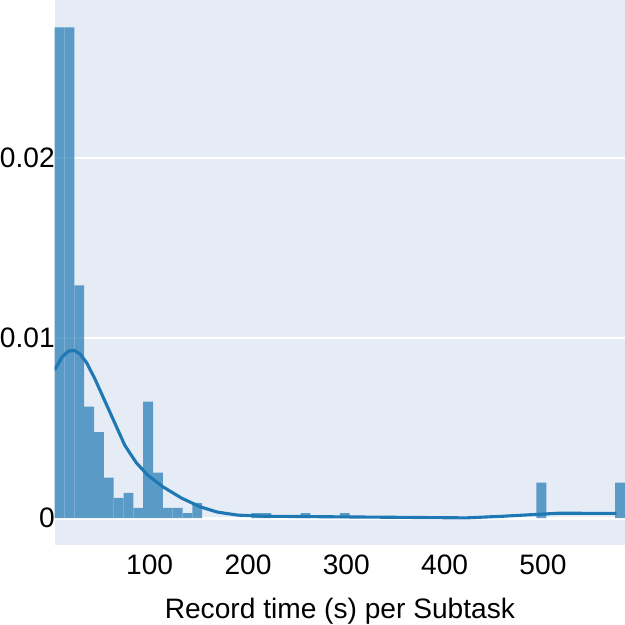}
        \caption{Recording duration per subtask.}
        \label{supp:record:dist}
    \end{subfigure}
    \vspace{0.5cm}
    \caption{Distribution of \textbf{(a) Screenshot resolution} and \textbf{(b) Human recording time.}}
\end{figure}

\textbf{Manual Recording Cost.}
In Fig.~\ref{supp:res:pie}, we present the screenshot resolution distribution primarily used for action execution.

\textbf{Screenshot's resolutions.}
In Fig.~\ref{supp:record:dist}, we present the distribution of manual recording time per subtask, with an average of 55 sec.

\begin{figure}[h]
	\centering
	\includegraphics[width=1.0\linewidth]{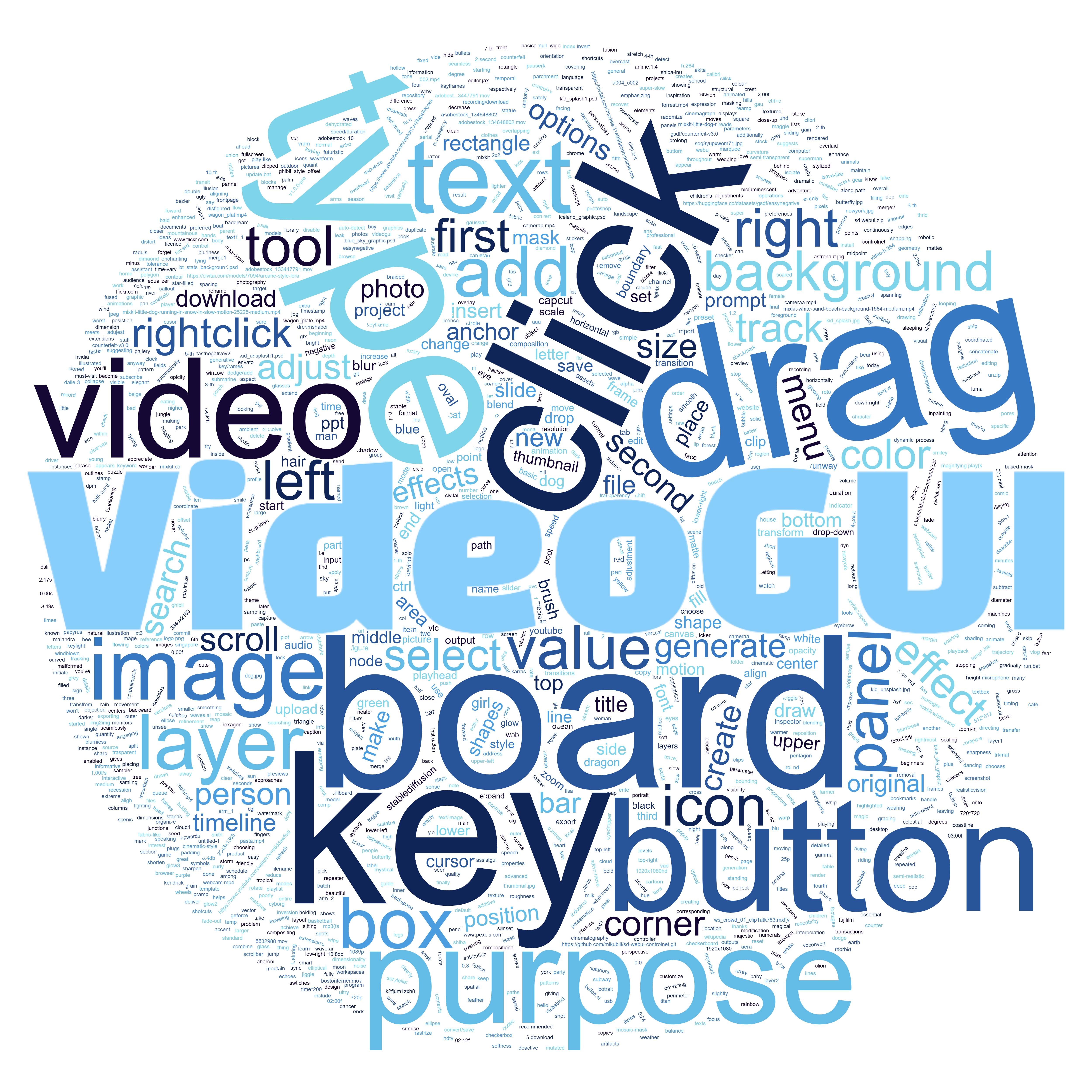}
    \caption{\our~World Clouds}
\label{fig:worldcloud}
\end{figure}

\textbf{World Cloud.}
In Fig.\ref{fig:worldcloud}, we present \our's Word Cloud, where the most frequent words are atomic actions (\eg~click, drag, type) and commonly used proper nouns (\eg, layer, background, pannel) in the GUI.

\clearpage
\section{Simulator Experiments}
\begin{figure}[h]
	\centering
	\includegraphics[width=1.0\linewidth]{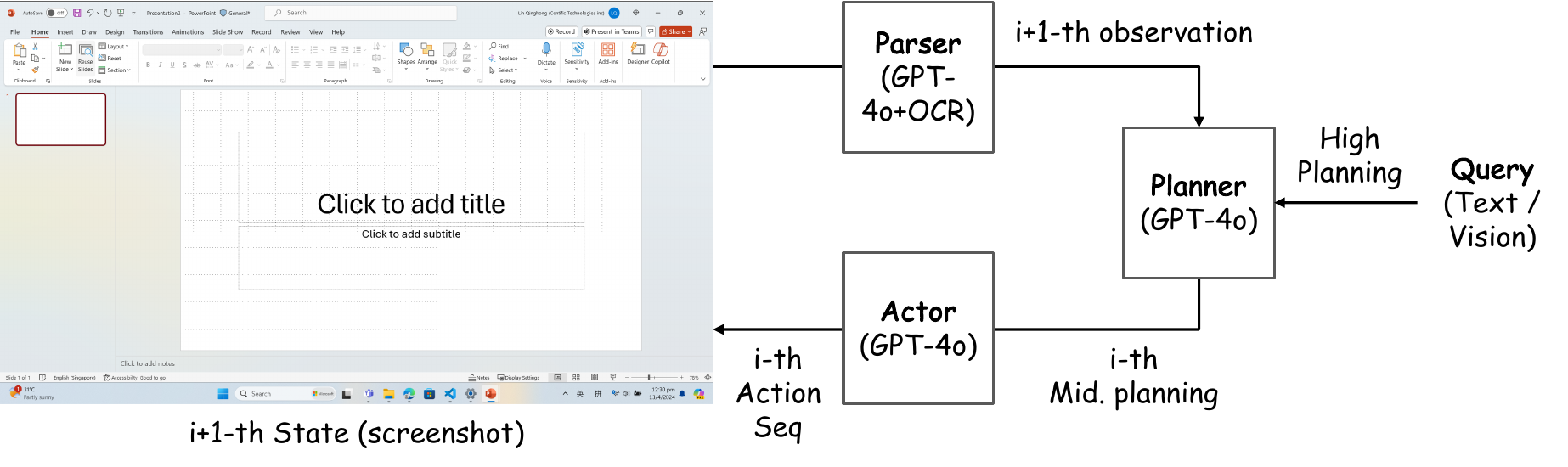}
    \caption{\textbf{Our Minimalist GUI Agent Framework} consists of three components: a Parser, a Planner, and an Actor. 
    The Planner receives input queries, which may be either vision previews or text instructions. It then conducts high-level planning and generates mid-level plans for the Actor. 
    The Actor executes these plans by performing a sequence of actions. 
    After action execution, the current state (screenshot) is captured and sent back to the Parser to gather observations. These observations are then relayed to the Planner for subsequent planning.}
\label{supp:miniagent}
\end{figure}

\textbf{Real-world Simulator.}
To simulate the real application scenario, we use the best performing LLM GPT-4o and build a simple agent baseline as shown in Fig.~\ref{supp:miniagent}.
We evaluate this agent on the most popular software (\ppt) to study its behavior.

\begin{table}[h]
\footnotesize
\centering
\resizebox{1\textwidth}{!}{
\begin{tabular}{ll ccc cc}
\toprule
\multirow{2}{*}{\textbf{Model}} & \multirow{2}{*}{\textbf{Settings}} & \multicolumn{3}{c}{\textbf{\our~Eval.}}& \multicolumn{2}{c}{\textbf{Full task Eval.}} \\
\cmidrule(lr){3-5} \cmidrule(lr){6-7} 
& & High Plan. & Mid Plan. & Action & Success Rate & {Rank (Arena)} $\downarrow$ \\
\midrule
\multirow{3}{*}{GUI Agent w/ GPT-4o~\cite{gpt4report}} 
 & Orig. Query (V) & 17.1 & 53.5 & 56.3 & 0 & 2.50 \\
 & w. GT High Plan. &  \color{gray}{100.0} & 53.5 & 56.3 & 0 & 1.88  \\
 & w. GT High \& Mid Plan. &  \color{gray}{100.0} & \color{gray}{100.0} & 56.3 & 0 &\textbf{1.38} \\
 \bottomrule
\end{tabular}
}
\caption{\textbf{Simulator Evaluation on \our's PPT \textit{full tasks}.}}
\label{supp:sim:fulltask}
\end{table}

Tab.~\ref{supp:sim:fulltask} presents the model performance on full task execution in our simulator environment. We see that completing the full task is extremely challenging for the GPT4o agent, with a notable 0 success rate for all variants. This again supports the design of our hierarchical evaluation, as the zero success rate simply implies the model/agent fail to execute the full task, without enough information in where they succeed or fail, or even how these models/agents perform relatively to each other.
Therefore, we introduce another metric, Rank (Arena), which compares the final outcome of their execution. Specifically, we ask human to perform manual inspection, and rank the comparing models by the similarities between the final results and the GT. 
We found that when injected with GT planning (both high or mid.-level), the full-task execution can be significantly improved. These results echoes our observations of low model performance in high-level and mid-level planning in the main paper, which are the bottlenecks of successful full-task executions.

\begin{figure}[t]
    \centering
    \includegraphics[width=1\textwidth]{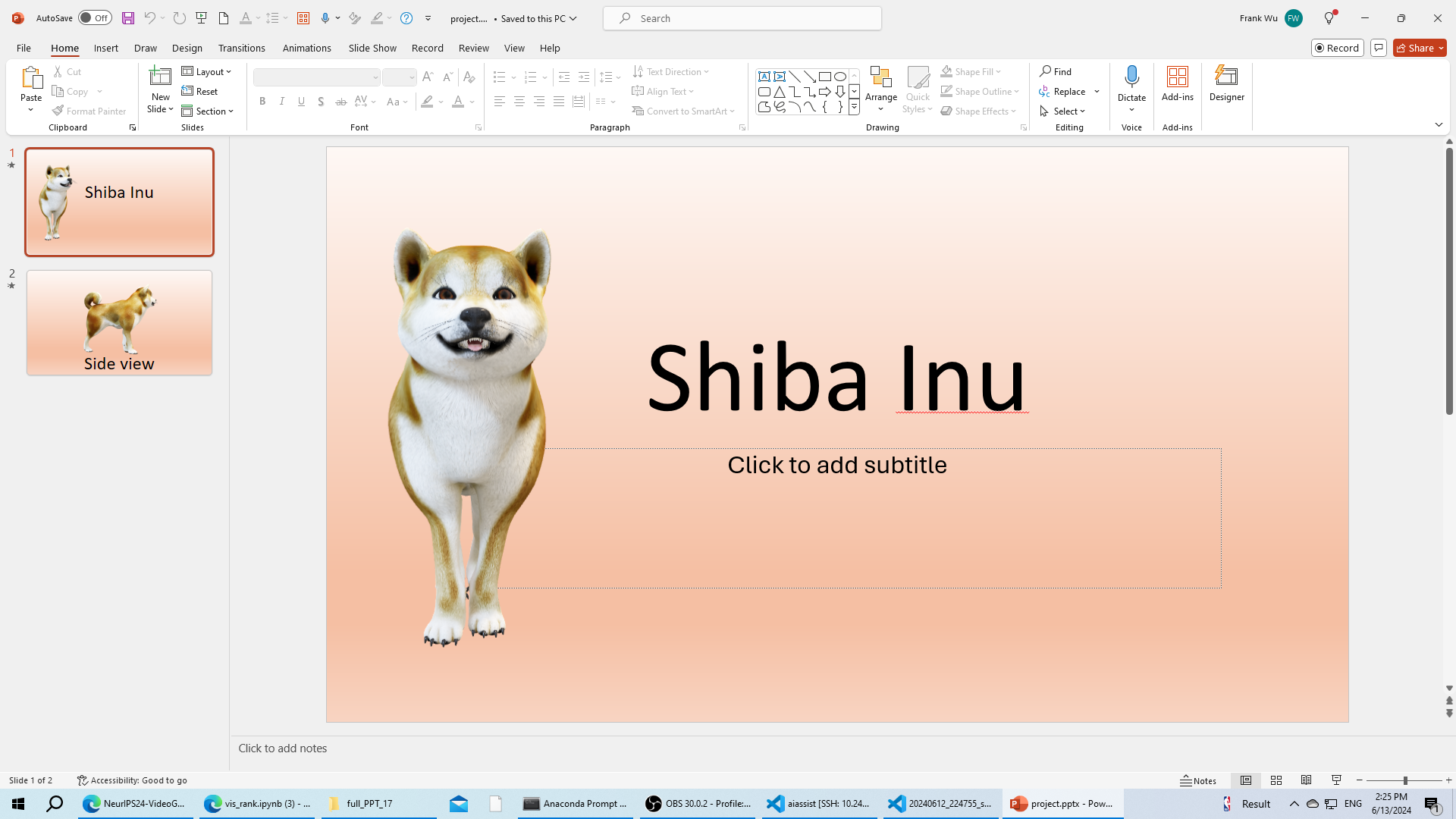}
    \caption{Final effect in \ppt~files.}
    \label{fig:ppt17_final}
    \vspace{2em}
    
    \begin{subfigure}[b]{0.32\textwidth}
        \centering
        \includegraphics[width=\textwidth]{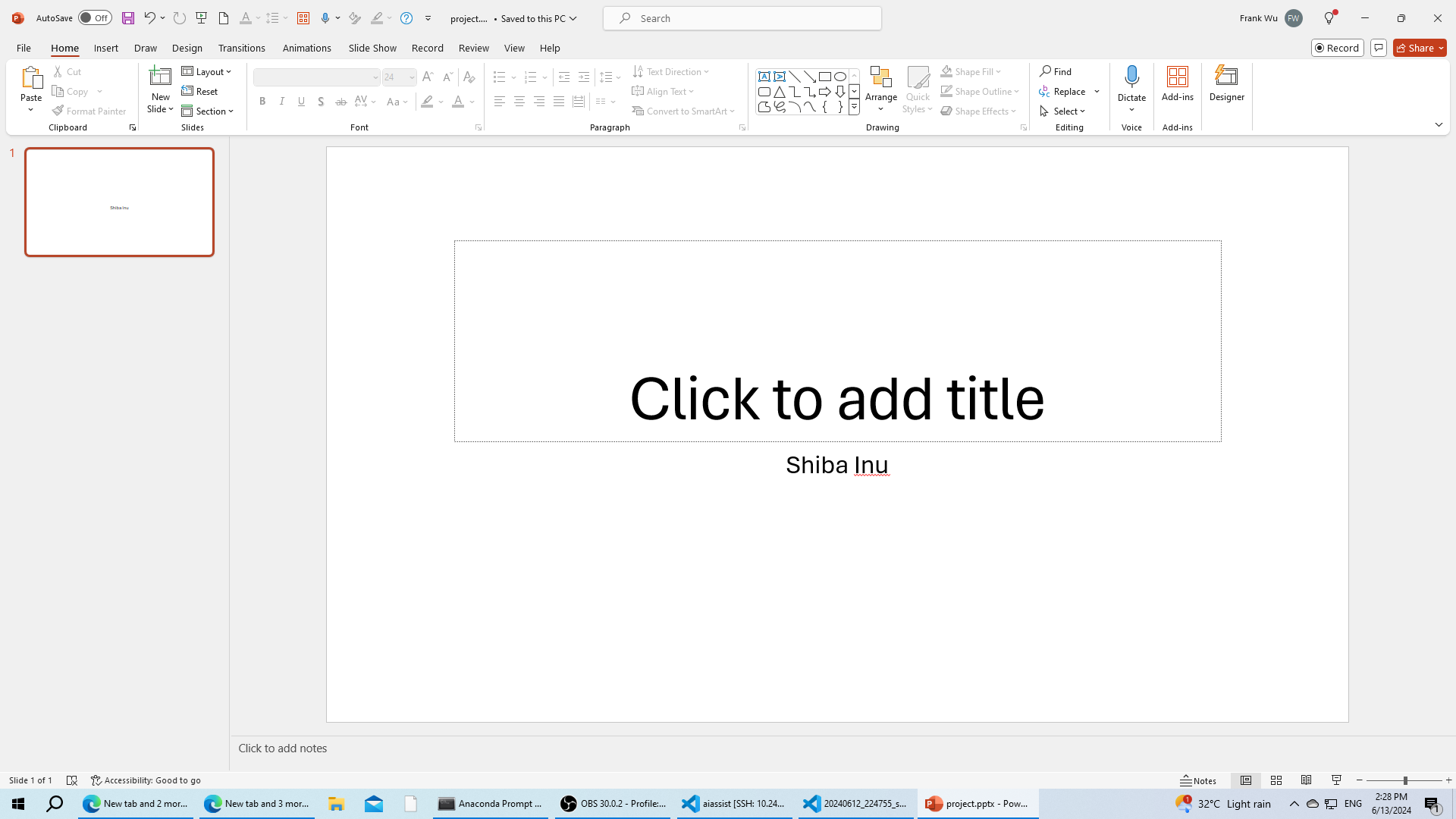}
        \caption{\textbf{GPT-4o}}
        \label{fig:ppt17_v}
    \end{subfigure}
    \hfill
    \begin{subfigure}[b]{0.32\textwidth}
        \centering
        \includegraphics[width=\textwidth]{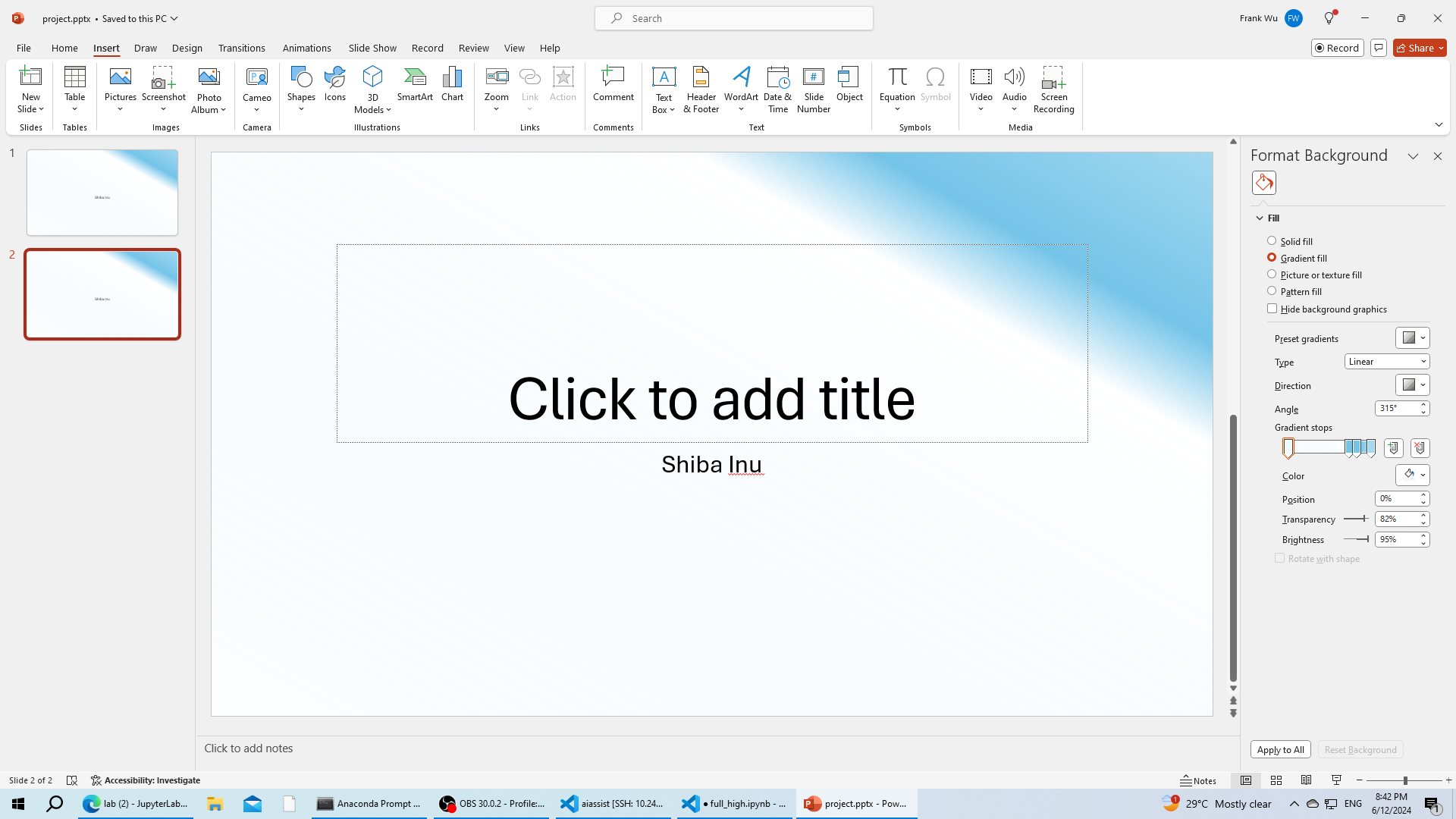}
        \caption{\textbf{GPT-4o w. GT High Plan}}
        \label{fig:ppt17_full}
    \end{subfigure}
    \hfill
    \begin{subfigure}[b]{0.32\textwidth}
        \centering
        \includegraphics[width=\textwidth]{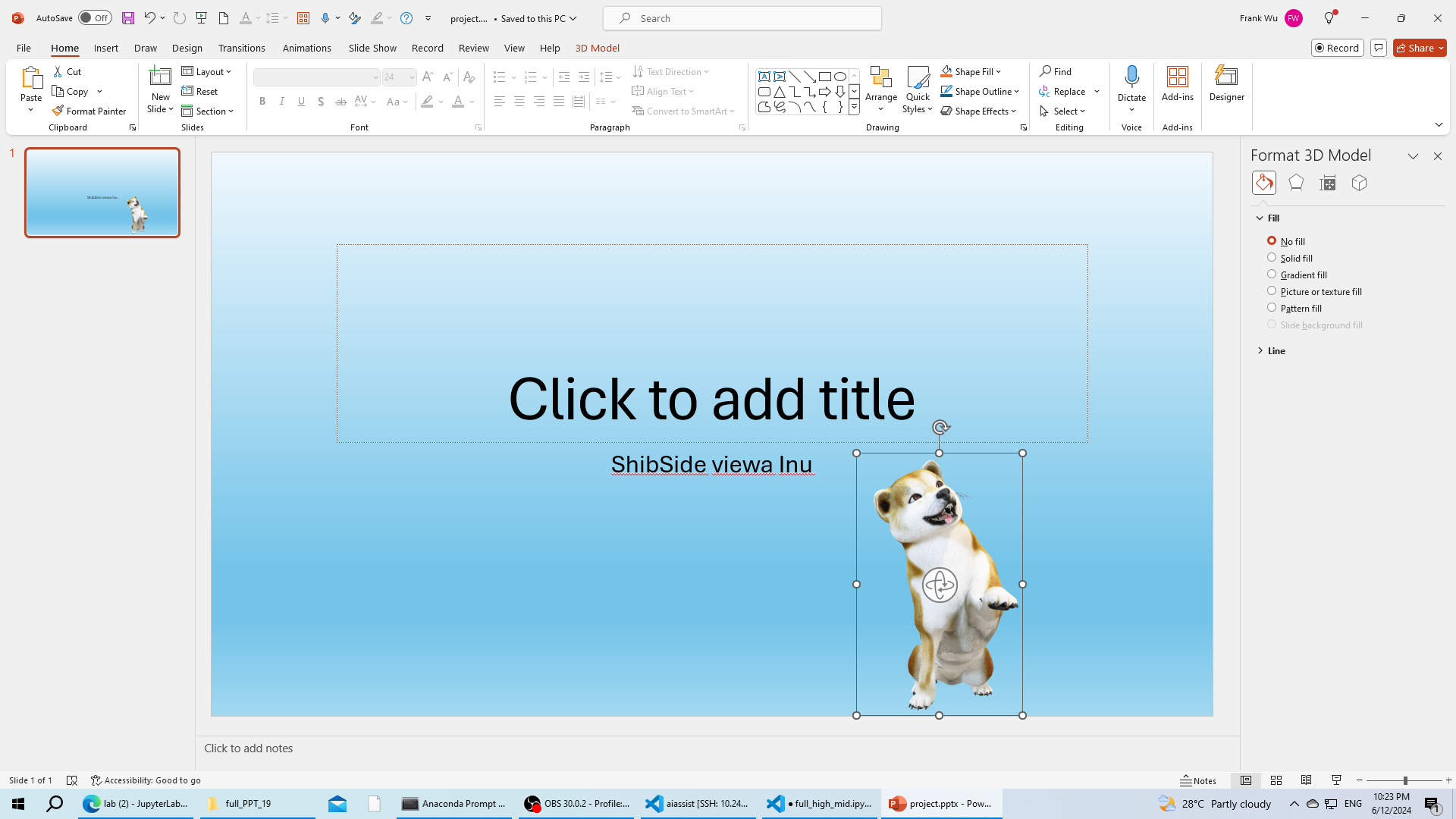}
        \caption{\textbf{GPT-4o w. GT High+Mid. Plan}}
        \label{fig:ppt17_full_mid}
    \end{subfigure}
    \vspace{2em}
    \caption{\textbf{Example of final outcome with our simple GPT-4o agent in simulated environment.} When provided with GT planning (c), the GUI agent successfully inserts the 3D model. However, it still fails to match the background color. }
    \label{fig:ppt17_sim}
\end{figure}

\begin{figure}[t]
    \centering
    \includegraphics[width=1\textwidth]{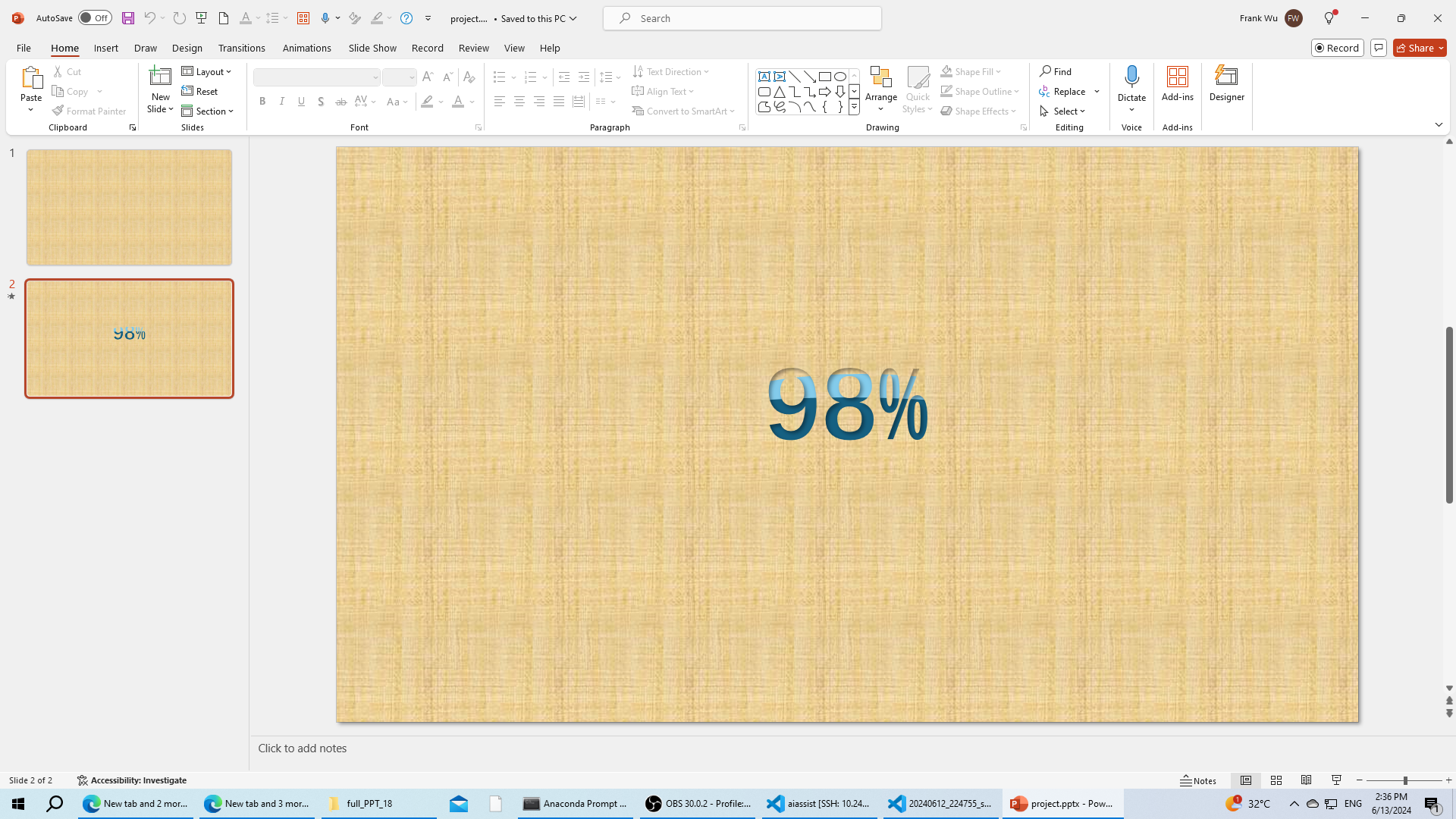}
    \caption{Final effect in \ppt~files.}
    \label{fig:ppt18_final}
    \vspace{2em}
    
    \begin{subfigure}[b]{0.32\textwidth}
        \centering
        \includegraphics[width=\textwidth]{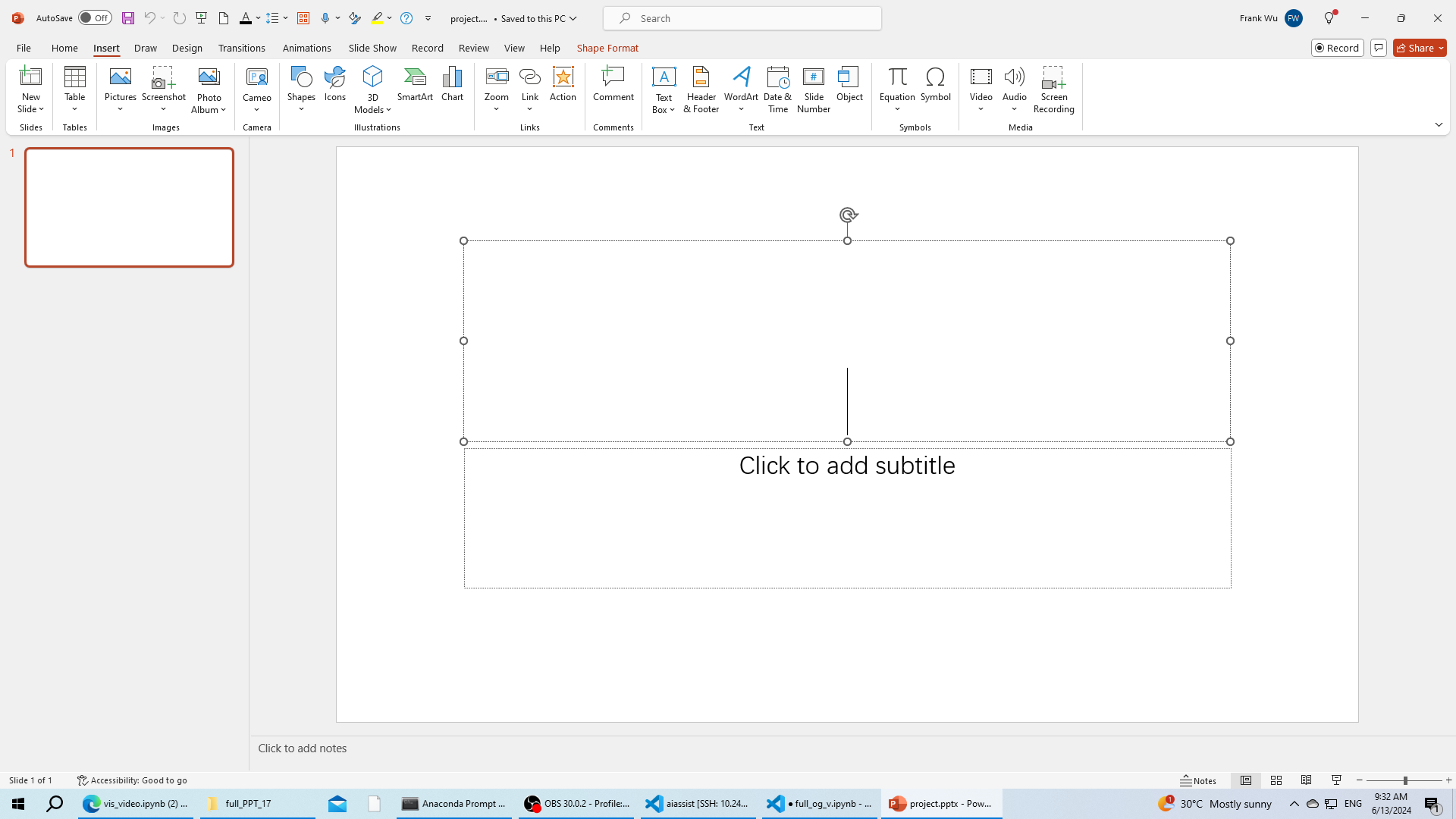}
        \caption{\textbf{GPT-4o}}
        \label{fig:ppt18_v}
    \end{subfigure}
    \hfill
    \begin{subfigure}[b]{0.32\textwidth}
        \centering
        \includegraphics[width=\textwidth]{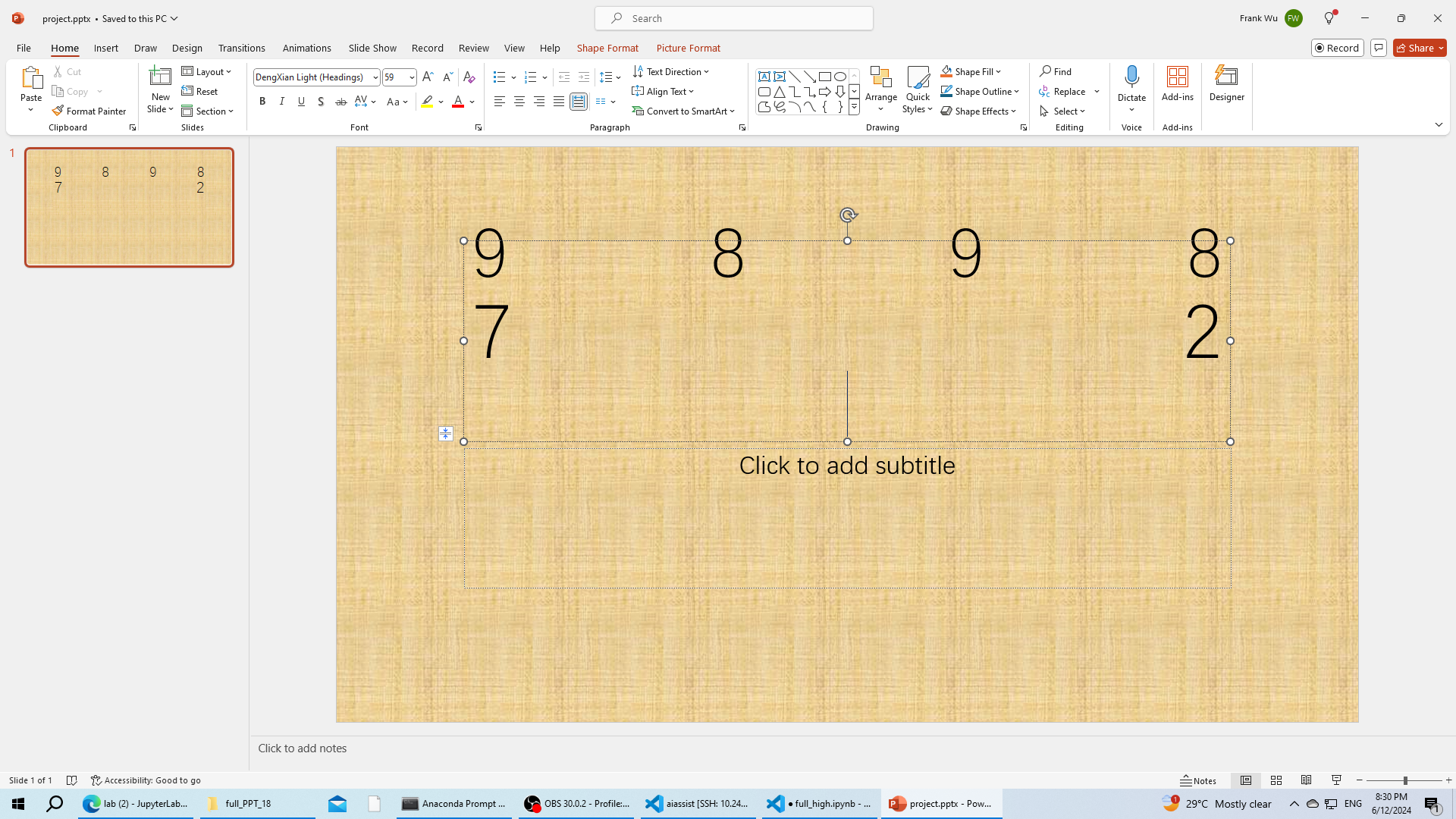}
        \caption{\textbf{GPT-4o w. GT High Plan}}
        \label{fig:ppt18_full_2}
    \end{subfigure}
    \hfill
    \begin{subfigure}[b]{0.32\textwidth}
        \centering
        \includegraphics[width=\textwidth]{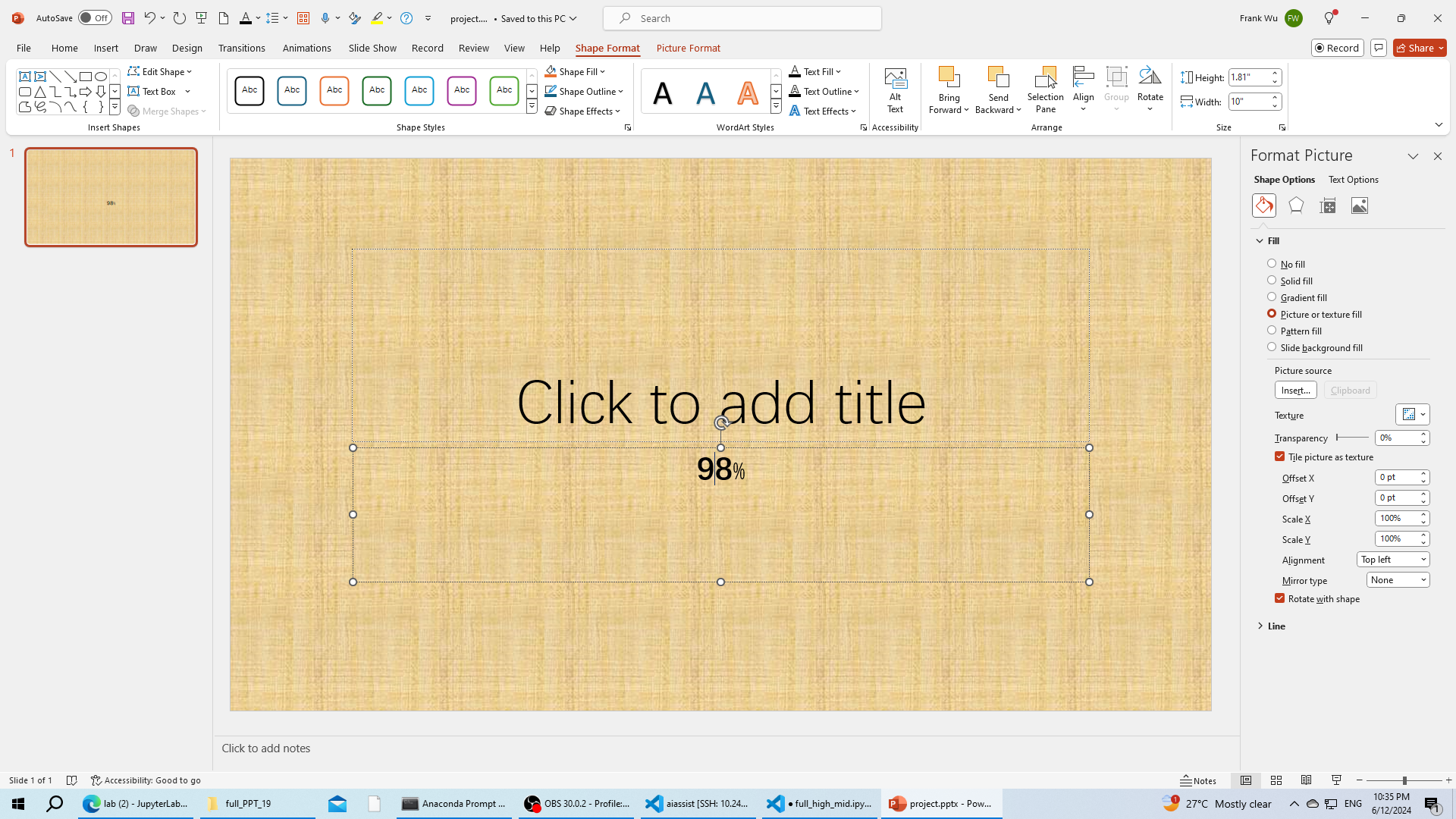}
        \caption{\textbf{GPT-4o w. GT High+Mid. Plan}}
        \label{fig:ppt18_full_mid_2}
    \end{subfigure}
    \vspace{2em}
    \caption{\textbf{Example of final outcome with our simple GPT-4o agent in simulated environment.} Guided by the GT  planning, both (b) and (c) successfully insert the textual background, while the (c) can accurately type `98\%'.}
    \label{fig:ppt18_sim}
\end{figure}

We visualize the final outcome of the three agent variants in Fig.~\ref{fig:ppt17_sim} and Fig.~\ref{fig:ppt18_sim}.


\begin{table}[h]
\footnotesize
\centering
\resizebox{1\textwidth}{!}{
\begin{tabular}{ll cc cc}
\toprule
\multirow{2}{*}{\textbf{Model}} & \multirow{2}{*}{\textbf{Settings}} & \multicolumn{2}{c}{\textbf{\our~Eval.}}& \multicolumn{2}{c}{\textbf{Subtask Eval.}} \\
\cmidrule(lr){3-4} \cmidrule(lr){5-6}
& & Mid Plan. & Action & Success Rate (\%) & Avg. Round $\downarrow$\\
\midrule
\multirow{2}{*}{GUI Agent w/ GPT-4o~\cite{gpt4report}} &  Orig. Query (V+T) &  53.5 & 56.3 & 20.0 & 5.4\\
 & w. GT Mid Plan. &  \color{gray}{100}  & 56.3 & \textbf{50.0} & \textcolor{gray}{\textbf{3.3}} \\
 \bottomrule
\end{tabular}
}
\caption{\textbf{Simulator Evaluation on \our's PPT \textit{subtasks}.}}
\label{supp:sim:subtask}
\end{table}

In Tab.~\ref{supp:sim:subtask}, we examine the performance of the GPT-4o agent in subtask competitions. Since subtasks do not necessitate high-level planning, we primarily investigate two variants: one with and one without manually provided middle-level planning, referred to as action sequences. Our study yields two key findings: 
(\textit{i}) Despite the simplicity of these tasks, the original GPT-4o agent achieves a success rate of only 20.0\%. With the assistance of manual plans, there is a 30\% increase in success rate. 
(\textit{ii}) For simple subtasks, the agent typically requires more extensive procedural execution compared to manual demonstrations (+2.1), which often represent the optimal pathway. This redundancy is exacerbated in complex tasks. Therefore, enhancing planning capabilities is essential for achieving efficient system with accurate success rates.
\clearpage
\section{Dataset Examples}

\textbf{Data samples.}
In this section, we display the visual-preview data samples, which are mainly focused on visual creation or editing.

\begin{table}[h]
    \centering
    \small
\begin{tabularx}{0.98\textwidth}{X|X|X|X}
        \toprule
        \multicolumn{4}{c}{\textbf{Visual preview}} \\
        \hline
        \multicolumn{4}{c}{\includegraphics[width=0.95\textwidth]{supp/datasample/ppt18_preview.pdf}} \\
        \midrule
        \footnotesize{\textbf{Full task}} &
        \footnotesize{\dothh} &
        \footnotesize{\dotmm} & \footnotesize{\dotaa}\\  
        \hline
{\scriptsize
\textbf{Visual query:} How to create this effect in \ppt?

\textcolor{gray}{\textbf{Textual query:} Create a slide that displays a large percentage figure of "98\%" against a textured, beige background that appears to be fabric or canvas. The numerals are rendered in a bold, stylized font. The visual effect in this image is a wave-like effect. The blue percentage numerals appear to be rising out of the beige fabric-like background, creating a dynamic appearance. This gradient of wave creates a sense of depth and dimensionality, making the wave appear to have volume and curvature. The lighter blue at the top catches the light more, giving an illusion of the wave crest rising up, while the darker blue below suggests shadow and recession.}

}
&
{\scriptsize
a. Format the background for the canvas

b. Change the background texture to parchment. Add a text box, add 98\%, increase the font size and bold effect

c. Change the background texture to papyrus, increase the font size of 98\%, change color to white, center it in the middle

d. Add a rectangle, remove outline, change the texture to papyrus

e. Send the rectangle to the back

\textcolor{darkyellow}{\textbf{f. Select the rectangle and the text. Merge shape and subtract, add buttom right shadow}}

g. Add shapes (e.g. Ovals) in between the two layers

h. Duplicate the slide, place it nicely and add Morph transition effect
}
&
{\scriptsize
f1. Drag to select the rectangle and text '98\%'

f2. Click on Shape Format button

f3. Click on Merge Shapes button

\textcolor{darkgreen}{\textbf{f4. Click on Subtract button}}

f5. Click on Presets button

f6. Click on shadow with buttom right

}
&
{\scriptsize
\textcolor{darkblue}{d1. Click, [322, 424]}

\includegraphics[width=0.2\textwidth]{supp/datasample/ppt18_action.pdf}
}
\\
\bottomrule
\end{tabularx}
\vspace{0.2em}
    \caption{Video Creation (\ie~animation) example with \textbf{\ppt.}}
    \label{ppt18}
\end{table}
\begin{table}[h]
    \centering
    \small
\begin{tabularx}{0.98\textwidth}{X|X|X|X}
        \toprule
        \multicolumn{4}{c}{\textbf{Visual preview}} \\
        \hline
        \multicolumn{4}{c}{\includegraphics[width=0.95\textwidth]{supp/datasample/pr10_preview.pdf}} \\
        \midrule
        \footnotesize{\textbf{Full task}} &
        \footnotesize{\dothh} &
        \footnotesize{\dotmm} & \footnotesize{\dotaa}\\  
        \hline
{\scriptsize
\textbf{Visual query:} How to transform from [start] to [end] in \pr?

\textcolor{gray}{\textbf{Textual query:} Add a rectangle mosaic mask to the red billboard and track it.}

}
&
{\scriptsize
\textbf{a.} Drag the timestamp to the beginning of the video

\textcolor{darkyellow}{\textbf{b. Add Mosaic effect on the top clip}}

\textbf{c.} Adjust the granularity of the Mosaic to 120

\textbf{d.} Add a rectangle mask to cover the bilboard and track it
}
&
{\scriptsize
{b1.} Click on Effects

{b2.} Click on Search box in Effects panel

{b3.} Key board Type Mosaic

\textcolor{darkgreen}{\textbf{b4. Click on 'Mosaic' effect}}

{b5.} Drag the Mosaic effect to the top clip.
}
&
{\scriptsize
\textcolor{darkblue}{b4. Click, [1667, 410]}

\includegraphics[width=0.2\textwidth]{supp/datasample/pr10_action.pdf}
}
\\
\bottomrule
\end{tabularx}
\vspace{0.2em}
    \caption{Video Editing example with \textbf{\pr.}}
    \label{pr10}
\end{table}

\begin{table}[h]
    \centering
    \small
\begin{tabularx}{0.98\textwidth}{X|X|X|X}
        \toprule
        \multicolumn{4}{c}{\textbf{Visual preview}} \\
        \hline
        \multicolumn{4}{c}{\includegraphics[width=0.95\textwidth]{supp/datasample/rw9_preview.pdf}} \\
        \midrule
        \footnotesize{\textbf{Full task}} &
        \footnotesize{\dothh} &
        \footnotesize{\dotmm} & \footnotesize{\dotaa}\\  
        \hline
{\scriptsize
\textbf{Visual query:} How to create this effect in Runway?

\textcolor{gray}{\textbf{Textual query:} Create a video about "A man in a dark green jacket stands in the center of a futuristic industrial setting with yellow machines and monitors, under bright overhead lights, creating a cinematic portrait effect" with the dolly zoom effect.}

}
&
{\scriptsize
\textbf{a.} Open Text/Image to Video Tool

\textbf{b.} Generate preview picture with text "A man in a dark green jacket stands in the center of a futuristic industrial setting with yellow machines and monitors, under bright overhead lights, creating a cinematic portrait effect.

\textbf{c.} Select the third image as the image input

\textcolor{darkyellow}{\textbf{d. Adjust camera settings. Set Zoom to -3}}

\textbf{e.} Select the background in Motion Brush. Set its Proximity to 10

\textbf{f.} Select the subject in Motion Brush. Set its Proximity to 2

\textbf{g.} Generate the video
}
&
{\scriptsize
\textcolor{darkgreen}{\textbf{d1. Click on Camera Settings.}}

d2. Click on the value of Zoom.

d3. Key board Type -3
}
&
{\scriptsize
\textcolor{darkblue}{d1. Click, [50, 840]}

\includegraphics[width=0.2\textwidth]{supp/datasample/rw9_action.pdf}
}
\\
\bottomrule
\end{tabularx}
\vspace{0.2em}
    \caption{Video Creation example with \textbf{Runway.}}
    \label{rw9}
\end{table}
\begin{table}[h]
    \centering
    \small
\begin{tabularx}{0.98\textwidth}{X|X|X|X}
        \toprule
        \multicolumn{4}{c}{\textbf{Visual preview}} \\
        \hline
        \multicolumn{4}{c}{\includegraphics[width=0.65\textwidth]{supp/datasample/sd10_preview.pdf}} \\
        \midrule
        \footnotesize{\textbf{Full task}} &
        \footnotesize{\dothh} &
        \footnotesize{\dotmm} & \footnotesize{\dotaa}\\  
        \hline
{\scriptsize
\textbf{Visual query:} How to transform from [start] to [end] in StableDiffusion-WebUI?

\textcolor{gray}{\textbf{Textual query:} Replace the 512*512 photo of a cat to a 720*720 photo of dragon by DPM++ method.}

}
&
{\scriptsize
\textbf{a.} Open img2img Tool and drag photo of cat into the file upload box

\textbf{b.} Put "image of a dragon" into prompt box

\textbf{c.} Put "cartoon" into negative prompt box

\textcolor{darkyellow}{\textbf{d. Set "Sampling method" to "DPM++ 2M Karras"}}

\textbf{e.} Set Width to 720 and Height to 720

\textbf{f.} Set Sampling steps to 25, Batch Size to 4 and CFG Scale to 4

\textbf{g.} Generate the image
}
&
{\scriptsize
d1. scroll down 7

d2. Click on options of Sampling method.

\textcolor{darkgreen}{\textbf{d3. Click on "DPM++ 2M".}}

d4. Click on options of Schedule type.

d5. Click on Karras.
}
&
{\scriptsize
\textcolor{darkblue}{d3. Click, [229, 277]}

\includegraphics[width=0.2\textwidth]{supp/datasample/sd10_action.pdf}
}
\\
\bottomrule
\end{tabularx}
\vspace{0.2em}
    \caption{Image Editing example with \textbf{StableDiffusion-WebUI.}}
    \label{sd10}
\end{table}

\begin{table}[h]
    \centering
    \small
\begin{tabularx}{0.98\textwidth}{X|X|X|X}
        \toprule
        \multicolumn{4}{c}{\textbf{Visual preview}} \\
        \hline
        \multicolumn{4}{c}{\includegraphics[width=0.95\textwidth]{supp/datasample/ae18_preview.pdf}} \\
        \midrule
        \footnotesize{\textbf{Full task}} &
        \footnotesize{\dothh} &
        \footnotesize{\dotmm} & \footnotesize{\dotaa}\\  
        \hline
{\scriptsize
\textbf{Visual query:} How to transform from [start] to [end] in Adobe Effects?

\textcolor{gray}{\textbf{Textual query:} Isolate the dog with Green Screen.}

}
&
{\scriptsize
\textbf{a.} Select and apply Keylight effect to the BostonTerrier.mov layer

\textbf{b.} Use the eyedropper tool to select the green background

\textbf{c.} Adjust Keylight view mode to Screen Matte

\textbf{d.} Modify Screen Gain and Screen Balance parameters

\textcolor{darkyellow}{\textbf{e. Adjust Clip Black and Clip White parameters in Screen Matte}}

\textbf{f.} Switch view mode back to Final Result and hide background layer
}
&
{\scriptsize
e1. Click on Expand icon of Screen Matte

\textcolor{darkgreen}{\textbf{e2. Click on Parameter of Clip Black 0.0}}

e3. Key board Type 10

e4. Click on Parameter of Clip White 100.0

e5. Key board Type 85
}
&
{\scriptsize
\textcolor{darkblue}{e2. Click, [193, 401]}

\includegraphics[width=0.2\textwidth]{supp/datasample/ae18_action.pdf}
}
\\
\bottomrule
\end{tabularx}
\vspace{0.2em}
    \caption{Video Editing example with \textbf{Adobe Effects.}}
    \label{ae18}
\end{table}
\begin{table}[h]
    \centering
    \small
\begin{tabularx}{0.98\textwidth}{X|X|X|X}
        \toprule
        \multicolumn{4}{c}{\textbf{Visual preview}} \\
        \hline
        \multicolumn{4}{c}{\includegraphics[width=0.65\textwidth]{supp/datasample/ps9_preview.pdf}} \\
        \midrule
        \footnotesize{\textbf{Full task}} &
        \footnotesize{\dothh} &
        \footnotesize{\dotmm} & \footnotesize{\dotaa}\\  
        \hline
{\scriptsize
\textbf{Visual query:} How to transform from [start] to [end] in \ps?

\textcolor{gray}{\textbf{Textual query:} Use quick selection tool to put the pencil in the black background.}

}
&
{\scriptsize
\textcolor{darkyellow}{\textbf{a. Use quick selection tool to select the pencil}}

\textbf{b.} Create a mask

\textbf{c.} Create a solid black background layer

\textbf{d.} Refine the mask. Set the smooth to 8, Feather to 7 px, Contrast to 72\%, and Shift Edge to -3\%;
}
&
{\scriptsize
\textcolor{darkgreen}{\textbf{a1. RightClick on Quick Selection Tool.}}

a2. Click on Quick Selection Tool.

a3. Drag the orange pencil from  right to  left. (Purpose:  select the orange pencil)
}
&
{\scriptsize
\textcolor{darkblue}{b4. RightClick, [25, 271]}

\includegraphics[width=0.15\textwidth]{supp/datasample/ps9_action.pdf}
}
\\
\bottomrule
\end{tabularx}
\vspace{0.2em}
    \caption{Image Editing example with \textbf{\ps.}}
    \label{ps9}
\end{table}
\begin{table}[h]
    \centering
    \small
\begin{tabularx}{0.98\textwidth}{X|X|X|X}
        \toprule
        \multicolumn{4}{c}{\textbf{Visual preview}} \\
        \hline
        \multicolumn{4}{c}{\includegraphics[width=0.95\textwidth]{supp/datasample/dv15_preview.pdf}} \\
        \midrule
        \footnotesize{\textbf{Full task}} &
        \footnotesize{\dothh} &
        \footnotesize{\dotmm} & \footnotesize{\dotaa}\\  
        \hline
{\scriptsize
\textbf{Visual query:} How to transform from [start] to [end] in DaVinci?

\textcolor{gray}{\textbf{Textual query:} Use Depth Map to blur the background.}

}
&
{\scriptsize
\textcolor{darkyellow}{\textbf{a. Add a serial node with depth map}}

\textbf{b.} Add a serial node with lens blur

\textbf{c.} Connect nodes and inverse the depth map node

\textbf{d.} Disable Depth Map Preview
}
&
{\scriptsize
a1. Click on Color panel.

a2. Click on Effects.

a3. Click on Search bar in Effects panel.

a4. Key board Type Dep

a5. RightClick on the video node in node editor.

a6. Click on "Add Node > Add Serial".

\textcolor{darkgreen}{\textbf{a7. Drag Depth Map from  Effects panel to video node 02. (Purpose: add Depth Map to the video node 02)}}
}
&
{\scriptsize
\textcolor{darkblue}{a7. Drag, [2175, 305]$\rightarrow$[1758, 370]}

\includegraphics[width=0.2\textwidth]{supp/datasample/dv15_action.pdf}
}
\\
\bottomrule
\end{tabularx}
\vspace{0.2em}
    \caption{Video Editing example with \textbf{DaVinci.}}
    \label{dv15}
\end{table}
\begin{table}[h]
    \centering
    \small
\begin{tabularx}{0.98\textwidth}{X|X|X|X}
        \toprule
        \multicolumn{4}{c}{\textbf{Visual preview}} \\
        \hline
        \multicolumn{4}{c}{\includegraphics[width=0.95\textwidth]{supp/datasample/cc20_preview.pdf}} \\
        \midrule
        \footnotesize{\textbf{Full task}} &
        \footnotesize{\dothh} &
        \footnotesize{\dotmm} & \footnotesize{\dotaa}\\  
        \hline
{\scriptsize
\textbf{Visual query:} How to transform from [start] to [end] in CapCut?

\textcolor{gray}{\textbf{Textual query:} Add Stickers "Heart", Effects "Blur" and Filters "Glow" to the video.}

}
&
{\scriptsize
\textcolor{darkyellow}{\textbf{a. Add "Heart" Sticker to the video}}

\textbf{b.} Add "Blur" Effect to the video

\textbf{c.} Add "Glow" Filter to the video
}
&
{\scriptsize
a1. Click on Click on Stickers Tool.

\textcolor{darkgreen}{\textbf{a2. Drag "heart" from  Stickers Pool to  video track. (Purpose:  add "heart" to the video track)}}

}
&
{\scriptsize
\textcolor{darkblue}{a2. Drag, [599, 464]$\rightarrow$[265, 1197]}

\includegraphics[width=0.2\textwidth]{supp/datasample/cc20_action.pdf}
}
\\
\bottomrule
\end{tabularx}
\vspace{0.2em}
    \caption{Video Editing example with \textbf{CapCut.}}
    \label{cc20}
\end{table}

\clearpage
\section{Prompts Templates}
\textbf{Procedural Planning.}
In Tab.\ref{temp:high_plan} and Tab.\ref{temp:mid_plan}, we present the prompt templates for high-level and mid-level planning, respectively. These templates are conditioned on the query formulation, such as whether the start or end visual effects are provided, or paired with the textual query.

\begin{table}[!h]
\centering
\begin{tcolorbox}[colback=gray!10, colframe=black, boxrule=0.5mm, width=\textwidth, rounded corners, boxsep=0mm, left=2mm, right=2mm, top=2mm, bottom=2mm]
\begin{tabularx}{\textwidth}{X}

\texttt{def get\_high\_prompt(vis\_start=True, vis\_end=True, txt=None, software=None):}
\\\\
\quad\texttt{PROMPT =} f"You are a software assistant professional at \texttt{\{software\}}."

\\
\quad\texttt{if vis\_start and vis\_end:}
    \\
    \quad\texttt{\quad\quad PROMPT +=} "Given two sequence of image frames about the initial visual effect and the final visual effect"
\\
\quad\texttt{elif vis\_end:}
\\
    \quad\texttt{\quad\quad PROMPT +=} "Given a sequence of image frames about the final visual effect"
\\
\quad\texttt{else:}
\\
    \quad\texttt{\quad\quad PROMPT +=} " You are provided"

\\
\quad\texttt{if txt:}
\\
    \quad\texttt{\quad\quad PROMPT +=} " with a task textual description"

\\
\quad\texttt{PROMPT +=} " Your goal is to recognize the effect software demonstrates and pinpoint the key functions or operations, necessary to replicate this distinctive pattern."

\\
\quad\texttt{PROMPT +=} """\\
**High-Level Planning**: \\
Distill the process into essential stages or components, emphasizing the unique functions or operations, such as a specific design technique.
Concentrate on brevity and precision in describing each stage, highlighting the unique aspects that contribute to the overall effect.
\\\\
Please format your response as follows (we use Powerpoint as an example):
\\
```\\
1: Insert a Circle and Change its color as black.\\
2: Add Text 'Happy' inside the Circle.\\
3: Apply the 'Fly-in' animation for the Circle.\\
'''
\\\\
Each stage should be concise yet comprehensive, focusing on the key functionalities or operations that lead to the visual outcome in PowerPoint.
Notably, avoid detailed step-by-step actions. 
Strive to keep the number of stages as few as possible, only including those that are crucial for achieving the unique effect.\\
"""
\\\\
\quad\texttt{if txt:}\\
    \quad\texttt{\quad\quad PROMPT +=} f"**This is the textual descriptions** \texttt{\{txt\}}"

\quad\texttt{return PROMPT}
\end{tabularx}
\end{tcolorbox}
\caption{\textbf{High-level Planning Prompt} conditioned on the interleaved instruction query.}
\label{temp:high_plan}
\end{table}

\begin{table}[!h]
\centering
\begin{tcolorbox}[colback=gray!10, colframe=black, boxrule=0.5mm, width=\textwidth, rounded corners, boxsep=0mm, left=2mm, right=2mm, top=2mm, bottom=2mm]
\begin{tabularx}{\textwidth}{X}

\texttt{def get\_prompt(vis=True, txt=None, software=None)}:
\\\\
    \quad\texttt{PROMPT =} f"You have been assigned the task of planning a sequence of actions in \texttt{\{software\}} software to achieve a desired goal state based on certain conditions. \
    Your objective is to outline the fundamental actions needed.\n\n"

    \\
    \texttt{\quad if vis and not txt:}\\
        \quad\texttt{\quad\quad PROMPT +=} "**You are provided with two screenshots which indicate the initial state as well as goal state.**\n"
    \\\\
    \texttt{\quad elif vis and txt:}\\
        \quad\texttt{\quad\quad PROMPT +=} "**You are provided with a screenshot to indicate your initial state.**\n"

    \\
    \texttt{\quad if txt:}\\
        \quad\texttt{\quad\quad PROMPT +=} f"**The goal is: \texttt{\{txt\}}**"
\\\\
    \texttt{\quad PROMPT +=} """\\
Please format your response as follows:
\\
```\\
1. Click the 'xxx'.\\
2. Type 'yyy'.\\
3.: Click the 'zzz'.\\
'''
\\
Ensure that each step is clearly described to facilitate step-by-step reproduction of the actions.
"""\\
\quad\texttt{return PROMPT}
\end{tabularx}
\end{tcolorbox}
\caption{\textbf{Middle-level Planning Prompt} conditioned on the interleaved instruction query.}
\label{temp:mid_plan}
\end{table}

\textbf{Action -- Click.}
In Tab.~\ref{temp:click}, we show the template used by LLM to estimate click coordinates based on image resolution. With SoM's assistance, we use the Tab.~\ref{temp:click:som} template to predict the mark index.
With OCR's assistance, we use the Tab.~\ref{temp:click:ocr} template.

\begin{table}[!h]
\centering
\begin{tcolorbox}[colback=gray!10, colframe=black, boxrule=0.5mm, width=\textwidth, rounded corners, boxsep=0mm, left=2mm, right=2mm, top=2mm, bottom=2mm]
\begin{tabularx}{\textwidth}{X}
I'm working on a computer task that involves clicking on some elements (like a button).\\
You are provided with a screenshot with a resolution of width: \texttt{\{width\}} and height: \texttt{\{height\}}.\\
Could you assist me in navigating to the "\texttt{\{element\}}"?\\
Please provide the location in the following format:\\
```
[x, y]
'''
\\
Ensure that your response contains only the coordinates.
\end{tabularx}
\end{tcolorbox}
\caption{\textbf{Click action template} that prompts LLMs output click's coordinate [x,y]}
\label{temp:click}
\end{table}

\begin{table}[!h]
\centering
\begin{tcolorbox}[colback=gray!10, colframe=black, boxrule=0.5mm, width=\textwidth, rounded corners, boxsep=0mm, left=2mm, right=2mm, top=2mm, bottom=2mm]
\begin{tabularx}{\textwidth}{X}
The screenshot has been divided into areas and marked with numbers. Where is \texttt{\{element\}}? Answer by mark index like [x].
\end{tabularx}
\end{tcolorbox}
\caption{\textbf{Click action template} that prompts LLMs~(with SoM~\cite{gpt4vsom}) output coordinate.}
\label{temp:click:som}
\end{table}

\begin{table}[!h]
\centering
\begin{tcolorbox}[colback=gray!10, colframe=black, boxrule=0.5mm, width=\textwidth, rounded corners, boxsep=0mm, left=2mm, right=2mm, top=2mm, bottom=2mm]
\begin{tabularx}{\textwidth}{X}
I'm working on a computer task that involves clicking on some elements (like a button).\\
Below are the OCR detection results (element name - bounding coordinates [[x1, y1], [x2, y2]]), which are separated by a colon ";".\\

\texttt{\{ocr\_result\}}
\\
Could you assist me in navigating to the "\texttt{\{element\}}"?\\
Please provide the location in the following format:\\
```
[x, y]
'''
\\
Ensure that your response contains only the coordinates.
\end{tabularx}
\end{tcolorbox}
\caption{\textbf{Click action template} that prompts LLMs (with OCR~\cite{azureocr}) output click's coordinate [x,y]}
\label{temp:click:ocr}
\end{table}

\textbf{Action -- Drag.}
In Tab.~\ref{temp:drag}, we show the template used by LLM to estimate drag coordinates based on image resolution. With SoM's assistance, we use the Tab.~\ref{temp:drag:som} template to predict the start and end mark index.
With OCR's assistance, we use the Tab.~\ref{temp:drag:ocr} template.

\begin{table}[h]
\centering
\begin{tcolorbox}[colback=gray!10, colframe=black, boxrule=0.5mm, width=\textwidth, rounded corners, boxsep=0mm, left=2mm, right=2mm, top=2mm, bottom=2mm]
\begin{tabularx}{\textwidth}{X}
I am working on a computer task that involves dragging elements from one place to another\\
You are provided with a screenshot with a resolution of width: \texttt{\{width\}} and height: \texttt{\{height\}}.\\
Could you assist me in navigating for action "\texttt{\{narration\}}"?\\
Please provide the location in the following format:\\
```
[x1, y1] -> [x2, y2]
'''\\
where [x1, y1] are the start coordinates and [x2, y2] are the destination coordinates.\\
Ensure that your response contains only the coordinates.\\
\end{tabularx}
\end{tcolorbox}
\caption{\textbf{Drag action template} that prompts LLMs output drag's coordinate [x1,y1] -> [x2, y2].}
\label{temp:drag}
\end{table}

\begin{table}[h]
\centering
\begin{tcolorbox}[colback=gray!10, colframe=black, boxrule=0.5mm, width=\textwidth, rounded corners, boxsep=0mm, left=2mm, right=2mm, top=2mm, bottom=2mm]
\begin{tabularx}{\textwidth}{X}
The screenshot has been divided into areas and marked with numbers. \\
To assist with dragging an item, please provide the start and end mark numbers.\\
How to \texttt{\{element\}}? Provide the mark indices as follows:\\
```
[x]->[y]
'''\\
where [x] represents the starting index and [y] represents the ending index.
\end{tabularx}
\end{tcolorbox}
\caption{\textbf{Drag action template} that prompts LLMs (with SoM~\cite{gpt4vsom}) output SoM mark.}
\label{temp:drag:som}
\end{table}

\begin{table}[h]
\centering
\begin{tcolorbox}[colback=gray!10, colframe=black, boxrule=0.5mm, width=\textwidth, rounded corners, boxsep=0mm, left=2mm, right=2mm, top=2mm, bottom=2mm]
\begin{tabularx}{\textwidth}{X}
I am working on a computer task that involves dragging elements from one place to another
Below are the OCR detection results (element name - bounding coordinates [[x1, y1], [x2, y2]]), which are separated by a colon ";".
\\
\texttt{\{ocr\_result\}}
\\
Could you assist me in navigating for action "{narration}"?\\
Please provide the location in the following format:\\
```
[x1, y1] -> [x2, y2]
'''\\
where [x1, y1] are the start coordinates and [x2, y2] are the destination coordinates.
Ensure that your response contains only the coordinates.\\
\end{tabularx}
\end{tcolorbox}
\caption{\textbf{Drag action template} that prompts LLMs (with OCR~\cite{azureocr}) output drag's coordinate [x1,y1] -> [x2, y2].}
\label{temp:drag:ocr}
\end{table}

\textbf{Action -- Type / Press.}
In Tab.~\ref{temp:type}, we present the template used by LLM to generate pyautogui code for keyboard actions.

\begin{table}[h]
\centering
\begin{tcolorbox}[colback=gray!10, colframe=black, boxrule=0.5mm, width=\textwidth, rounded corners, boxsep=0mm, left=2mm, right=2mm, top=2mm, bottom=2mm]
\begin{tabularx}{\textwidth}{X}
I'm working on a computer task involving typing or pressing keys. \\
Could you assist me in crafting a Python script using pyautogui to accomplish \texttt{\{goal\}}? where the key input element is "\texttt{\{element\}}".\\
I've already set up the environment. \\

Please provide the executable code directly and refrain from including other outputs or additional code blocks.
Ensure that your response contains only one code block formatted as follows:\\
```python\\
import pyautogui\\
pyautogui.press('ctrl')\\
```
\end{tabularx}
\end{tcolorbox}
\caption{\textbf{Type / Press action template} that prompts LLMs output \texttt{pyautogui} code.}
\label{temp:type}
\end{table}

\textbf{Action -- Scroll.}
In Tab.~\ref{temp:scroll}, we present the template used by LLM to predict scroll action, which is used for high-level planning. For mid-level planning, we remove the commentary component.

\begin{table}[h]
\centering
\begin{tcolorbox}[colback=gray!10, colframe=black, boxrule=0.5mm, width=\textwidth, rounded corners, boxsep=0mm, left=2mm, right=2mm, top=2mm, bottom=2mm]
\begin{tabularx}{\textwidth}{X}
I'm currently engaged in a computer-based task and need your assistance.\\
You are provided with an image of my screenshot.\\
Could you advise whether I need to scroll to see the complete element "\texttt{\{element\}}"?
Please note that even if the element appears partially, I still need to scroll to see it completely.
\\\\
{'A': 'No need to scroll.', 'B': 'Scroll down.', 'C': 'Scroll up.'}
\\\\
Please select the appropriate option and format your response as follows (Wrap options in square brackets):
\\
```
[A]
'''
\\\\
**Notably, only output options with square brackets**
\end{tabularx}
\end{tcolorbox}
\caption{\textbf{Scroll action template} that prompts LLMs to output a decision like scrolling (up/down) or not.}
\label{temp:scroll}
\end{table}

\textbf{Evaluation.}
In Tab.~\ref{temp:eval}, we display the evaluation template for \gptv.

\begin{table}[h]
\centering
\begin{tcolorbox}[colback=gray!10, colframe=black, boxrule=0.5mm, width=\textwidth, rounded corners, boxsep=0mm, left=2mm, right=2mm, top=2mm, bottom=2mm]
\begin{tabularx}{\textwidth}{X}
You are tasked with evaluating the quality of a software procedure plan.
Assess the prediction provided by an AI model against the human-generated ground truth and assign a correctness score to the prediction.

\textbf{Evaluation Criteria:}\\
1. \textit{Conciseness and Clarity}: The procedure plan should be straightforward and to the point.\\
2. \textit{Element Accuracy}: Pay attention to the precision of specific details like types of animation, text content, and design elements (e.g., 3d shape, color, shape). The prediction should accurately reflect these aspects as mentioned in the ground truth.\\
3. \textit{Commentary}: Provide a brief commentary in your response summarizing the accurate and inaccurate aspects of the prediction as evidence to support your scoring decision.\\
\\
\textbf{Correctness Score} (must be an integer):\\
- 0: Completely incorrect\\
- 1 to 3: Partially correct (with 1 being least accurate and 3 being more accurate)\\
- 4 to 5: Fully correct (with 4 being good and 5 being perfect)\\
\\
\textbf{Ground truth:}\\
\texttt{\{GT\}}\\
\\
\textbf{Prediction:}\\
\texttt{\{Pred\}}\\
\\
Considering the detailed elements and the overall process, please format your response as follows:

\texttt{[comment]: Summary of evaluation.}\\
\texttt{[score]: x}
\end{tabularx}
\end{tcolorbox}
\caption{Evaluation Prompt Template}
\label{temp:eval}
\end{table}


\clearpage
\bibliographystyle{unsrt}
\bibliography{ref}